%% file: 00_main.tex
\documentclass[10pt,twocolumn,letterpaper]{article}

\usepackage{iccv}
\usepackage{times}
\usepackage{epsfig}
\usepackage{graphicx}
\usepackage{amsmath}
\usepackage{amssymb}

\usepackage{subcaption}
\usepackage{commath}
\usepackage{booktabs}
\usepackage[numbers,sort]{natbib}
\usepackage{siunitx}
\usepackage{bold-extra}
\usepackage[T1]{fontenc}
\usepackage{contour}
\usepackage{microtype}      
\usepackage{booktabs}       
\usepackage{multirow}
\usepackage{multicol}
\usepackage{tabu}
\usepackage[symbol]{footmisc}
\usepackage{xr-hyper}

\newif\ifdraft
\draftfalse

\usepackage[pagebackref=true,breaklinks=true,colorlinks,bookmarks=false]{hyperref}

\iccvfinalcopy 

\def\iccvPaperID{} 

\ificcvfinal\fi

\input{commands.tex}

\begin{document}

\title{Nerfies: Deformable Neural Radiance Fields}

\author{
Keunhong Park\textsuperscript{1}\footnotemark
\and Utkarsh Sinha\textsuperscript{2}
\and Jonathan T. Barron\textsuperscript{2}
\and Sofien Bouaziz\textsuperscript{2}
\and Dan B Goldman\textsuperscript{2}
\and Steven M. Seitz\textsuperscript{1,2}
\and Ricardo Martin-Brualla\textsuperscript{2} \smallskip
\and
\textsuperscript{1}University of Washington\qquad 
\textsuperscript{2}Google Research \\ \rurl{nerfies.github.io}
}

\twocolumn[{
\renewcommand\twocolumn[1][]{#1}
\maketitle
\input{f01_teaser}
}]

\footnotetext[1]{Work done while the author was an intern at Google.}

\input{01_abstract}
\input{02_intro}
\input{03_related_work}

\input{04_method}
\input{05_evaluation}

\input{06_conclusion}

\input{A1_appendix}

{\small
\bibliographystyle{ieee_fullname}
\bibliography{00_main}
}

\end{document}

%% file: commands.tex
\usepackage{soul}
\usepackage{amsmath}
\usepackage{amssymb}
\usepackage{amsfonts}
\usepackage{mathtools}
\usepackage{bm}
\usepackage{xfrac}
\usepackage{xspace}
\usepackage[makeroom]{cancel}
\usepackage{makecell}

\usepackage{color}
\usepackage[table,xcdraw]{xcolor}

\usepackage[abs]{overpic}

\definecolor{myyellow}{rgb}{1,1, 0.6}
\definecolor{myorange}{rgb}{1, 0.8, 0.6}
\definecolor{myred}{rgb}{1, 0.6, 0.6}

\newcommand\rurl[1]{%
  \href{https://#1}{\nolinkurl{#1}}%
}

\renewcommand{\figref}[1]{Fig.~\ref{#1}}
\newcommand{\tabref}[1]{Tab.~\ref{#1}}

\newcommand{\mat}[1]{\bm{\mathrm{#1}}}

\newcommand{\latent}{\mat{\omega}_i}
\newcommand{\appearance}{\mat{\psi}_i}

\DeclareMathOperator{\tr}{tr}
\DeclareMathOperator{\diag}{diag}

\newcommand{\clamp}[1]{\operatorname{clamp}(#1)}

\newcommand{\sethree}[0]{\mathfrak{se}(3)}
\newcommand{\SETHREE}[0]{\text{SE}(3)}
\newcommand{\sothree}[0]{\mathfrak{so}(3)}
\newcommand{\SOTHREE}[0]{\text{SO}(3)}
\newcommand{\logrot}[0]{\mat{r}}
\newcommand{\unitlogrot}[0]{\hat{\logrot}}
\newcommand{\skewmat}[1]{[\mat{#1}]_{\times}}

\newcommand{\mypar}[1]{\noindent\textbf{#1:}}

\newcommand{\CIRCLE}[1]{\raisebox{.5pt}{\footnotesize \textcircled{\raisebox{-.6pt}{#1}}}}

\def\figcell#1#2#3{\begin{subfigure}{#1\columnwidth}\centering\includegraphics[width=\textwidth]{#2}
\def\temp{#3}\ifx\temp\empty\else\caption*{\centering{#3}}\fi\end{subfigure}}

\def\figcellc#1#2#3{\begin{subfigure}{#1\columnwidth}\centering\includegraphics[width=\textwidth]{#2} \def\temp{#3}\ifx\temp\empty\else\caption{#3}\fi\end{subfigure}}

\def\figcellt#1#2#3#4{\begin{subfigure}{#1\columnwidth}\centering\includegraphics[width=\textwidth,#4]{#2} \def\temp{#3}\ifx\temp\empty\else\caption*{#3}\fi\end{subfigure}}

\newcommand{\figcellbox}[1]{\fcolorbox{black}{white}{#1}}

\def\figcelltb#1#2#3#4{\begin{subfigure}{#1\columnwidth}\centering\figcellbox{\includegraphics[width=\textwidth,#4]{#2}} \def\temp{#3}\ifx\temp\empty\else\caption*{#3}\fi\end{subfigure}}

\def\figcellb#1#2#3{\begin{subfigure}{#1\columnwidth}\centering\figcellbox{\includegraphics[width=\textwidth]{#2}}
\def\temp{#3}\ifx\temp\empty\else\caption*{\centering{#3}}\fi\end{subfigure}}

\makeatletter
\newcommand{\thickhline}{%
    \noalign {\ifnum 0=`}\fi \hrule height 1pt
    \futurelet \reserved@a \@xhline
}
\newcolumntype{"}{@{\hskip\tabcolsep\vrule width 1pt\hskip\tabcolsep}}
\makeatother

\newcommand{\textimage}[4]{
	\begin{overpic}[width=2.35cm,unit=1mm,clip,trim=#1]{figures/vrig_results_v2/#2}
	\put (9.9,0.7) {\sethlcolor{white}\footnotesize\hl{$#3 / #4$}}
    \end{overpic}
}

%% file: f01_teaser.tex
\fboxsep=0pt 
\fboxrule=0.4pt 
\newcommand{\teaserbox}[1]{\fcolorbox{black}{white}{#1}}

\begin{center}
    \vspace{-5mm}
	\captionsetup[sub]{labelformat=parens}
	\begingroup
        \captionsetup{type=figure}
    	\begin{subfigure}{0.34\columnwidth}
        	\includegraphics[height=39mm]{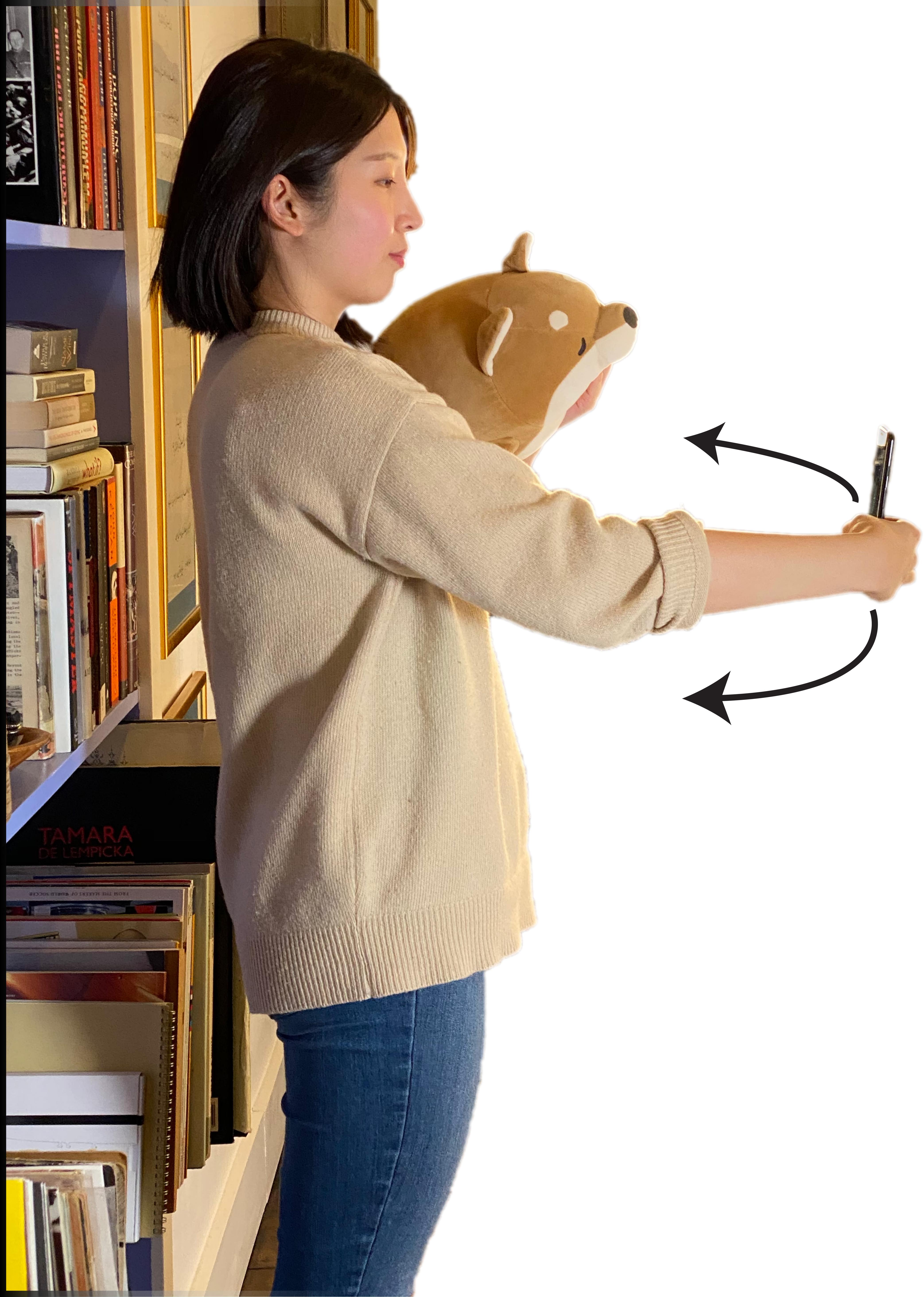}
    		\subcaption{\centering casual capture}
    	\end{subfigure}
    	\begin{subfigure}{0.35\columnwidth}
    		\centering
        	\includegraphics[height=39mm]{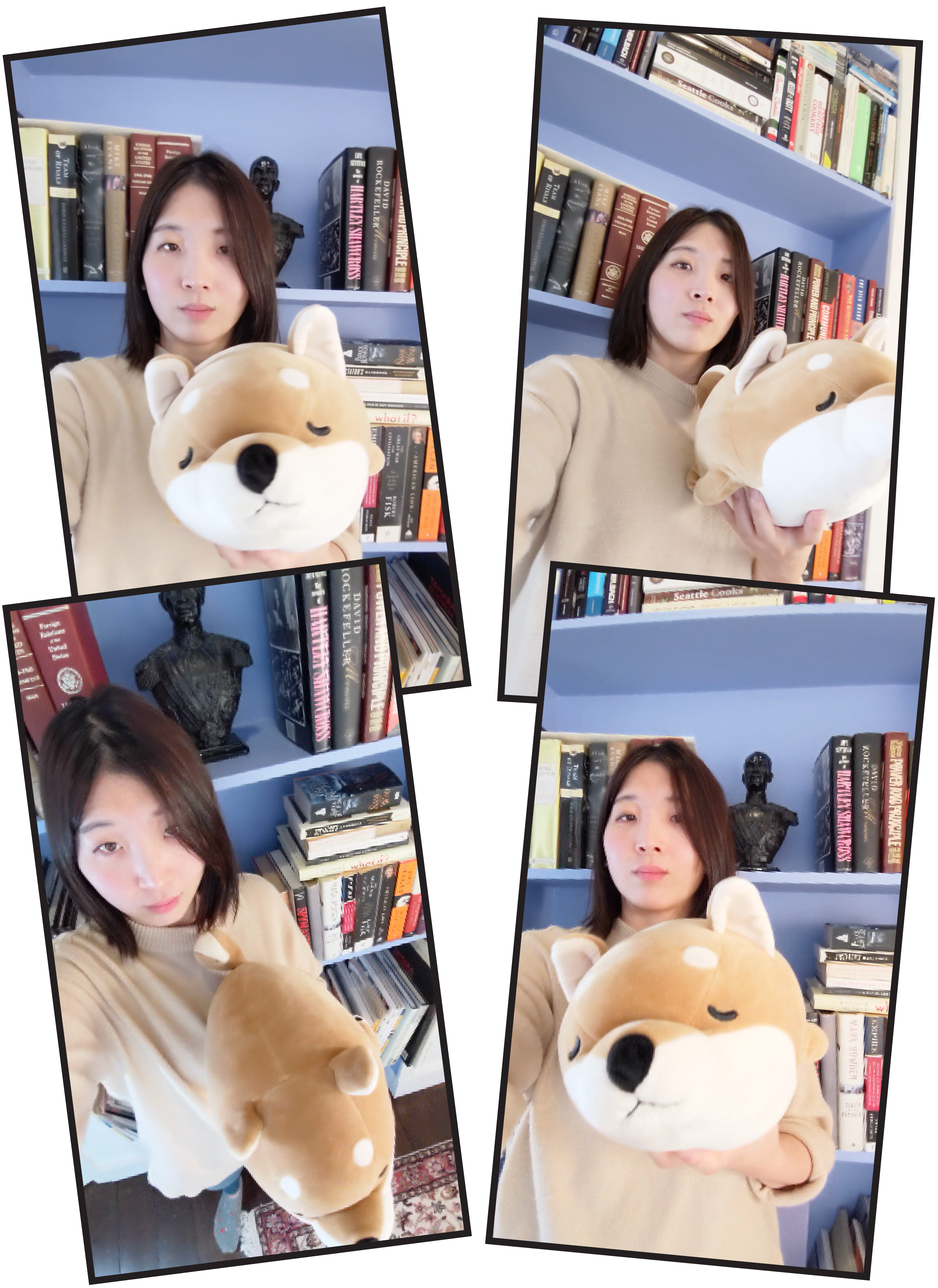}
    		\subcaption{input images}
    	\end{subfigure}
    	\begin{subfigure}{0.92\columnwidth}
    		\centering
        	\begin{subfigure}{0.46\columnwidth}
            	\teaserbox{\includegraphics[height=39mm,clip,trim=0 50 0 30]{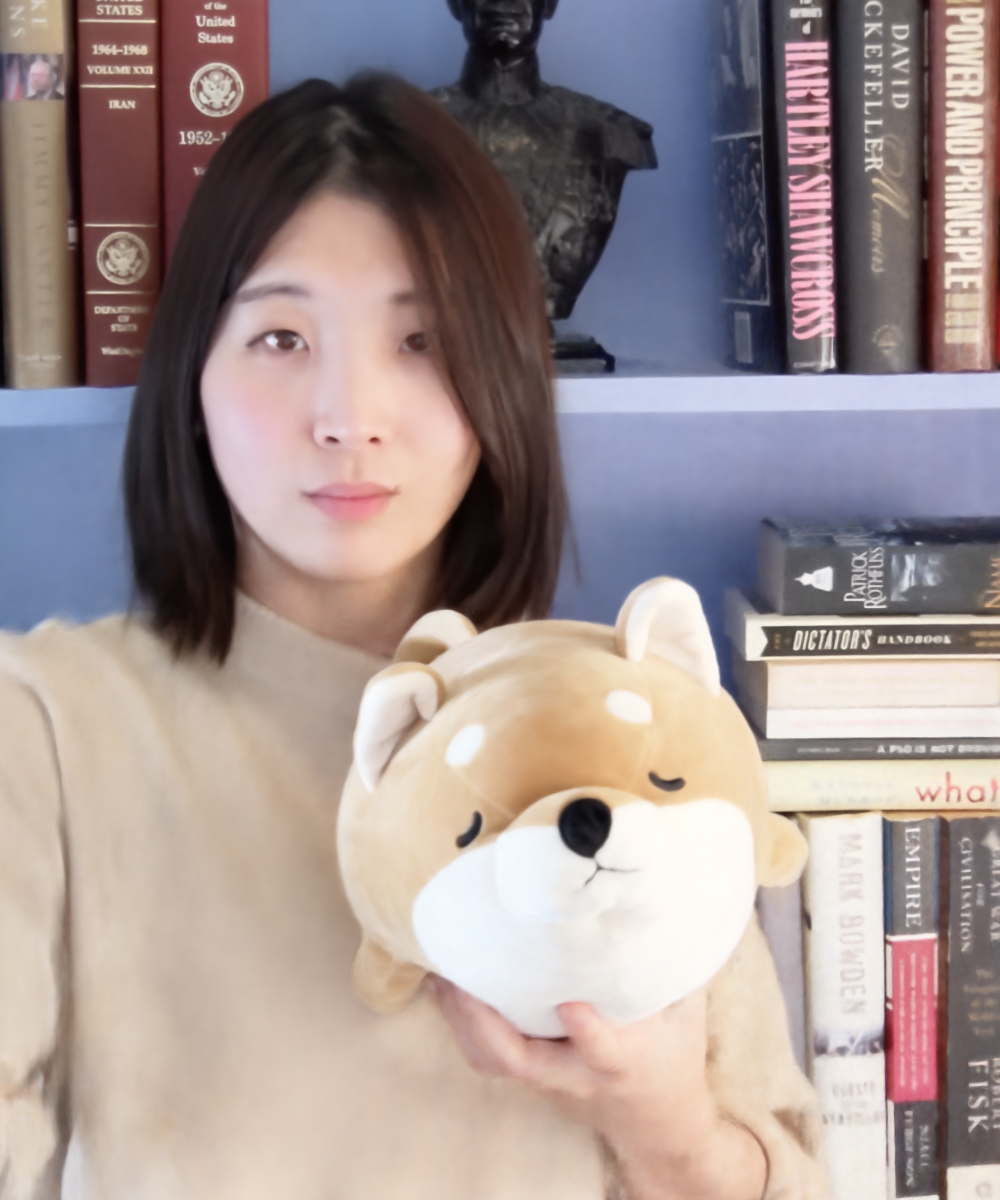}}
        	\end{subfigure}
        	\begin{subfigure}{0.46\columnwidth}
            	\teaserbox{\includegraphics[height=39mm,clip,trim=0 50 0 30]{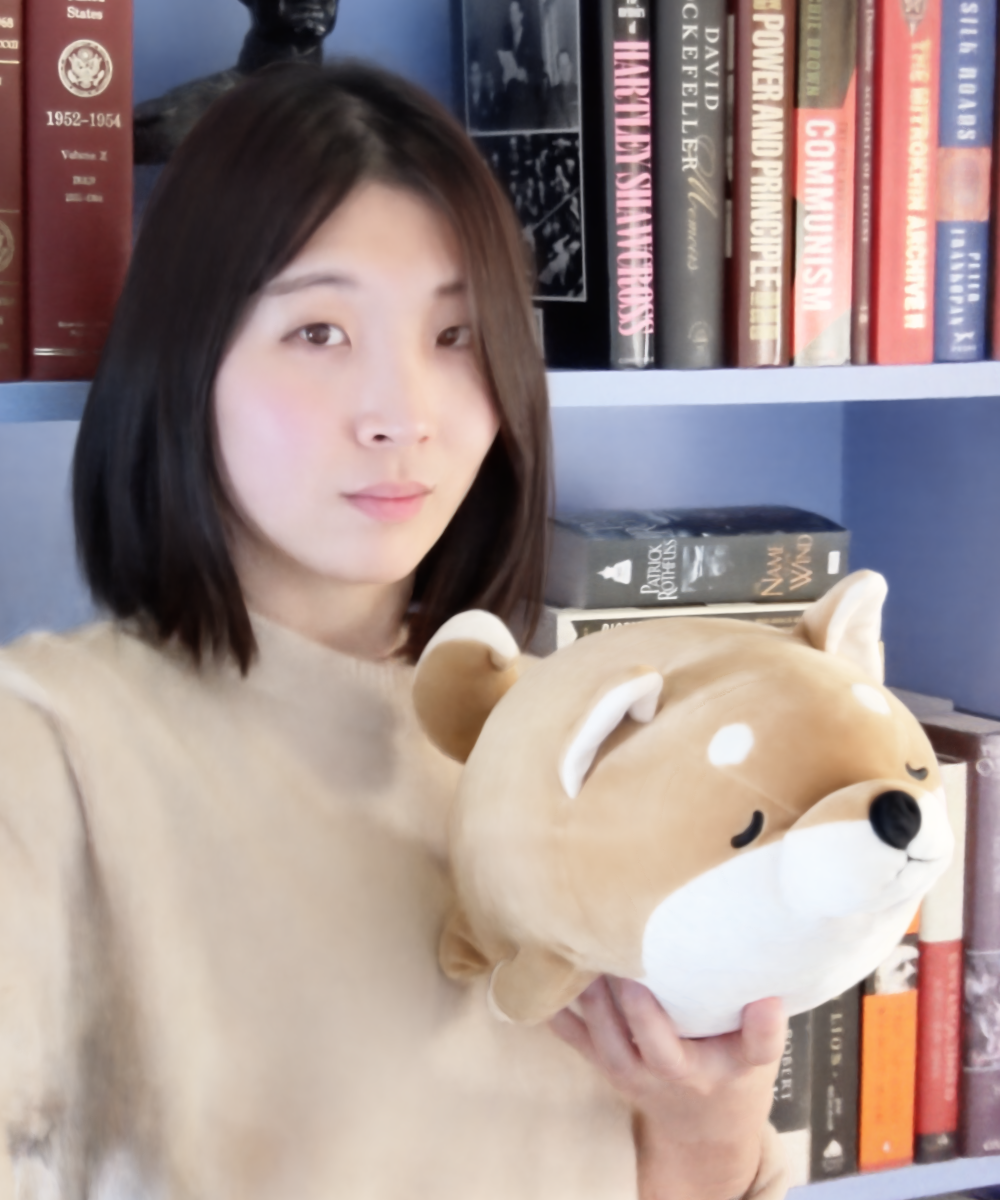}}
        	\end{subfigure}
        	\subcaption{nerfie novel views }
    	\end{subfigure}
    	\begin{subfigure}{0.41\columnwidth}
    		\centering
        	\teaserbox{\includegraphics[height=39mm,clip,trim=0 50 0 0]{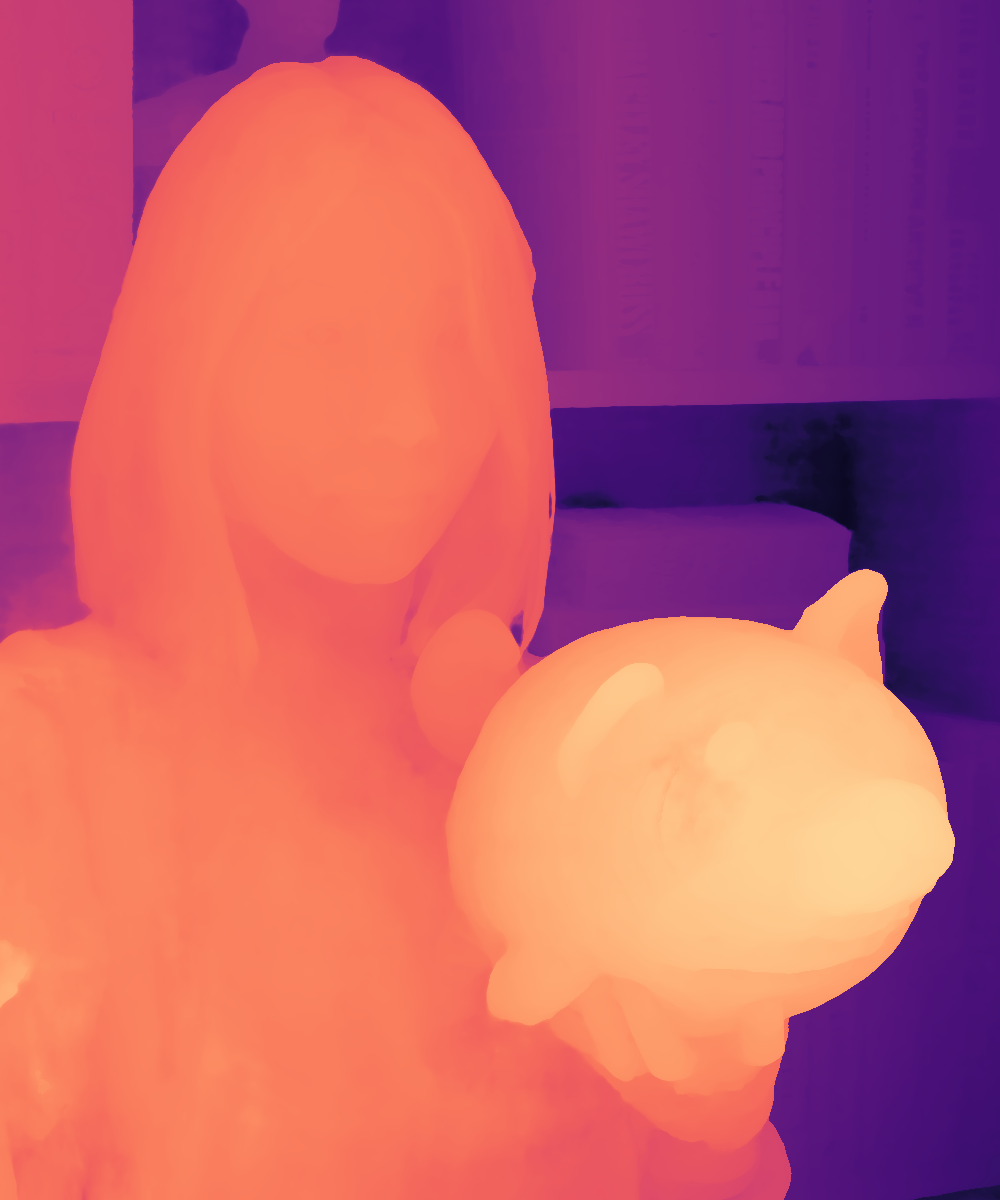}}
        	\subcaption{novel view depth}
    	\end{subfigure}
	\endgroup
	\vspace{-2mm}
	\setcounter{figure}{0} 
	\captionsetup{type=figure}
	\captionof{figure}{We reconstruct photo-realistic {\em nerfies} from a user casually waving a mobile phone (a). Our system uses selfie photos/videos (b) to produce a free-viewpoint representation with accurate renders (c) and geometry (d). Please see video.}
   \label{fig:teaser}
\end{center}

%% file: 01_abstract.tex
\begin{abstract}
We present the first method capable of photorealistically reconstructing deformable scenes using photos/videos captured casually from mobile phones.
Our approach augments neural radiance fields (NeRF) by optimizing an additional continuous volumetric deformation field that warps each observed point into a canonical 5D NeRF.  
We observe that these NeRF-like deformation fields are prone to local minima, and propose a coarse-to-fine optimization method for coordinate-based models that allows for more robust optimization.
By adapting principles from geometry processing and physical simulation to NeRF-like models, we propose an elastic regularization of the deformation field that further improves robustness.
We show that our method can turn casually captured selfie photos/videos into deformable NeRF models that allow for photorealistic renderings of the subject from arbitrary viewpoints, which we dub ``nerfies.'' We evaluate our method by collecting time-synchronized data using a rig with two mobile phones, yielding train/validation images of the same pose at different viewpoints. We show that our method faithfully reconstructs non-rigidly deforming scenes and reproduces unseen views with high fidelity.
\end{abstract}
\vspace{-3pt}

%% file: 02_intro.tex
\section{Introduction}

High quality 3D human scanning has come a long way -- but the best results currently require a specialized lab with many synchronized lights and cameras, e.g., \cite{collet2015high,dou2016fusion4d,guo2019therelightables}.
What if you could capture a photorealistic model of yourself (or someone else) just by waving your mobile phone camera? Such a capability would dramatically increase accessibility and applications of 3D modeling technology.

Modeling people with hand-held cameras is especially challenging due both to 1) nonrigidity -- our inability to stay perfectly still, and 2) challenging materials like hair, glasses, and earrings that violate assumptions used in most reconstruction methods.  In this paper we introduce an approach to address both of these challenges, by generalizing Neural Radiance Fields (NeRF)~\cite{mildenhall2020nerf} to model shape deformations. 
Our technique recovers high fidelity 3D reconstructions from short videos, providing free-viewpoint visualizations while accurately capturing hair, glasses, and other complex, view-dependent materials, as shown in Figure~\ref{fig:teaser}.  A special case of particular interest is capturing a 3D self-portrait -- we call such casual 3D selfie reconstructions \emph{nerfies}.

Rather than represent shape explicitly, NeRF \cite{mildenhall2020nerf} uses a neural network to encode color and density as a function of location and viewing angle, and generates novel views using volume rendering.  Their approach produces 3D visualizations of unprecedented quality, faithfully representing thin structures, semi-transparent materials, and view-dependent effects. 
To model non-rigidly deforming scenes, we generalize NeRF by introducing an additional component: A canonical NeRF model serves as a template for all the observations, supplemented by a deformation field for each observation that warps 3D points in the frame of reference of an observation into the frame of reference of the canonical model. We represent this deformation field as a multi-layer perceptron (MLP), similar to the radiance field in NeRF. This deformation field is conditioned on a per-image learned latent code, allowing it to vary between observations.

Without constraints, the deformation fields are prone to distortions and over-fitting. 
We employ a similar approach to the elastic energy formulations that have seen success for mesh fitting \cite{ARAP_modeling:2007,bouaziz2014dynamics,chao2010simple,sumner2007embedded}.
However, our volumetric deformation field formulation greatly simplifies such regularization, because we can easily compute the Jacobian of the deformation field through automatic differentiation, and directly regularize its singular values.

To robustly optimize the deformation field, we propose a novel coarse-to-fine optimization scheme that modulates the components of the input positional encoding of the deformation field network by frequency. By zeroing out the high frequencies at the start of optimization, the network is limited to learn smooth deformations, which are later refined as higher frequencies are introduced into the optimization.

For evaluation, we capture image sequences from a rig of two synchronized, rigidly attached, calibrated cameras, and use the reconstruction from one camera to predict views from the other.
We plan to release the code and data.

In summary, our contributions are: \CIRCLE{1}~an extension to NeRF to handle non-rigidly deforming objects that optimizes a deformation field per observation; \CIRCLE{2}~rigidity priors suitable for deformation fields defined by neural networks; \CIRCLE{3}~a coarse-to-fine regularization approach that modulates the capacity of the deformation field to model high frequencies during optimization; \CIRCLE{4}~a system to reconstruct free-viewpoint selfies from casual mobile phone captures.

%% file: 03_related_work.tex
\section{Related Work}

\mypar{Non-Rigid Reconstruction} Non-rigid reconstruction decomposes a scene into a geometric model and a deformation model that deforms the geometric model for each observation. Earlier works focused on sparse representations such as keypoints projected onto 2D images~\cite{bregler2000recovering, torresani2008nonrigid}, making the problem highly ambiguous. Multi-view captures~\cite{collet2015high,dou2016fusion4d} simplify the problem to one of registering and fusing 3D scans~\cite{li2012temporally}. DynamicFusion~\cite{newcombe2015dynamicfusion} uses a single RGBD camera moving in space, solving jointly for a canonical model, a deformation, and camera pose. 
More recently, learning-based methods have been used to find correspondences useful for non-rigid reconstruction~\cite{bozic2020neural, schmidt2015dart}.
Unlike prior work, our method does not require depth nor multi-view capture systems and works on monocular RGB inputs. 
Most similar to our work, Neural Volumes~\cite{lombardi2019neural} learns a 3D representation of a deformable scene using a voxel grid and warp field regressed from a 3D CNN. However, their method requires dozens of synchronized cameras and our evaluation shows that it does not extend to sequences captured from a single camera. 
Yoon~\etal~\cite{yoon2020novel} reconstruct dynamic scenes from moving camera trajectories, but their method relies on strong semantic priors, in the form of monocular depth estimation, which are combined with multi-view cues.
OFlow~\cite{Niemeyer_2019_ICCV} solves for temporal flow-fields using ODEs, and thus requires temporal information. 
ShapeFlow~\cite{jiang2020shapeflow} learns 3D shapes a divergence-free deformations of a learned template. Instead, we propose an elastic energy regularization.

\mypar{Domain-Specific Modeling}
Many reconstruction methods use domain-specific knowledge to model the shape and appearance of categories with limited topological variation, such as faces~\cite{blanz1999morphable, bouaziz2013online, booth2018large}, human bodies~\cite{loper2015smpl, xu2020ghum}, and animals~\cite{cashman2012shape, zuffi20173d}. Although some methods show impressive results in monocular face reconstruction from color and RGBD cameras~\cite{zollhofer2018state}, such models often lack detail (e.g., hair), or do not model certain aspects of a category (e.g., eyewear or garments). Recently, image translation networks have been applied to improve the realism of composited facial edits~\cite{fried2019text, kim2018deepvideoportraits}. In contrast, our work does not rely on domain-specific knowledge, enabling us to model the whole scene, including eyeglasses and hair for human subjects.

\begin{figure*}[t]
\centering
\includegraphics*[width = 0.85\linewidth]{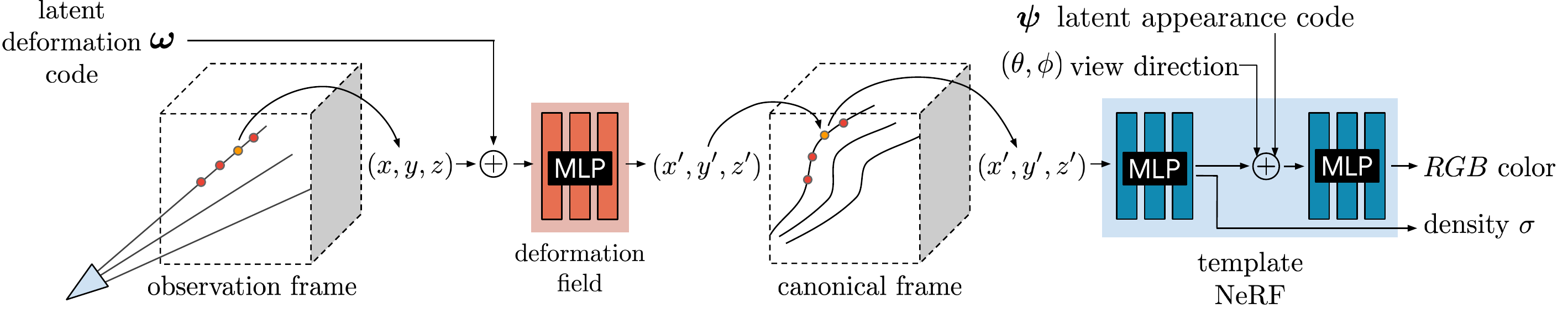}
\vspace{-5pt}
\caption{We associate a latent deformation code ($\mat{\omega}$) and an appearance code ($\mat{\psi}$)  to each image. We trace the camera rays in the observation frame and transform samples along the ray to the canonical frame using a deformation field encoded as an MLP that is conditioned on the deformation code $\boldsymbol{\omega}$. We query the template NeRF~\cite{mildenhall2020nerf} using the transformed sample $(x',y',z')$, viewing direction $(\theta,\phi)$ and appearance code $\boldsymbol{\psi}$ as inputs to the MLP and integrate samples along the ray following NeRF. 
}
\label{fig:architecture}

\vspace{-5pt}
\end{figure*}

\mypar{Coordinate-based Models}
Our method builds on the recent success of coordinate-based models, which encode a spatial field in the weights of a multilayer perceptron (MLP) and require significantly less memory compared to discrete representations. These methods
have been used to represent shapes~\cite{Chen_2019_CVPR, mescheder2019occupancy, park2019deepsdf} and scenes~\cite{mildenhall2020nerf,sitzmann2019srns}. Of particular interest are NeRFs~\cite{mildenhall2020nerf}, that use periodic positional encoding layers~\cite{sitzmann2019siren, tancik2020fourfeat} to increase resolution, and whose formulation has be extended to handle different lighting conditions~\cite{bi2020neural, martinbrualla2020nerfw}, transient objects~\cite{martinbrualla2020nerfw}, large scenes~\cite{liu2020neural, zhang2020nerf} and to model object categories~\cite{schwarz2020graf}. Our work extends NeRFs to handle non-rigid scenes. 

\mypar{Concurrent Work} Two concurrent works~\cite{pumarola2020dnerf, tretschk2021nonrigid} propose to represent deformable scenes using a translation field in conjunction with a template. This is similar to our framework with the following differences: \CIRCLE{1}~we condition the deformation with a per-example latent~\cite{bojanowski2018optimizing} instead of time~\cite{pumarola2020dnerf}; \CIRCLE{2}~propose an as-rigid-as-possible regularization of the deformation field while NR-NeRF~\cite{tretschk2021nonrigid} penalizes the divergence of the translation field; \CIRCLE{3}~propose a coarse-to-fine regularization to prevent getting stuck in local minima; and \CIRCLE{4}~propose an improved SE(3) parameterization of the deformation field. Other concurrent works~\cite{xian2020spacetime,li2020neural} reconstruct space-time videos by recovering time-varying NeRFs while leveraging external supervision such as monocular depth estimation and flow-estimation to resolve ambiguities.

%% file: 04_method.tex
\section{Deformable Neural Radiance Fields}

Here we describe our method for modeling non-rigidly deforming scenes given a set of casually captured images of the scene.  We decompose a non-rigidly deforming scene into a template volume represented as a neural radiance field (NeRF)~\cite{mildenhall2020nerf} (\secref{sec:nerf}) and a per-observation deformation field (\secref{sec:deformation-field}) that associates a point in observation coordinates to a point on the template (overview in \figref{fig:architecture}). The deformation field is our key extension to NeRF and allows us to represent moving subjects.  Jointly optimizing a NeRF together with a deformation field leads to an under-constrained optimization problem. We therefore introduce an elastic regularization on the deformation (\secref{sec:arap}), a background regularization (\secref{sec:background-regularization}), and a continuous, coarse-to-fine annealing technique that avoids bad local minima (\secref{sec:annealing}). 

\subsection{Neural Radiance Fields}
\label{sec:nerf}

A neural radiance field (NeRF) is a continuous, volumetric representation. It is a function $F: (\mat{x},\mat{d}, \appearance) \rightarrow (\mat{c}, \sigma)$ which maps a 3D position $\mat{x} = (x,y,z)$ and viewing direction $\mat{d}=(\phi,\theta)$ to a color $\mat{c}=(r,g,b)$ and density $\sigma$. In practice, NeRF maps the inputs $\mat{x}$ and $\mat{d}$ using a sinusoidal positional encoding $\gamma: \mathbb{R}^3 \rightarrow \mathbb{R}^{3 + 6m}$ defined as
$
    \gamma(\mat{x}) = \left(\mat{x}, \cdots,\sin{(2^{k}\pi\mat{x})},\cos{(2^{k}\pi\mat{x})},\cdots\right)
$, where $m$ is a hyper-parameter that controls the total number of frequency bands

and $k\in\{0,\ldots,m-1\}$. 
This function projects a coordinate vector $\mat{x} \in \mathbb{R}^3$ to a high dimensional space using a set of sine and cosine functions of increasing frequencies. This allows the MLP to model high-frequency signals in low-frequency domains as shown in~\cite{tancik2020fourfeat}.
Coupled with volume rendering techniques, NeRFs can represent scenes with photo-realistic quality. We build upon NeRF to tackle the problem of capturing deformable scenes.

Similar to NeRF-W~\cite{martinbrualla2020nerfw}, we also provide an appearance latent code $\appearance$ for each observed frame $i\in\{1,\ldots,n\}$ that modulates the color output to handle appearance variations between input frames, e.g., exposure and white balance.

The NeRF training procedure relies on the fact that given a 3D scene, two intersecting rays from two different cameras should yield the same color. Disregarding specular reflection and transmission, this assumption is true for all static scenes. Unfortunately, many scenes are not completely static; e.g., it is hard for people to stay completely still when posing for a photo, or worse, when waving a phone when capturing themselves in a selfie video.

\subsection{Neural Deformation Fields}
\label{sec:deformation-field}

With the understanding of this limitation, we extend NeRF to allow the reconstruction of non-rigidly deforming scenes. Instead of directly casting rays through a NeRF, we use it as a canonical template of the scene. This template contains the relative structure and appearance of the scene while a rendering will use a non-rigidly deformed version of the template (see~\figref{fig:observation_vs_canonical} for an example). DynamicFusion~\cite{newcombe2015dynamicfusion} and Neural Volumes~\cite{lombardi2019neural} also model a template and a per-frame deformation, but the deformation is defined on mesh points and on a voxel grid respectively, whereas we model it as a continuous function using an MLP.

We employ an observation-to-canonical deformation for every frame $i\in\{1,\ldots,n\}$, where $n$ is the number of observed frames. This defines a mapping $T_i: \mat{x} \rightarrow \mat{x}'$ that maps all observation-space coordinates $\mat{x}$ to a canonical-space coordinate $\mat{x}'$. We model the deformation fields for all time steps using a mapping $T: (\mat{x}, \latent) \rightarrow \mat{x}'$, which is conditioned on a per-frame learned latent deformation code $\latent$. Each latent code encodes the state of the scene in frame $i$. Given a canonical-space radiance field $F$ and a observation-to-canonical mapping $T$, the observation-space radiance field can be evaluated as:
\begin{align}
    G(\mat{x},\mat{d}, \appearance, \latent) = F\left(T(\mat{x}, \latent),\mat{d}, \appearance\right)\,.
\end{align}
When rendering, we simply cast rays and sample points in the observation frame and then use the deformation field to map the sampled points to the template, see \figref{fig:architecture}.

A simple model of deformation is a displacement field $V: (\mat{x}, \latent) \rightarrow \mat{t}$, defining the transformation as $T(\mat{x}, \latent) = \mat{x} + V(\mat{x}, \latent)$. This formulation is sufficient to represent all continuous deformations; however, rotating a group of points with a translation field requires a different translation for each point, making it difficult to rotate regions of the scene simultaneously. We therefore formulate the deformation using a dense SE(3) field $W: (\mat{x}, \latent) \rightarrow \SETHREE$. An SE(3) transform encodes rigid motion, allowing us to rotate a set of distant points with the same parameters.  

We encode a rigid transform as a screw axis~\cite{lynch2017modern} $\mathcal{S} = (\logrot; \mat{v})\in\mathbb{R}^6$.
Note that $\logrot\in\sothree$ encodes a rotation where $\unitlogrot = \logrot/\norm{\logrot}$ is the axis of rotation and $\theta = \norm{\logrot}$ is the angle of rotation. The exponential of $\logrot$ (also known as Rodrigues' formula~\cite{rodrigues1816}) yields a rotation matrix $e^{\logrot} \in \SOTHREE$:
\begin{align}
    e^{\logrot} \equiv e^{[\logrot]_\times} = \mat{I} + \frac{\sin\theta}{\theta}[\logrot]_\times + \frac{1-\cos\theta}{\theta^2}[\logrot]_\times^2\,,
\end{align}
where $[\mat{x}]_\times$ denotes the cross-product matrix of a vector $\mat{x}$.

Similarly, the translation encoded by the screw motion $\mathcal{S}$ can be recovered as $\mat{p} = \mat{G}\mat{v}$ where
\begin{align}
    \mat{G} &= \mat{I} + \frac{1-\cos{\theta}}{\theta^2}[\logrot]_\times + \frac{\theta - \sin\theta}{\theta^3}[\logrot]^2_\times\,.
\end{align}
Combining these formulas and using the exponential map, we get the transformed point as $\mat{x}' = e^{\mathcal{S}} \mat{x} = e^{\logrot}  \mat{x} +  \mat{p}$.

As mentioned before, we encode the transformation field in an MLP $W: (\mat{x},\mat{\omega}_i) \rightarrow (\logrot,\mat{v})$
using a NeRF-like architecture, and represent the transformation of every frame $i$ by conditioning on a latent code $\latent$. We optimize the latent code through an embedding layer~\cite{bojanowski2018optimizing}. Like with the template, we map the input $\mat{x}$ using positional encoding $\gamma_\alpha$ (see \secref{sec:annealing}). An important property of the $\sethree$ representation is that $e^{\mathcal{S}}$ is the identity when $\mathcal{S} = \mat{0}$. We therefore initialize the weights of the last layer of the MLP from $\mathcal{U}(-10^{-5},10^{-5})$ to initialize the deformation near the identity.

\input{f0x_observation_vs_canonical}

\subsection{Elastic Regularization}
\label{sec:arap}

The deformation field adds ambiguities that make optimization more challenging. For example, an object moving backwards is visually equivalent to it shrinking in size, with many solutions in between.
These ambiguities lead to under-constrained optimization problems which yield implausible results and artifacts (see \figref{fig:latent_interpolation}). It is therefore crucial to introduce priors that lead to a more plausible solution.

It is common in geometry processing and physics simulation to model non-rigid deformations using elastic energies measuring the deviation of local deformations from a rigid motion ~\cite{ARAP_modeling:2007, sumner2007embedded, chao2010simple, bouaziz2014dynamics}. In the vision community, these energies have been extensively used for the reconstruction and tracking of non-rigid scenes and objects ~\cite{zollhofer2014real, newcombe2015dynamicfusion, dou2016fusion4d} making them good candidates for our approach. 
While they have been most commonly used for discretized surfaces, e.g., meshes, we can apply a similar concept in the context of our continuous deformation field.

\input{f0x_elastic_example}

\mypar{Elastic Energy} For a fixed latent code $\latent$, our deformation field $T$ is a non-linear mapping from observation-coordinates in $\mathbb{R}^3$ to canonical coordinates in $\mathbb{R}^3$. The Jacobian $\mat{J}_{T}(\mat{x})$ of this mapping at a point $\mat{x} \in \mathbb{R}^3$ describes the best linear approximation of the transformation at that point. We can therefore control the local behavior of the deformation through $\mat{J}_T$ \cite{sifakis2012FEM}. Note that unlike other approaches using discretized surfaces, our continuous formulation allows us to directly compute $\mat{J}_T$ through automatic differentiation of the MLP.
There are several ways to penalize the deviation of the Jacobian $\mat{J}_{T}$ from a rigid transformation. Considering the singular-value decomposition of the Jacobian $\mat{J}_{T} = \mat{U}\mat{\Sigma}\mat{V}^T$, multiple approaches \cite{chao2010simple, bouaziz2014dynamics} penalize the deviation from the closest rotation as $\norm{\mat{J}_{T}-\mat{R}}_F^2$, where $\mat{R}=\mat{VU}^T$ and $\norm{\cdot}_F$ is the Frobenius norm. We opt to directly work with the singular values of $\mat{J}_T$ and measure its deviation from the identity. The log of the singular values gives equal weight to a contraction and expansion of the same factor, and we found it to perform better. We therefore penalize the deviation of the log singular values from zero:
\begin{align}
    L_\text{elastic}(\mat{x}) = \norm{\log{\mat{\Sigma}}-\log{\mat{I}}}^2_F = \norm{\log{\mat{\Sigma}}}^2_F\,,
\end{align}
where $\log$ here is the matrix logarithm. 

\mypar{Robustness} Although humans are mostly rigid, there are some movements which can break our assumption of local rigidity, e.g., facial expressions which locally stretch and compress our skin. We therefore remap the elastic energy defined above using a robust loss:
\begin{gather}
    L_\text{elastic-r}(\mat{x}) = \rho\left(\norm{\log{\mat{\Sigma}}}_F, c\right), \\
    \rho(x, c) = \frac{2 (\sfrac{x}{c})^2}{(\sfrac{x}{c})^2 + 4} \,.
\end{gather}
where $\rho(\cdot)$ is the Geman-McClure robust error function~\cite{geman1985bayesian} parameterized with hyperparameter $c=0.03$ as per Barron~\cite{barron2019general}. This robust error function causes the gradients of the loss to fall off to zero for large values of the argument, thereby reducing the influence of outliers during training.

\mypar{Weighting} We allow the deformation field to behave freely in empty space, since the subject moving relative to the background requires a non-rigid deformation somewhere in space. We therefore weight the elastic penalty at each sample along the ray by its contribution to the rendered view, \ie $w_i$ in Eqn. 5 of NeRF~\cite{mildenhall2020nerf}.

\input{f0x_coarse_to_fine_optimization}

\subsection{Background Regularization}
\label{sec:background-regularization}

The deformation field is unconstrained and therefore everything is free to move around. We optionally add a regularization term which prevents the background from moving. Given a set of 3D points in the scene which we know should be static, we can penalize any deformations at these points. For example, camera registration using structure from motion produces a set of 3D feature points that behave rigidly across at least some set of observations. Given these static 3D points $\{\mat{x}_1\,\ldots,\mat{x}_K\}$, we penalize movement as:
\begin{align}
    L_\text{bg} = \frac{1}{K}\sum_{k=1}^K \norm{T(\mat{x}_k) - \mat{x}_k}_2 \,.
\end{align}
In addition to keeping the background points from moving, this regularization also has the benefit of aligning the observation coordinate frame to the canonical coordinate frame. 

\subsection{Coarse-to-Fine Deformation Regularization}
\label{sec:annealing}

A common trade-off that arises during registration and flow estimation is the choice between modeling minute versus large motions, that can lead to overly smooth results or incorrect registration (local minima). Coarse-to-fine strategies circumvent the issue by first solving the problem in low-resolution, where motion is small, and iteratively upscaling the solution and refining it~\cite{lucas1981iterative}. We observe that our deformation model suffers from similar issues, and propose a coarse-to-fine regularization to mitigate them.

Recall the positional encoding parameter $m$ introduced in \secref{sec:nerf} that controls the number of frequency bands used in the encoding. Tancik \etal~\cite{tancik2020fourfeat} show that controls it the smoothness of the network: a low value of $m$ results in a low-frequency bias (low resolution) while a higher value of $m$ results in a higher-frequency bias (high resolution).

Consider a motion like in \figref{fig:coarse_to_fine}, where subject rotates their head and smiles. 
With a small $m$ for the deformation field, the model cannot capture the minute motion of the smile; conversely, with a larger $m$, the model fails to correctly rotate the head because the template overfits to an underoptimized deformation field.
To overcome this trade-off, we propose a coarse-to-fine approach that starts with a low-frequency bias and ends with a high-frequency bias.

\input{f0x_latent_interpolation_v2}

Tancik \etal~\cite{tancik2020fourfeat} show that positional encoding can be interpreted in terms of the Neural Tangent Kernel (NTK)~\cite{jacot2018neural} of NeRF's MLP: a stationary interpolating kernel where $m$ controls a tunable ``bandwidth'' of that kernel. A small number of frequencies induces a wide kernel which causes under-fitting of the data, while a large number of frequencies induces a narrow kernel causing over-fitting.
With this in mind, we propose a method to smoothly anneal the bandwidth of the NTK by introducing a parameter $\alpha$ that windows the frequency bands of the positional encoding, akin to how coarse-to-fine optimization schemes solve for coarse solutions that are subsequently refined at higher resolutions.  
We define the weight for each frequency band $j$ as:
\begin{align}
    \label{eq:posenc_weights}
    w_j(\alpha) = \frac{(1 - \cos(\pi \clamp{\alpha-j,0,1})}{2} \,,
\end{align}
where linearly annealing the parameter $\alpha \in [0,m]$ can be interpreted as sliding a truncated Hann window (where the left side is clamped to 1 and the right side is clamped to 0) across the frequency bands. The positional encoding is then defined as 
$
    \gamma_{\alpha}(\mat{x}) = \left(\mat{x}, \cdots,w_k(\alpha)\sin{(2^{k}\pi\mat{x})},w_k(\alpha)\cos{(2^{k}\pi\mat{x})},\cdots\right)
$.
During training, we set $\alpha(t) = \tfrac{mt}{N}$ where $t$ is the current training iteration, and $N$ is a hyper-parameter for when $\alpha$ should reach the maximum number of frequencies $m$. We provide further analysis in the supplementary materials.

\input{f0x_hair}

\section{Nerfies: Casual Free-Viewpoint Selfies}

So far we have presented a generic method of reconstructing non-rigidly deforming scenes. We now present a key application of our system -- reconstructing high quality models of human subjects from casually captured selfies, which we dub \emph{``nerfies''}. Our system takes as input a sequence of selfie photos or a selfie video in which the user is standing mostly still. Users are instructed to wave the camera around their face, covering viewpoints within a $45^{\circ}$ cone. We observe that 20 second captures are sufficient.
In our method, we assume that the subject stands against a static background to enable a consistent geometric registration of the cameras. 
We filter blurry frames using the variance of the Laplacian~\cite{pech2000diatom}, keeping about $600$ frames per capture.

\mypar{Camera Registration} We seek a registration of the cameras with respect to the static background. We use COLMAP~\cite{schoenberger2016sfm} to compute pose for each image and camera intrinsics. This step assumes that enough features are present in the background to register the sequence.

\mypar{Foreground Segmentation} In some cases, SfM will match features on the moving subject, causing significant misalignment in the background. This is problematic in video captures with correlated frames. In those cases, we found it helpful to discard image features on the subject, which can be detected using a foreground segmentation network.

%% file: f0x_observation_vs_canonical.tex

\begin{figure}[t]
	\centering
	\begin{subfigure}{0.49\columnwidth}
		\centering
    	\includegraphics[width=\textwidth]{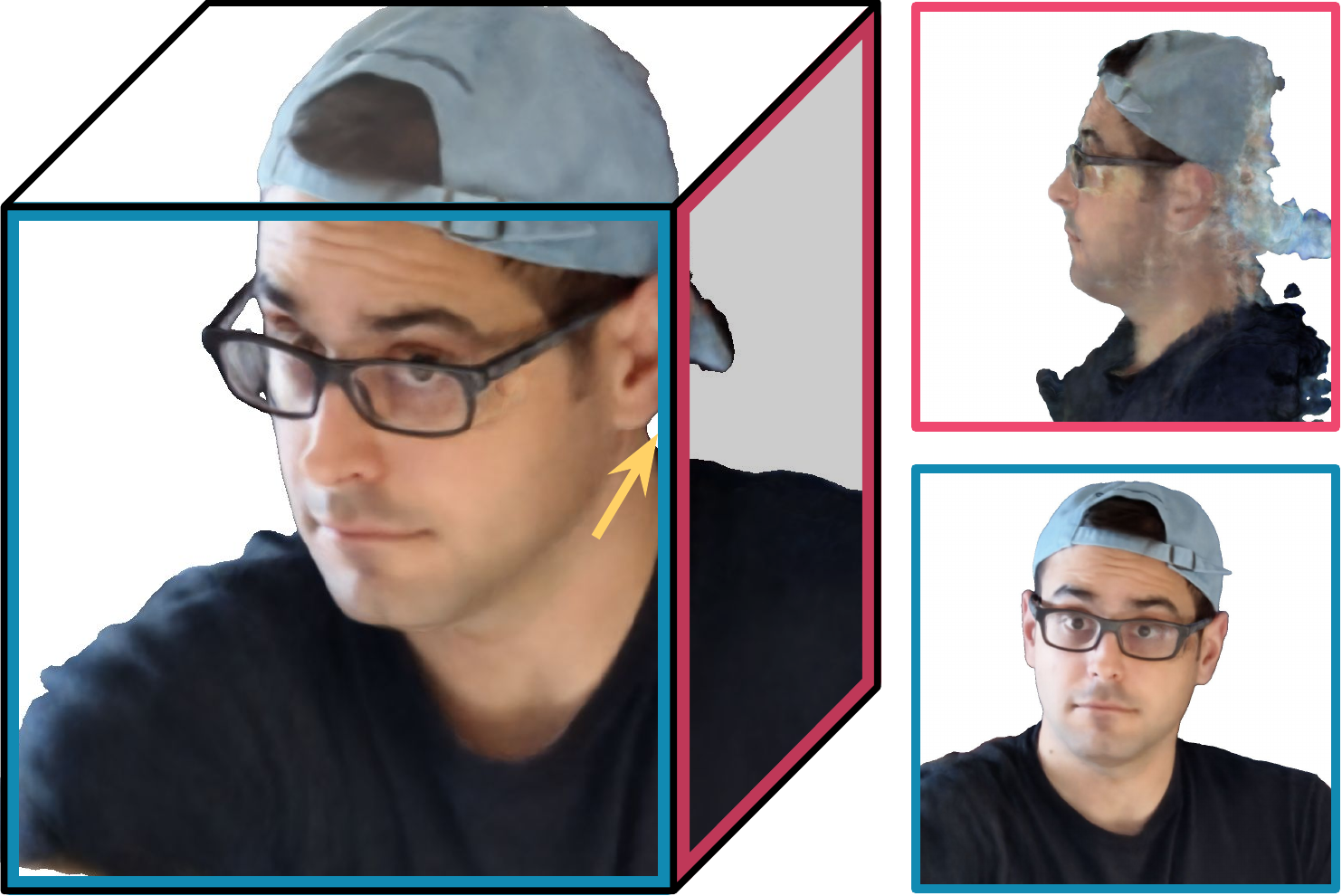}
		\caption*{observation frame}
	\end{subfigure}
	\begin{subfigure}{0.49\columnwidth}
		\centering
    	\includegraphics[width=\textwidth]{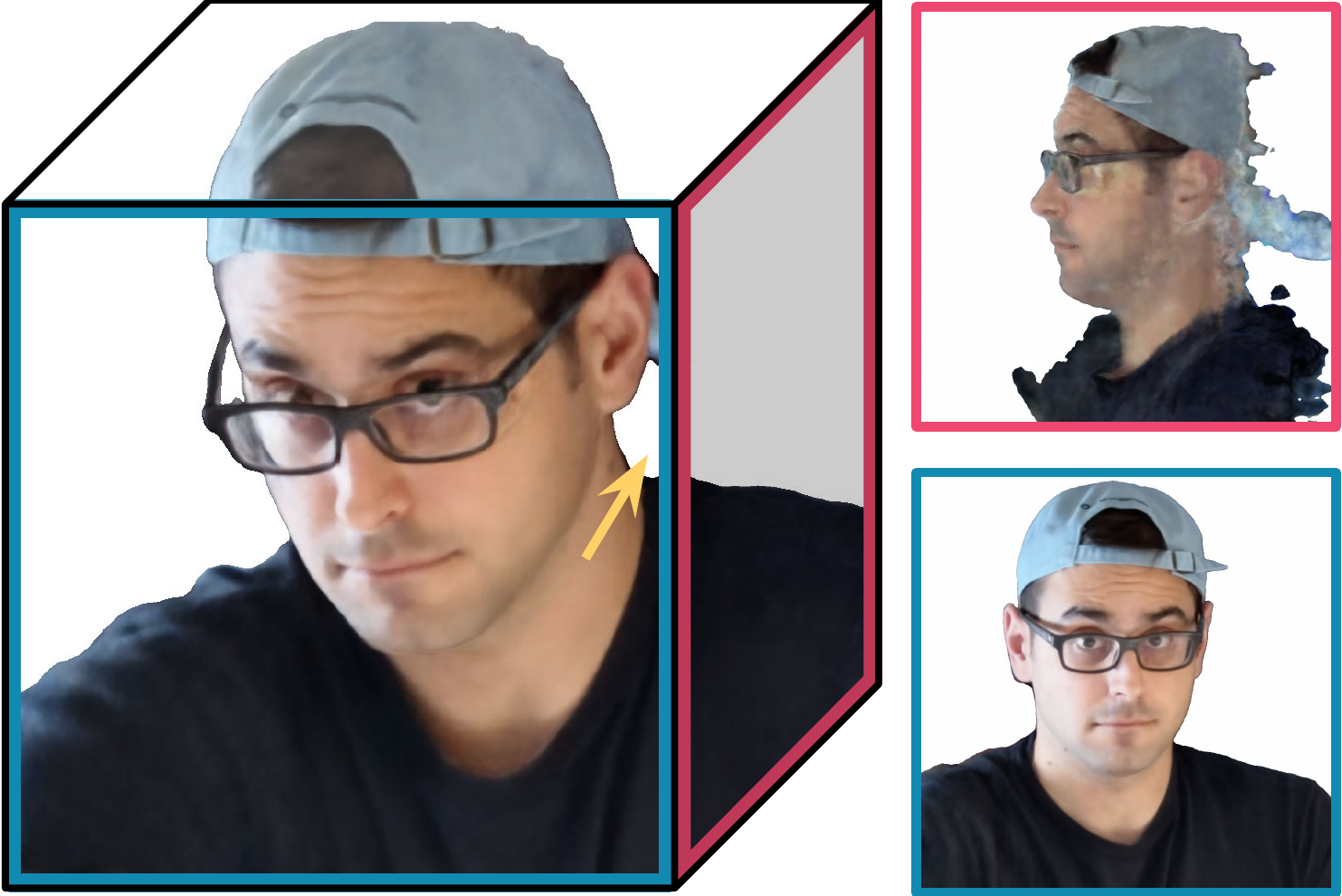}
		\caption*{canonical frame}
	\end{subfigure}
	\vspace{-8pt}
	\caption{
    Visualizations of the recovered 3D model in the observation and canonical frames of reference, with insets showing orthographic views in the forward and left directions. 
    Note the right-to-left and front-to-back displacements between the observation and canonical model, which are modeled by the deformation field for this observation.
	}
   \label{fig:observation_vs_canonical}
\end{figure}

%% file: f0x_elastic_example.tex
\fboxsep=0pt 
\fboxrule=0.4pt 

\begin{figure}[t]
	\centering
	\begin{subfigure}{1.0\columnwidth}
    	\centering
    	\figcelltb{0.24}{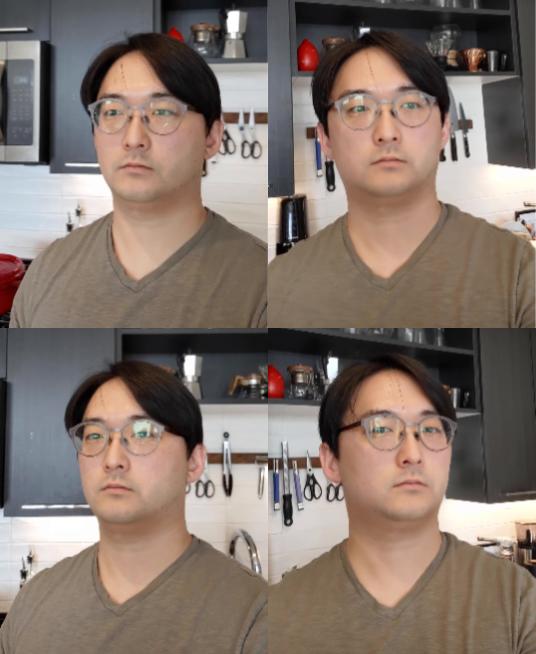}{example inputs}{clip,trim=0 10 0 10}\hspace{-1.0pt}
    	\figcelltb{0.24}{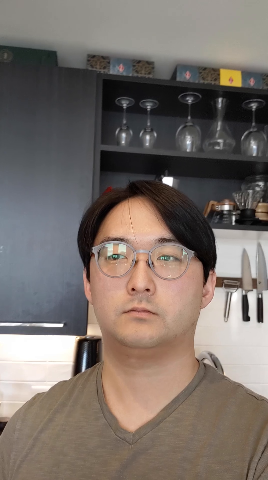}{ground truth}{clip,trim=40 60 10 160}\hspace{-1.0pt}
    	\figcelltb{0.24}{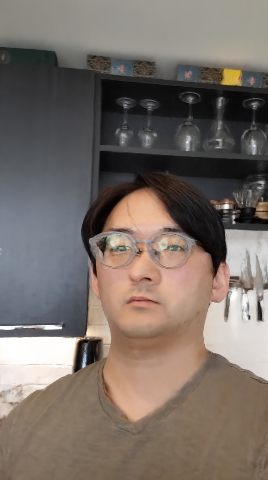}{elastic off}{clip,trim=40 60 10 160}\hspace{-1.0pt}
    	\figcelltb{0.24}{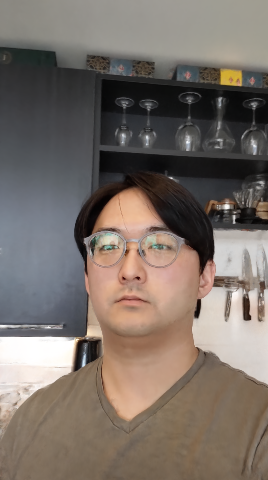}{elastic on}{clip,trim=40 60 10 160}
	\end{subfigure}
	\vspace{-8pt}
	\caption{Our elastic regularization helps when the scene is under-constrained. This capture only contains 20 input images with the cameras biased towards one side of the face resulting in an under constrained problem. Elastic regularization helps resolve the ambiguity and leads to less distortion.}
   \label{fig:elastic_example}
   
   \vspace{-12pt}
\end{figure}

%% file: f0x_coarse_to_fine_optimization.tex
\fboxsep=0pt 
\fboxrule=0.4pt 

\begin{figure}[t]
	\begin{subfigure}{0.96\columnwidth}
    	\centering
    	\figcelltb{0.24}{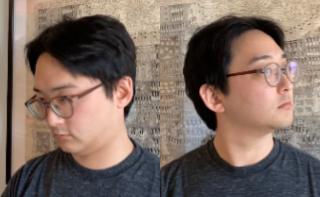}{head turn}{}
    	\figcelltb{0.17}{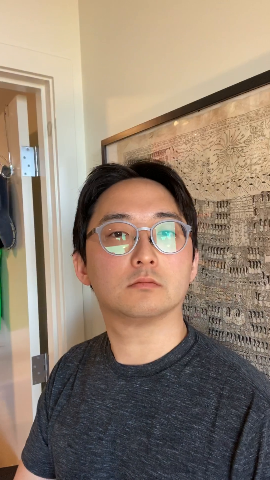}{}{clip,trim=60 120 50 165}
    	\figcelltb{0.17}{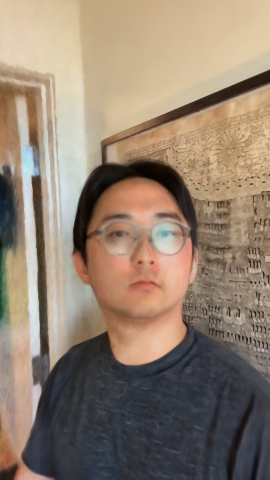}{}{clip,trim=60 120 50 165}
    	\figcelltb{0.17}{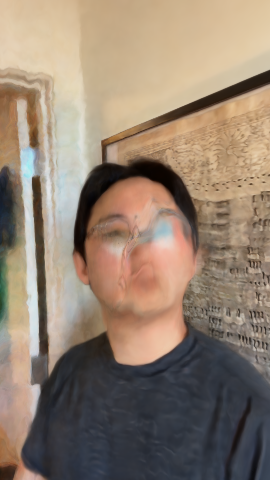}{}{clip,trim=60 120 50 165}
    	\figcelltb{0.17}{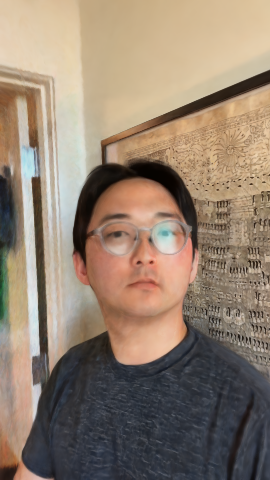}{}{clip,trim=60 120 50 165}
	\end{subfigure}
	\begin{subfigure}{0.96\columnwidth}
    	\centering
    	\figcelltb{0.24}{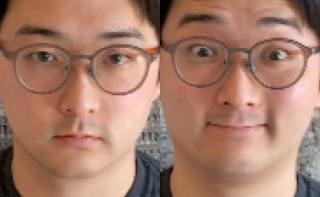}{\makecell{smile\\\xspace}}{}
    	\figcelltb{0.17}{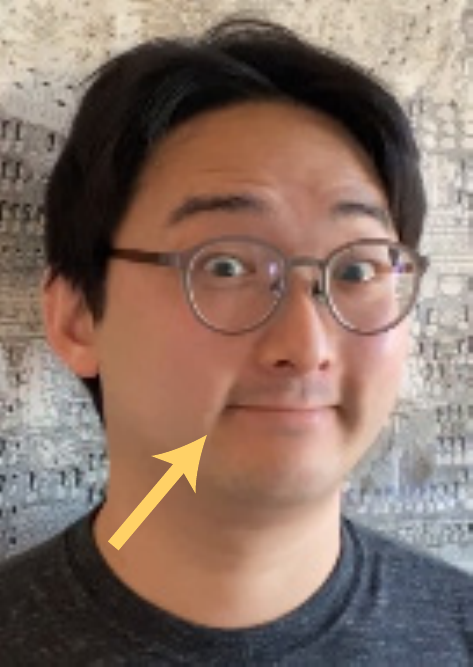}{gt}{clip,trim=0 10 0 20}
    	\figcelltb{0.17}{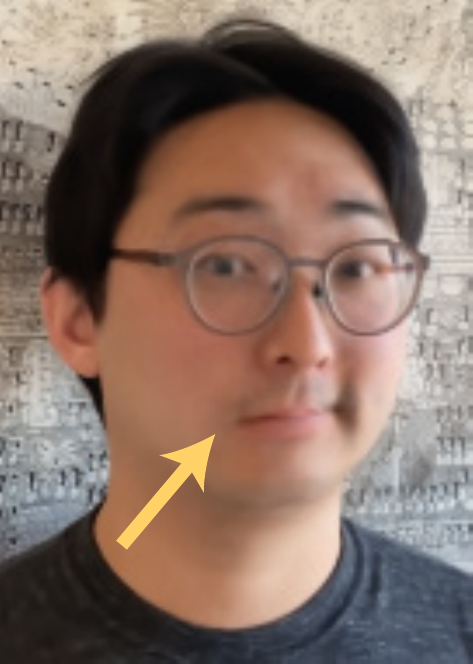}{$m=4$}{clip,trim=0 10 0 20}
    	\figcelltb{0.17}{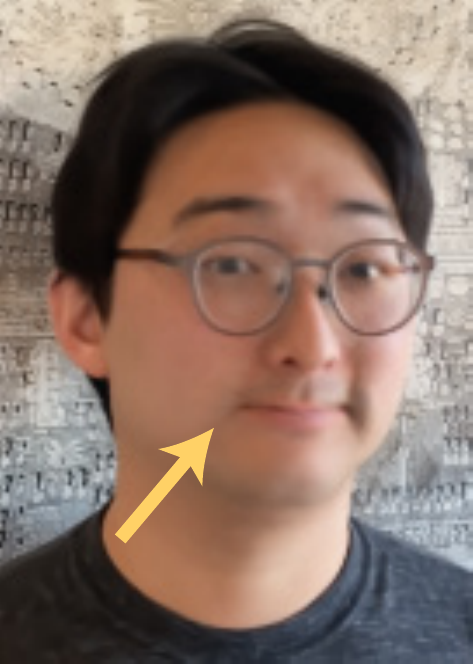}{$m=8$}{clip,trim=0 10 0 20}
    	\figcelltb{0.17}{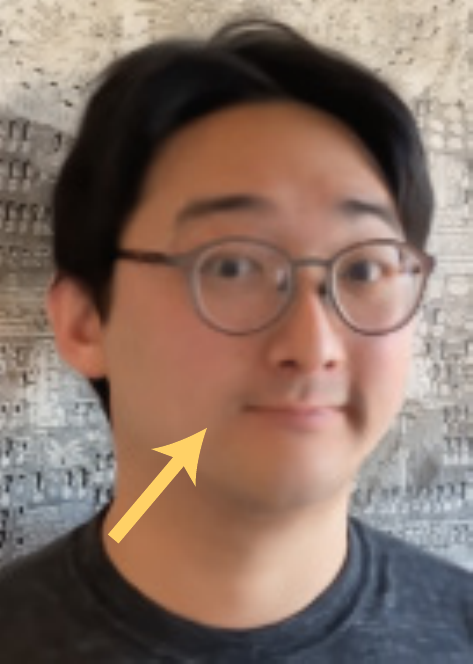}{c2f}{clip,trim=0 10 0 20}
	\end{subfigure}
	\vspace{-5pt}
	\caption{
In this capture, the subject rotates their head (top) and smiles (bottom). With $m=4$ positional encoding frequencies, the deformation model does not capture the smile, while it fails to rotate the head with $m=8$ frequencies. With coarse-to-fine regularization (c2f) the model captures both.
}
   \label{fig:coarse_to_fine}
   \vspace{-10pt}

\end{figure}

%% file: f0x_latent_interpolation_v2.tex

\fboxsep=0pt 
\fboxrule=1pt 

\definecolor{startcolor}{rgb}{0.067, 0.541, 0.698}
\definecolor{endcolor}{rgb}{0.937, 0.278, 0.435}

\newcommand{\kfbox}[1]{\fcolorbox{black}{white}{#1}}
\newcommand{\kfboxstart}[1]{\fcolorbox{startcolor}{white}{#1}}
\newcommand{\kfboxend}[1]{\fcolorbox{endcolor}{white}{#1}}
\newcommand{\latcell}[1]{\includegraphics[width=1.2cm,clip,trim=30 50 0 0]{#1}}
\newcommand{\latcellkfstart}[1]{\kfboxstart{\includegraphics[width=1.2cm,clip,trim=30 50 0 0]{#1}}}
\newcommand{\latcellkfend}[1]{\kfboxend{\includegraphics[width=1.2cm,clip,trim=30 50 0 0]{#1}}}

\begin{figure}[t]
    \setlength{\tabcolsep}{0.5pt}
    \renewcommand{\arraystretch}{0.25}
    \centering
    \begin{tabular}{ccccccc}
    
        \multirow{4}*[1.2cm]{\kfboxstart{\includegraphics[width=0.93cm,clip,trim=40 0 40 0]{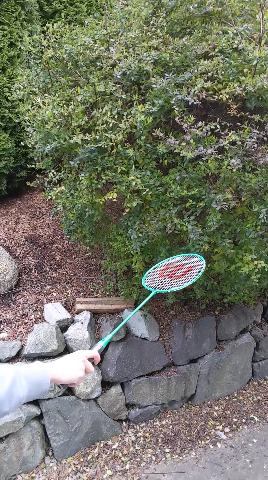}}\hspace{0.05cm}} &
        
        \latcellkfstart{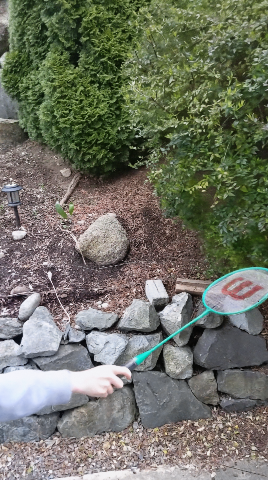} &
        \latcell{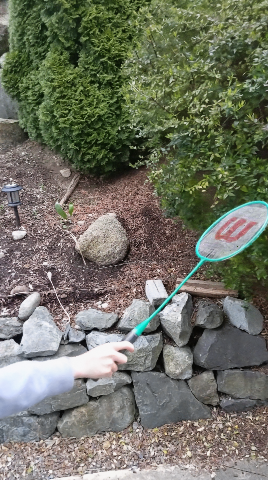} &
        \latcell{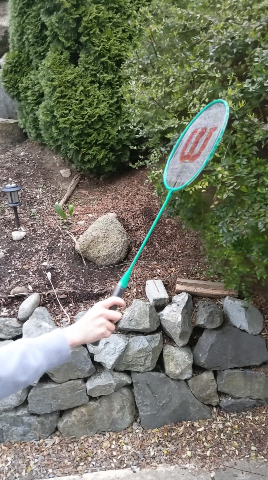} &
        \latcell{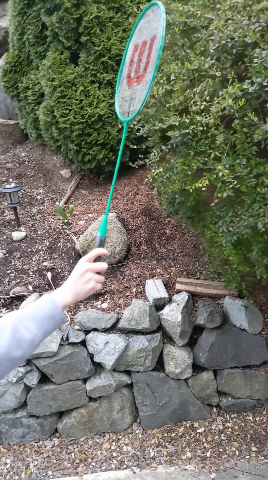} &
        \latcellkfend{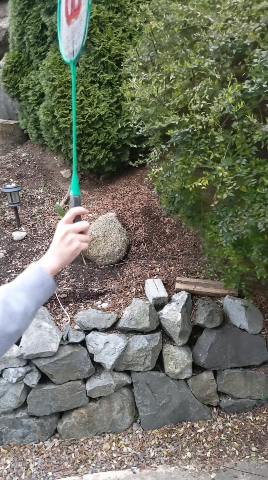} &
        
        \multirow{4}*[1.2cm]{\hspace{0.05cm}\kfboxend{\includegraphics[width=0.93cm,clip,trim=40 0 40 0]{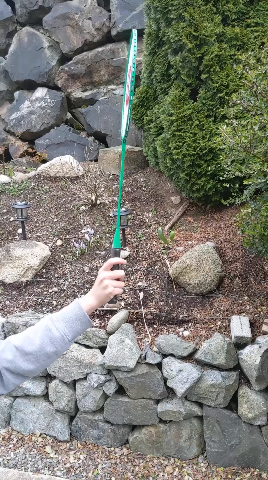}}}
        \vspace{2pt}
        \\
        
        & 
        \multicolumn{5}{c}{} & \vspace{-4pt}

        \\
       
        {\small start}
        &
        \latcellkfstart{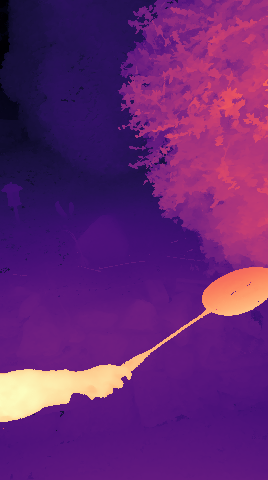} &
        \latcell{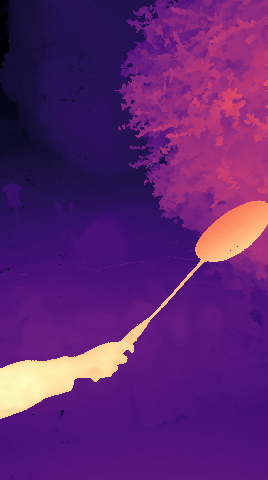} &
        \latcell{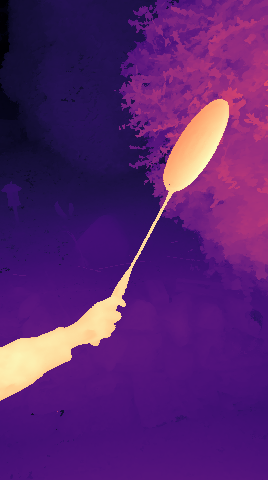} &
        \latcell{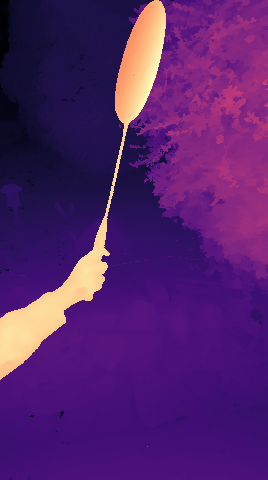} &
        \latcellkfend{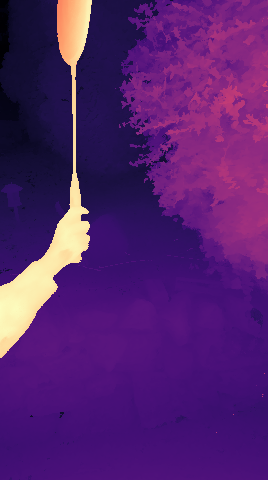} & 
        
        {\small end}
        \vspace{2pt}
        \\
        & \multicolumn{5}{c}{{\small interpolated frames from a novel view}} & 
        
    \end{tabular}
    \vspace{-8pt}
	\caption{
	    Novel views synthesized by linearly interpolating the deformation latent codes of two frames of the \textsc{Badminton} (left and right) show a smooth racquet motion.
	}
  \label{fig:latent_interpolation}
  \vspace{-12pt}
\end{figure}

%% file: f0x_hair.tex
\fboxsep=0pt 
\fboxrule=0.4pt 

\begin{figure}[t]
	\centering
	\begin{subfigure}{1.0\columnwidth}
    	\centering
    	\figcelltb{0.32}{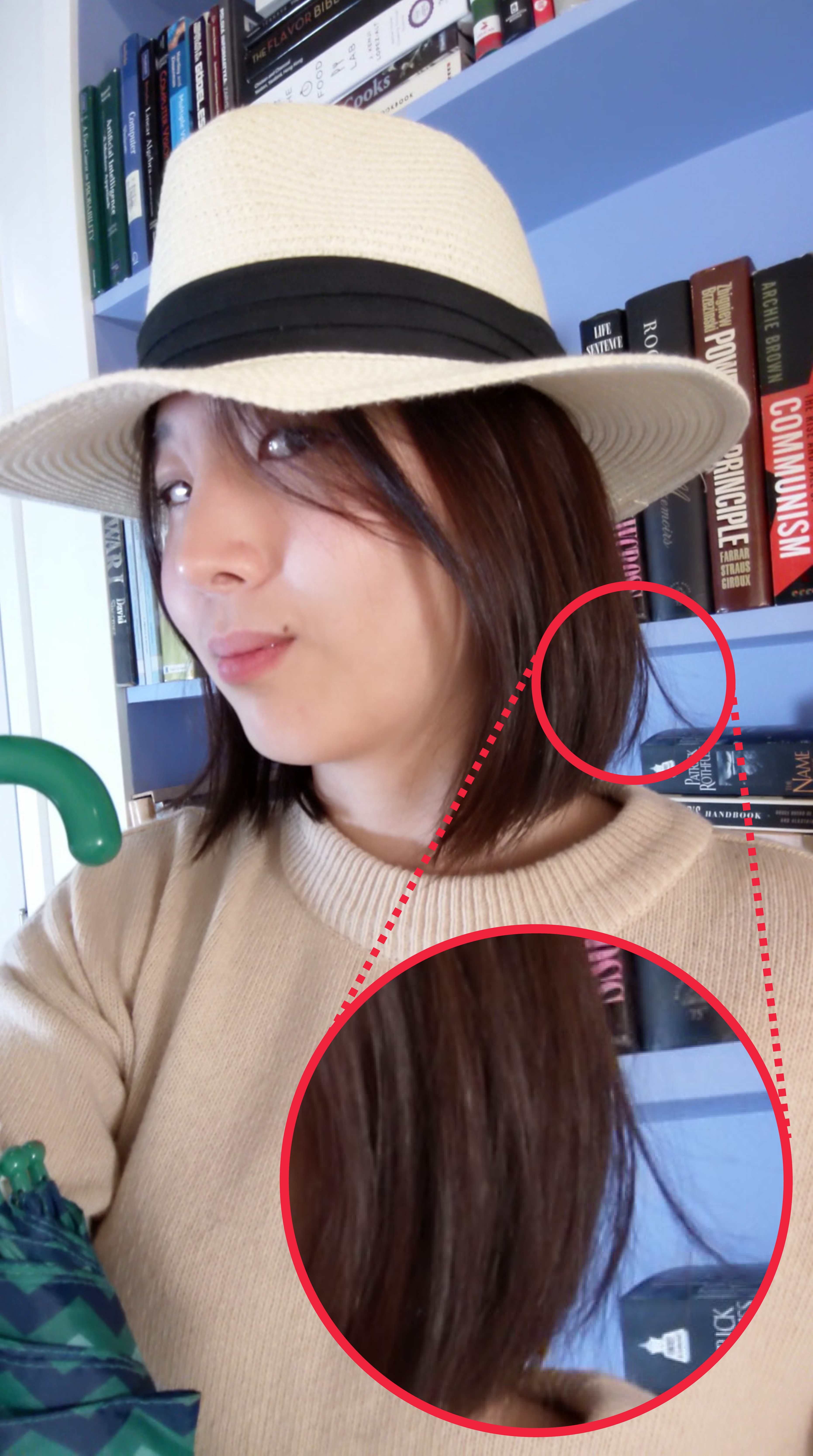}{ground truth}{clip,trim=0 30 0 430}\hspace{-1.0pt}
    	\figcelltb{0.32}{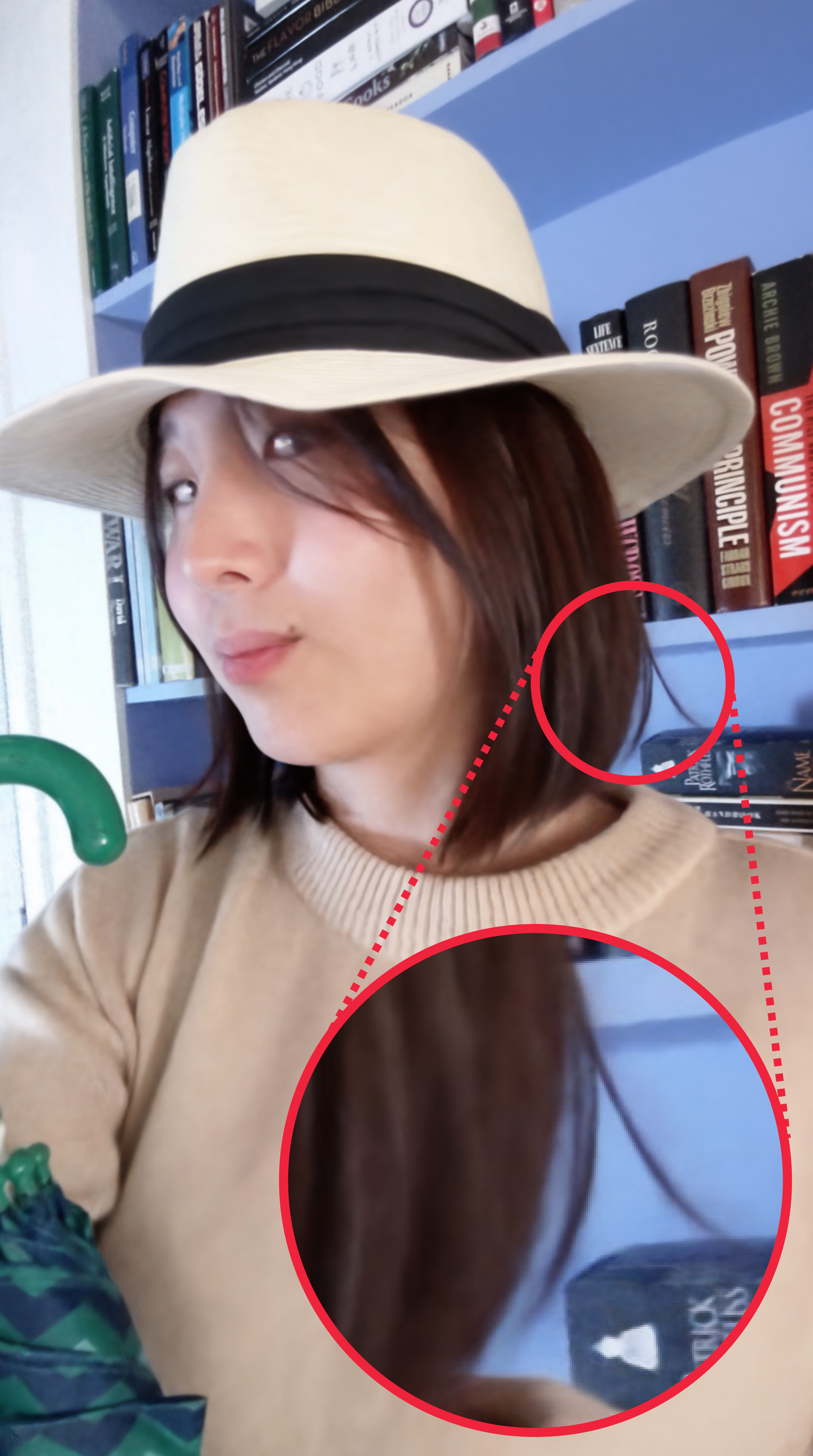}{rendered color}{clip,trim=0 30 0 430}\hspace{-1.0pt}
    	\figcelltb{0.32}{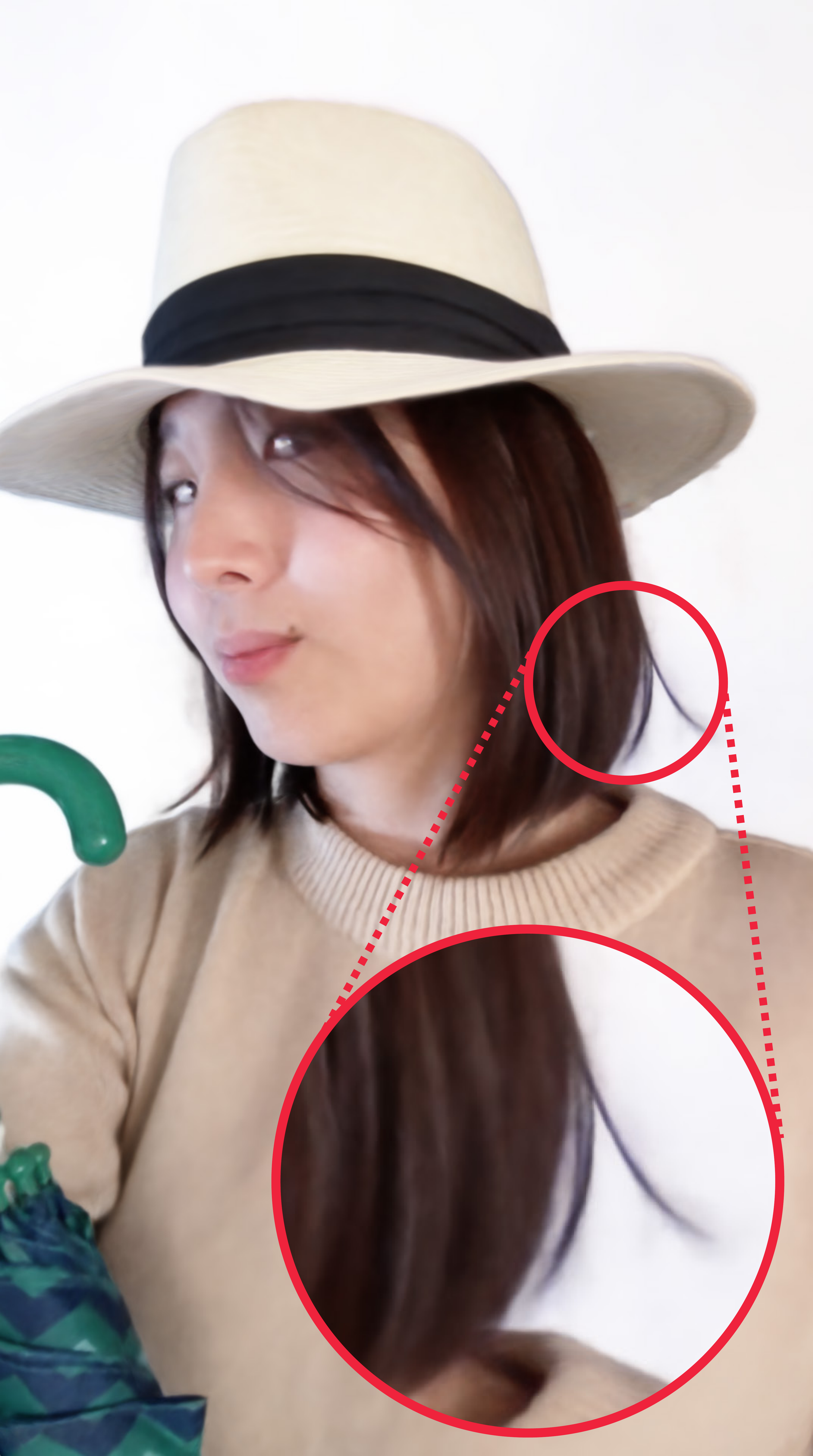}{render w/o bg}{clip,trim=0 30 0 430}
	\end{subfigure}
	\vspace{-8pt}
	\caption{Our method recovers thin hair strands.
	By adjusting the camera's far plane, we can render the subject against a flat white background.}
   \label{fig:qualitative_hair}
   \vspace{-15pt}
\end{figure}

%% file: 05_evaluation.tex
\section{Experiments}
\input{f0x_fullbody}

\subsection{Implementation Details}
Our NeRF template implementation closely follows the original~\cite{mildenhall2020nerf}, except we use a Softplus activation $\ln(1+e^x)$ for the density. We use a deformation network with depth 6, hidden size 128, and a skip connection at the 4th layer. We use 256 coarse and fine ray samples for full HD ($1920{\times}1080$) models and half that for the half resolution models. We use 8 dimensions for the latent deformation and appearance codes. For coarse-to-fine optimization we use 6 frequency bands and linearly anneal $\alpha$ from $0$ to $6$ over 80K iterations. We use the same MSE photometric loss as in NeRF~\cite{mildenhall2020nerf} and weight the losses as $L_\text{total} = L_\text{rgb} + \lambda L_\text{elastic-r} + \mu L_\text{bg}$
where we use $\lambda=\mu=10^{-3}$ for all experiments except when mentioned.
We train on 8 V100 GPUs for a week for full HD models, and for 16 hours for the half resolution models used for the comparisons in \tabref{table:results}, \figref{fig:additional-vrig-results-static}, and \figref{fig:additional-vrig-results-dynamic}. We provide more details in the Section A of the appendix.

\begin{figure}[t]
    \includegraphics[width=\columnwidth]{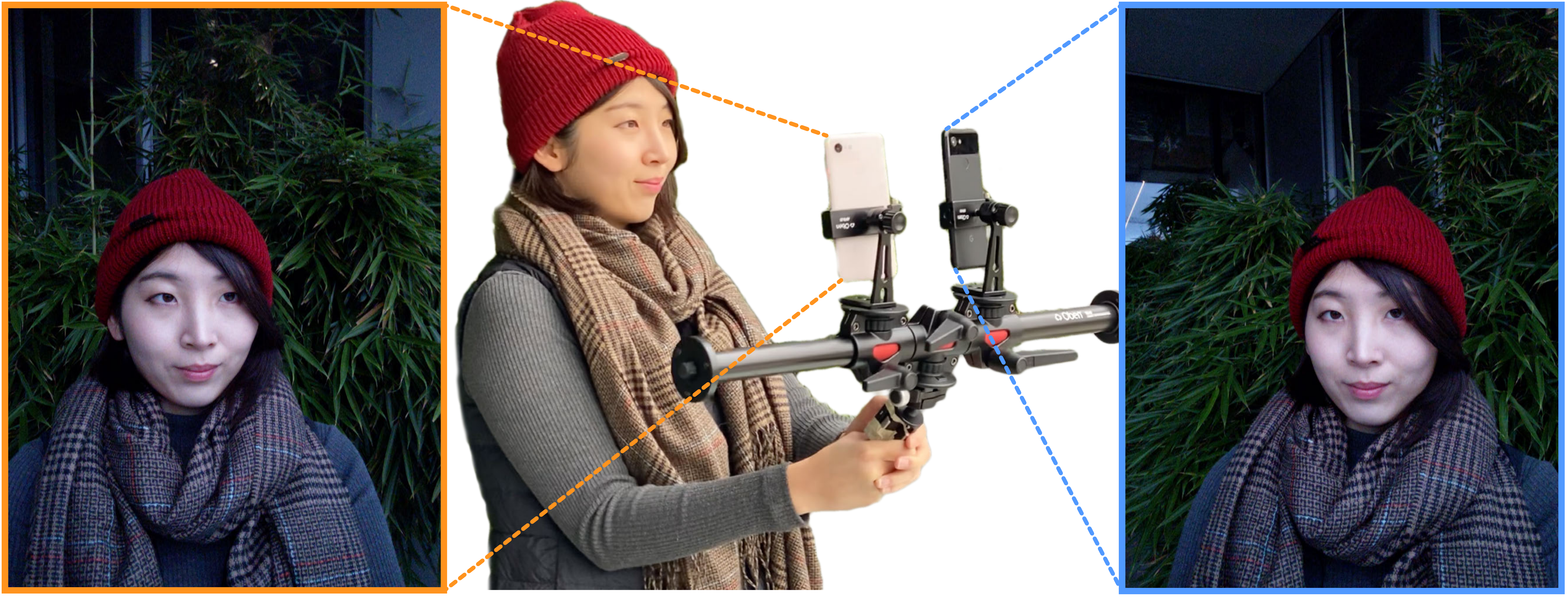}
    \vspace{-18pt}
	\caption{
	Validation rig used only for evaluation.
	}
  \label{fig:validation-rig}
  \vspace{-12pt}
\end{figure}

\subsection{Evaluation Dataset} In order to evaluate the quality of our reconstruction, we must be able to measure how faithfully we can recreate the scene from a viewpoint unseen during training. Since we are reconstructing non-rigidly deforming scenes, we cannot simply hold out views from an input capture, as the structure of the scene will be slightly different in every image. We therefore build a simple multi-view data capture rig for the sole purpose of evaluation. 
We found the multi-view dataset of Yoon~\etal~\cite{yoon2020novel} not representative of many capture scenarios, as it contains too few viewpoints (12) and exaggerated frame-to-frame motions due to temporal subsampling. 

Our rig (\figref{fig:validation-rig}) is a pole with two Pixel 3's rigidly attached. We have two methods for data capture: (a) for selfies we use the front-facing camera and capture time-synchronized photos using the method of Ansari~\etal~\cite{ansari2019wireless}, which achieves sub millisecond synchronization; or (b) we use the back-facing camera and record two videos which we manually synchronize based on the audio; we then subsample to 5 fps. We register the images using COLMAP~\cite{schoenberger2016sfm} with rigid relative camera pose constraints. Sequences captured with (a) contain fewer frames (40\textasciitilde78) but the focus, exposure, and time are precisely synchronized. Sequences captured with (b) have denser samples in time (between 193 and 356 frames) but the synchronization is less precise and exposure and focus may vary between the cameras.
We split each capture into a training set and a validation set.  
We alternate assigning the left view to the training set, and right to the validation, and vice versa. This avoids having regions of the scene that one camera has not seen.

\mypar{Quasi-static scenes} We capture 5 human subjects
using method (a), that attempt to stay as still as possible during capture, and a mostly still dog using method (b). 
\input{t0x_quantative_table}
\input{fa0_vrig_results_dynamic}
\input{fa0_vrig_results_static}

\mypar{Dynamic scenes} We capture 4 dynamic scenes containing deliberate motions of a human subject, a dog wagging its tail, and two moving objects
using method (b).

\subsection{Evaluation}
\label{sec:evaluation}

Here we provide quantitative and qualitative evaluations of our model. However, to best appreciate the quality of the reconstructed \emph{nerfies}, we encourage the reader to watch the supplementary video that contains many example results.

\mypar{Quantitative Evaluation}
We compare against NeRF and a NeRF + latent baseline, where NeRF is conditioned on a per-image learned latent code~\cite{bojanowski2018optimizing} to modulate density and color. We also compare with a variant of our system similar to the concurrent work of D-NeRF~\cite{pumarola2020dnerf}, which conditions a translational deformation field with a position encoded time variable $\gamma(t)$ instead of a latent code ({\footnotesize$\gamma(t)$+trans} in \tabref{table:results}).
We also compare with the high quality model of Neural Volumes (NV)~\cite{lombardi2019neural} using a single view as input to the encoder, and Neural Scene Flow Fields (NSFF)~\cite{li2020neural}. We do not evaluate the method of Yoon et al.~\cite{yoon2020novel} due to the lack of available code (note that NSFF outperforms it). 
NSFF and the {\footnotesize$\gamma(t)+trans$} baseline use temporal information while other baselines and our method do not. 
NSFF also uses auxilliary supervision such as estimated flow and relative depth maps; we do not. Note that the default hyper-parameters for NSFF~\cite{li2020nsff} provided with the official code performs poorly on our datasets --- we therefore contacted the authors to help us tune the hyper-parameters.
Photometric differences between the two rig cameras may exist due to different exposure/white balance settings and camera response curves. We therefore swap the per-frame appearance code $\psi_i$ for a per-camera $\{\psi_L, \psi_R\}\in\mathbb{R}^2$ instead for validation rig captures.

\tabref{table:results} reports LPIPS~\cite{zhang2018unreasonable} and PSNR metrics for the unseen validation views. PSNR favors blurry images and is therefore not an ideal metric for dynamic scene reconstruction; we find that LPIPS is more representative of visual quality. See \figref{fig:additional-vrig-results-static} and \figref{fig:additional-vrig-results-dynamic} for side-by-side images with associated PSNR/LPIPS metrics. Our method struggles with PSNR due to slight misalignments resulting from factors such as gauge ambiguity~\cite{mclauchlan1999gauge} while we outperform all baselines in terms of LPIPS for all sequences. 

\mypar{Ablation Study}
We evaluate each of our contributions: SE(3) deformations, elastic regularization, background regularization, and coarse-to-fine optimization. We ablate them one at a time, and all at once ({\footnotesize Ours (bare)} in \tabref{table:results}).
As expected, a stronger elastic regularization ({\footnotesize$\lambda=0.01$}) improves results for dynamic scenes compared to the baseline ({\footnotesize$\lambda=0.001$}) while minimally impacting quasi-static scenes. 
Removing the elastic loss hurts performance for quasi-static scenes while having minimal effect on the dynamic scene; this may be due to the larger influence of other losses in the presence of larger motion. Elastic regularization fixes distortion artifacts when the scene is under-constrained (e.g., \figref{fig:elastic_example}).
Disabling coarse-to-fine regularization mildly drops performance for quasi-static scenes while causing a significant drop for dynamic scenes. This is expected since large motions are a main source of local minima (e.g., \figref{fig:coarse_to_fine}). 
Our SE(3) deformations also quantitatively outperform translational deformations. 
Background regularization helps PSNR by reducing shifts in static regions and removing it performs worse.
Finally, removing all of our contributions performs the worst in terms of LPIPS.

\mypar{Qualitative Results}
We show results for the captures used in the quantitative evaluation in \figref{fig:additional-vrig-results-static} and \figref{fig:additional-vrig-results-dynamic}. Our method can reconstruct fine details such as strands of hair (e.g., in \textsc{\small{Curls}} of \tabref{table:results} and \figref{fig:qualitative_hair}), shirt wrinkles, and glasses (\figref{fig:qualitative_fullbody}).
Our method works on general scenes beyond human subjects as shown in \figref{fig:additional-vrig-results-static} and \figref{fig:additional-vrig-results-dynamic}.
In addition, we can create smooth animations by interpolating the deformation latent codes of any input state as shown in \figref{fig:latent_interpolation}.

\mypar{Elastic Regularization} \figref{fig:elastic_example} shows an example where the user only captured 20 images mostly from one side of their face, while their head tracked the camera. This results in ambiguous geometry. Elastic regularization helps in such under-constrained cases, reducing distortion significantly.

\input{f0x_failure_cases}

\mypar{Depth Visualizations}
We visualize the quality of our reconstruction using depth renders of the density field. Unlike NeRF\cite{mildenhall2020nerf} that visualizes the expected ray termination distance, we use the \emph{median} depth termination distance, which we found to be less biased by residual density in free space (see Fig.~\ref{fig:qualitative_hair}). We define it as the depth of the first sample with accumulated transmittance $T_i \geq 0.5$ (Eqn.~3 of NeRF~\cite{mildenhall2020nerf}).

\mypar{Limitations}
Our method struggles with topological changes e.g., opening/closing of the mouth (see \figref{fig:failure_cases}) and may fail for certain frames in the presence of rapid motion (see supplementary).
As mentioned in \secref{sec:background-regularization}, our deformations are unconstrained so static regions may shift; this contributes to the disjunction between PSNR and LPIPS in \tabref{table:results}. Future work may address this by modeling static regions separately as in \cite{martinbrualla2020nerfw,li2020neural}.
Finally, the quality of our method depends on camera registration, and when SfM fails so do we.

%% file: f0x_fullbody.tex
\fboxsep=0pt 
\fboxrule=0.4pt 

\begin{figure}[t]
	\centering
	\begin{subfigure}{1.0\columnwidth}
    	\centering
    	\figcellb{0.33}{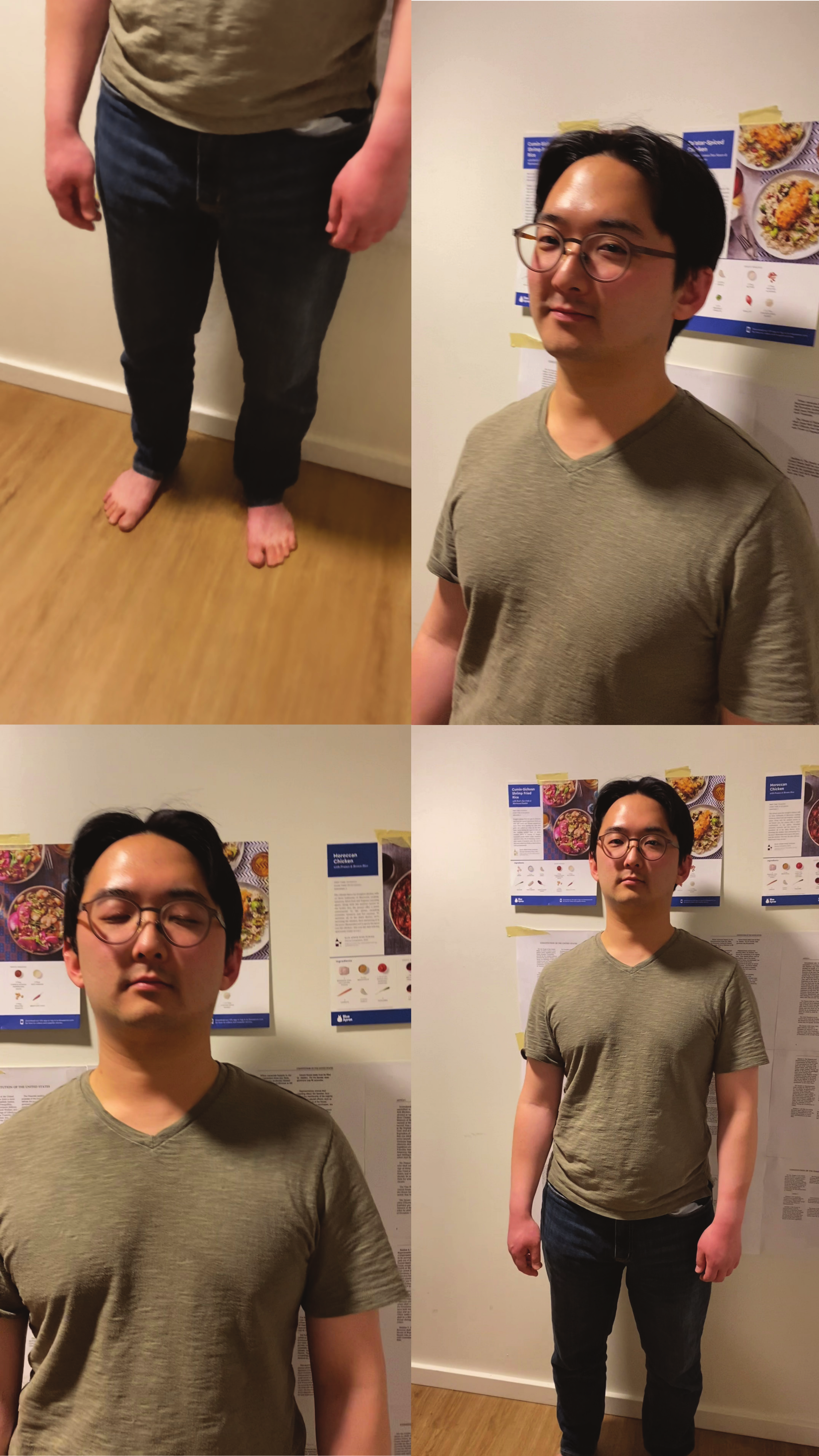}{example inputs}\hspace{-1.0pt}
    	\figcellb{0.3265}{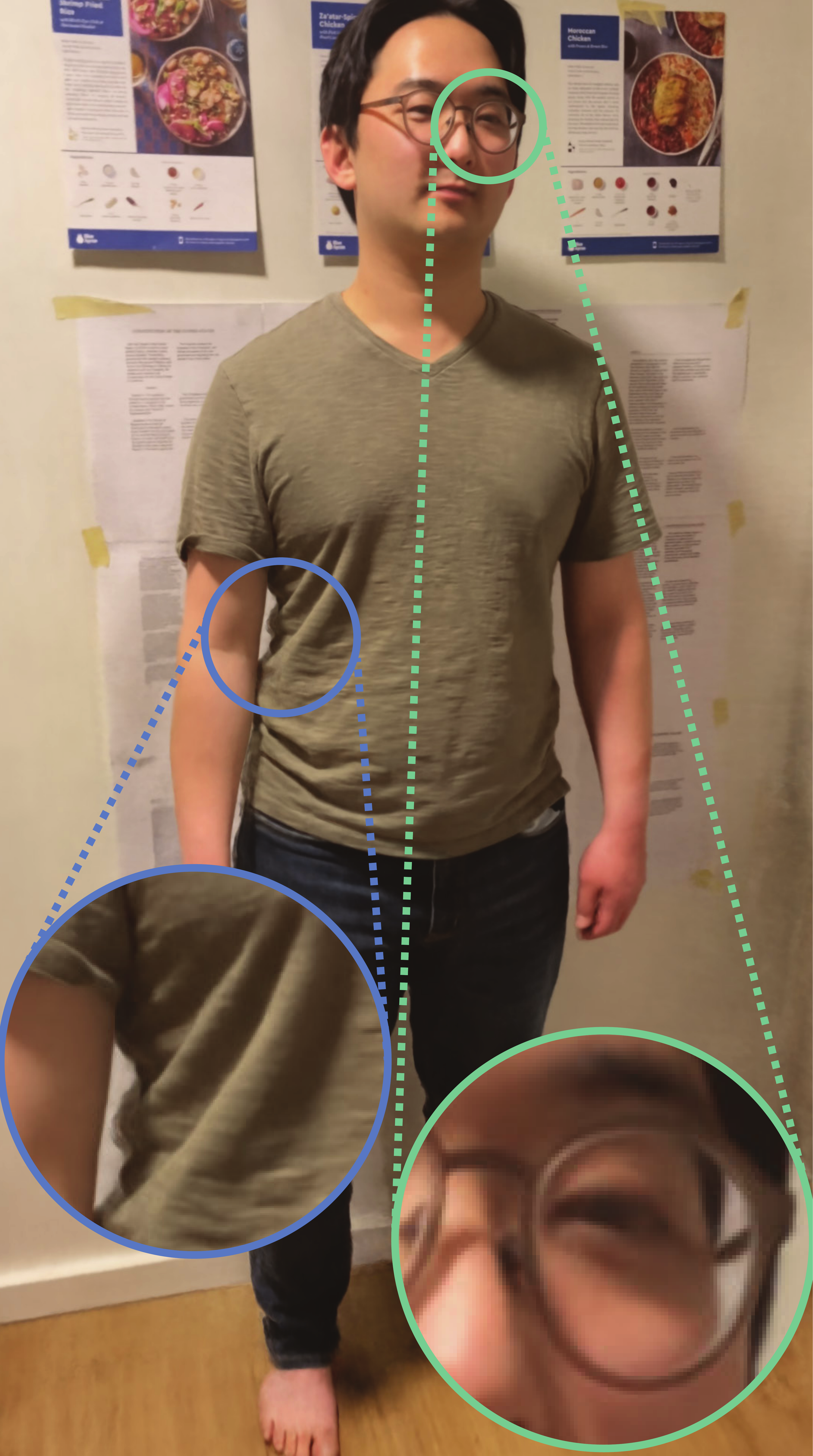}{rendered rgb}\hspace{-1.0pt}
    	\figcellb{0.33}{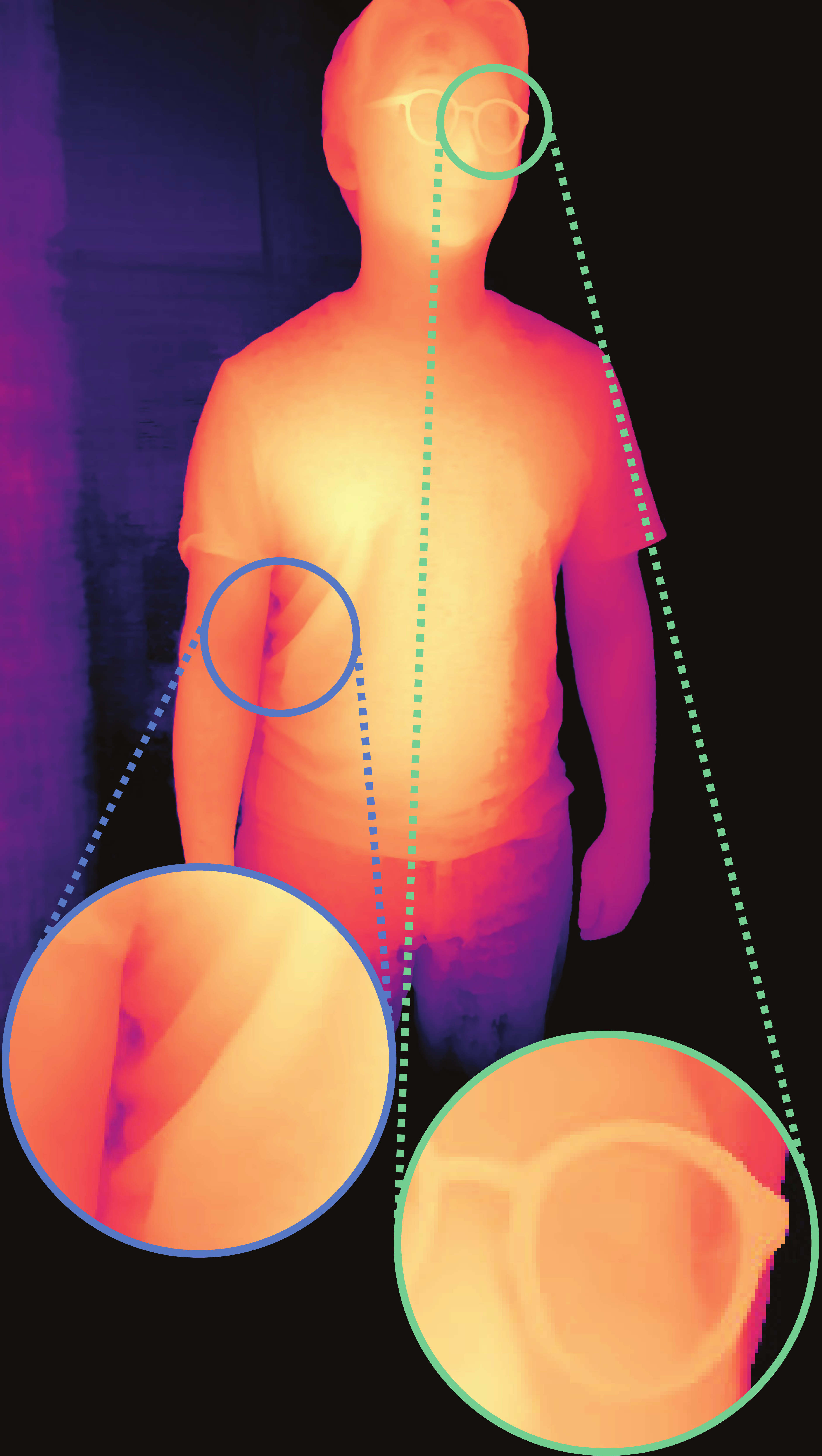}{rendered depth}
	\end{subfigure}
	\vspace{-8pt}
	\caption{Our method reconstructs full body scenes captured by a second user with high quality details. }
  \label{fig:qualitative_fullbody}
  
   \vspace{-8pt}
\end{figure}

%% file: t0x_quantative_table.tex
\newcommand{\tablefirst}[0]{\cellcolor{myred}}
\newcommand{\tablesecond}[0]{\cellcolor{myorange}}
\newcommand{\tablethird}[0]{\cellcolor{myyellow}}

\begin{table*}[t]
\caption{
Quantitative evaluation on validation captures against baselines and ablations of our system, we color code each row as \colorbox{myred}{\textbf{best}}, \colorbox{myorange}{\textbf{second best}}, and \colorbox{myyellow}{\textbf{third best}}. \textsuperscript{\textdagger}denotes use of temporal information. Please see Sec.~\ref{sec:evaluation} for more details.
\label{table:results}
}
\centering
\resizebox{\linewidth}{!}{
\setlength{\tabcolsep}{2pt}
\input{t0x_quantitative_table_contents_0909-crop}
}
\end{table*}

%% file: t0x_quantitative_table_contents_0909-crop.tex

\begin{tabular}{l||cc|cc|cc|cc|cc|cc|cc||cc|cc|cc|cc|cc}

\toprule

& \multicolumn{ 14 }{c||}{\makecell{{\small Quasi-Static }}}
& \multicolumn{ 10 }{c}{\makecell{{\small Dynamic }}}
\\ \hline

& \multicolumn{ 2 }{c}{
  \makecell{
  \textsc{\small Glasses }
\\(78 images)
  }
}
& \multicolumn{ 2 }{c}{
  \makecell{
  \textsc{\small Beanie }
\\(74 images)
  }
}
& \multicolumn{ 2 }{c}{
  \makecell{
  \textsc{\small Curls }
\\(57 images)
  }
}
& \multicolumn{ 2 }{c}{
  \makecell{
  \textsc{\small Kitchen }
\\(40 images)
  }
}
& \multicolumn{ 2 }{c}{
  \makecell{
  \textsc{\small Lamp }
\\(55 images)
  }
}
& \multicolumn{ 2 }{c|}{
  \makecell{
  \textsc{\small Toby Sit }
\\(308 images)
  }
}
& \multicolumn{ 2 }{c||}{
  \makecell{
  \textsc{\small Mean }
  }
}
& \multicolumn{ 2 }{c}{
  \makecell{
  \textsc{\small Drinking }
\\(193 images)
  }
}
& \multicolumn{ 2 }{c}{
  \makecell{
  \textsc{\small Tail }
\\(238 images)
  }
}
& \multicolumn{ 2 }{c}{
  \makecell{
  \textsc{\small Badminton }
\\(356 images)
  }
}
& \multicolumn{ 2 }{c|}{
  \makecell{
  \textsc{\small Broom }
\\(197 images)
  }
}
& \multicolumn{ 2 }{c}{
  \makecell{
  \textsc{\small Mean }
  }
}
\\

& \multicolumn{1}{c}{ \footnotesize PSNR$\uparrow$ }
& \multicolumn{1}{c}{ \footnotesize LPIPS$\downarrow$ }
& \multicolumn{1}{c}{ \footnotesize PSNR$\uparrow$ }
& \multicolumn{1}{c}{ \footnotesize LPIPS$\downarrow$ }
& \multicolumn{1}{c}{ \footnotesize PSNR$\uparrow$ }
& \multicolumn{1}{c}{ \footnotesize LPIPS$\downarrow$ }
& \multicolumn{1}{c}{ \footnotesize PSNR$\uparrow$ }
& \multicolumn{1}{c}{ \footnotesize LPIPS$\downarrow$ }
& \multicolumn{1}{c}{ \footnotesize PSNR$\uparrow$ }
& \multicolumn{1}{c}{ \footnotesize LPIPS$\downarrow$ }
& \multicolumn{1}{c}{ \footnotesize PSNR$\uparrow$ }
& \multicolumn{1}{c|}{ \footnotesize LPIPS$\downarrow$ }
& \multicolumn{1}{c}{ \footnotesize PSNR$\uparrow$ }
& \multicolumn{1}{c||}{ \footnotesize LPIPS$\downarrow$ }
& \multicolumn{1}{c}{ \footnotesize PSNR$\uparrow$ }
& \multicolumn{1}{c}{ \footnotesize LPIPS$\downarrow$ }
& \multicolumn{1}{c}{ \footnotesize PSNR$\uparrow$ }
& \multicolumn{1}{c}{ \footnotesize LPIPS$\downarrow$ }
& \multicolumn{1}{c}{ \footnotesize PSNR$\uparrow$ }
& \multicolumn{1}{c}{ \footnotesize LPIPS$\downarrow$ }
& \multicolumn{1}{c}{ \footnotesize PSNR$\uparrow$ }
& \multicolumn{1}{c|}{ \footnotesize LPIPS$\downarrow$ }
& \multicolumn{1}{c}{ \footnotesize PSNR$\uparrow$ }
& \multicolumn{1}{c}{ \footnotesize LPIPS$\downarrow$ }
\\
\hline

  NeRF~\cite{mildenhall2020nerf}
  &$18.1$
  &$.474$
  
  &$16.8$
  &$.583$
  
  &$14.4$
  &$.616$
  
  &$19.1$
  &$.434$
  
  &$17.4$
  &$.444$
  
  &$22.8$
  &$.463$
  
  &$18.1$
  &$.502$
  
  &$18.6$
  &$.397$
  
  &$23.0$
  &$.571$
  
  &$18.8$
  &$.392$
  
  &$21.0$
  &$.667$
  
  &$20.3$
  &$.506$
  
  \\
  NeRF + latent
  &$19.5$
  &$.463$
  
  &$19.5$
  &$.535$
  
  &$17.3$
  &$.539$
  
  &$20.1$
  &$.403$
  
  &$18.9$
  &$.386$
  
  &$19.4$
  &$.385$
  
  &$19.1$
  &$.452$
  
  &$21.9$
  &$.233$
  
  &\tablethird$24.9$
  &$.404$
  
  &$20.0$
  &$.308$
  
  &$21.9$
  &$.576$
  
  &$22.2$
  &$.380$
  
  \\
  Neural Volumes~\cite{lombardi2019neural}
  &$15.4$
  &$.616$
  
  &$15.7$
  &$.595$
  
  &$15.2$
  &$.588$
  
  &$16.2$
  &$.569$
  
  &$13.8$
  &$.533$
  
  &$13.7$
  &$.473$
  
  &$15.0$
  &$.562$
  
  &$16.2$
  &$.198$
  
  &$18.5$
  &$.559$
  
  &$13.1$
  &$.516$
  
  &$16.1$
  &$.544$
  
  &$16.0$
  &$.454$
  
  \\
  NSFF\textsuperscript{\textdagger}
  &$19.6$
  &$.407$
  
  &$21.5$
  &$.402$
  
  &$18.0$
  &$.432$
  
  &$21.4$
  &$.317$
  
  &$20.5$
  &$.239$
  
  &\tablefirst$26.9$
  &$.208$
  
  &$21.3$
  &$.334$
  
  &\tablefirst$27.7$
  &\tablefirst$.0803$
  
  &\tablefirst$30.6$
  &$.245$
  
  &$21.7$
  &$.205$
  
  &\tablefirst$28.2$
  &\tablefirst$.202$
  
  &\tablefirst$27.1$
  &$.183$
  
  \\
  $\gamma(t)$ + Trans\textsuperscript{\textdagger}~\cite{li2020neural}
  &$22.2$
  &$.354$
  
  &$20.8$
  &$.471$
  
  &$20.7$
  &$.426$
  
  &$22.5$
  &$.344$
  
  &$21.9$
  &$.283$
  
  &\tablesecond$25.3$
  &$.420$
  
  &$22.2$
  &$.383$
  
  &\tablesecond$23.7$
  &$.151$
  
  &\tablesecond$27.2$
  &$.391$
  
  &\tablefirst$22.9$
  &$.221$
  
  &\tablesecond$23.4$
  &$.627$
  
  &\tablesecond$24.3$
  &$.347$
  
  \\ \hline
  Ours ($\lambda=0.01$)
  &$23.4$
  &\tablefirst$.305$
  
  &\tablethird$22.2$
  &\tablethird$.391$
  
  &\tablesecond$24.6$
  &\tablethird$.319$
  
  &\tablesecond$23.9$
  &\tablethird$.280$
  
  &\tablethird$23.6$
  &\tablethird$.232$
  
  &$22.9$
  &\tablesecond$.159$
  
  &\tablethird$23.4$
  &\tablefirst$.281$
  
  &$22.4$
  &$.0872$
  
  &$23.9$
  &\tablefirst$.161$
  
  &\tablesecond$22.4$
  &\tablefirst$.130$
  
  &$21.5$
  &\tablesecond$.245$
  
  &$22.5$
  &\tablefirst$.156$
  
  \\
  Ours ($\lambda=0.001$)
  &\tablefirst$24.2$
  &\tablesecond$.307$
  
  &\tablesecond$23.2$
  &\tablesecond$.391$
  
  &\tablefirst$24.9$
  &\tablefirst$.312$
  
  &$23.5$
  &\tablesecond$.279$
  
  &\tablefirst$23.7$
  &\tablefirst$.230$
  
  &$22.8$
  &\tablethird$.174$
  
  &\tablefirst$23.7$
  &\tablesecond$.282$
  
  &$21.8$
  &$.0962$
  
  &$23.6$
  &\tablethird$.175$
  
  &\tablethird$22.1$
  &\tablesecond$.132$
  
  &$21.0$
  &\tablethird$.270$
  
  &$22.1$
  &\tablesecond$.168$
  
  \\
  No elastic
  &$23.1$
  &$.317$
  
  &\tablefirst$24.2$
  &\tablefirst$.382$
  
  &$24.1$
  &$.322$
  
  &$22.9$
  &$.290$
  
  &\tablesecond$23.7$
  &\tablesecond$.230$
  
  &\tablethird$23.0$
  &$.257$
  
  &\tablesecond$23.5$
  &$.300$
  
  &$22.2$
  &\tablethird$.0863$
  
  &$23.7$
  &\tablesecond$.174$
  
  &$22.0$
  &\tablethird$.132$
  
  &$20.9$
  &$.287$
  
  &$22.2$
  &\tablethird$.170$
  
  \\
  No coarse-to-fine
  &\tablethird$23.8$
  &\tablethird$.312$
  
  &$21.9$
  &$.408$
  
  &$24.5$
  &$.321$
  
  &\tablefirst$24.0$
  &\tablefirst$.277$
  
  &$22.8$
  &$.242$
  
  &$22.7$
  &$.244$
  
  &$23.3$
  &$.301$
  
  &$22.3$
  &$.0960$
  
  &$24.3$
  &$.257$
  
  &$21.8$
  &$.151$
  
  &$21.9$
  &$.406$
  
  &\tablethird$22.6$
  &$.228$
  
  \\
  No SE3
  &$23.5$
  &$.314$
  
  &$21.9$
  &$.401$
  
  &\tablethird$24.5$
  &\tablesecond$.317$
  
  &\tablethird$23.7$
  &$.282$
  
  &$22.7$
  &$.235$
  
  &$22.9$
  &$.206$
  
  &$23.2$
  &\tablethird$.293$
  
  &$22.4$
  &$.0867$
  
  &$23.5$
  &$.191$
  
  &$21.2$
  &$.156$
  
  &$20.9$
  &$.276$
  
  &$22.0$
  &$.177$
  
  \\
  Ours (base)
  &\tablesecond$24.0$
  &$.319$
  
  &$20.9$
  &$.456$
  
  &$23.5$
  &$.345$
  
  &$22.4$
  &$.323$
  
  &$22.1$
  &$.254$
  
  &$22.7$
  &$.184$
  
  &$22.6$
  &$.314$
  
  &\tablethird$22.6$
  &$.127$
  
  &$24.3$
  &$.298$
  
  &$21.1$
  &$.173$
  
  &\tablethird$22.1$
  &$.503$
  
  &$22.5$
  &$.275$
  
  \\
  No BG Loss
  &$22.3$
  &$.317$
  
  &$21.5$
  &$.395$
  
  &$20.1$
  &$.371$
  
  &$22.5$
  &$.290$
  
  &$20.3$
  &$.260$
  
  &$22.3$
  &\tablefirst$.145$
  
  &$21.5$
  &$.296$
  
  &$22.3$
  &\tablesecond$.0856$
  
  &$23.5$
  &$.210$
  
  &$20.4$
  &$.161$
  
  &$20.9$
  &$.330$
  
  &$21.8$
  &$.196$
  
  \\
\bottomrule

\end{tabular}


%% file: fa0_vrig_results_dynamic.tex
\definecolor{bestcolor}{rgb}{0.85, 0.113, 0.188}

\renewcommand{\textimage}[4]{
	\begin{overpic}[width=2.35cm,unit=1mm,clip,trim=#1]{figures/vrig_results_v3/#2}
	\put (9.9,0.7) {\sethlcolor{white}\footnotesize\hl{$#3 / #4$}}
    \end{overpic}
}
\newcommand{\textimagelonger}[4]{
	\begin{overpic}[width=2.35cm,unit=1mm,clip,trim=#1]{figures/vrig_results_v3/#2}
	\put (8.5,0.7) {\sethlcolor{white}\footnotesize\hl{$#3 / #4$}}
    \end{overpic}
}
\newcommand{\animage}[2]{\includegraphics[width=2.35cm,clip,trim=#1]{figures/vrig_results_v3/#2}}
\newcommand{\bestnum}[1]{\textcolor{bestcolor}{#1}}

\begin{figure*}[t!]
    \setlength{\tabcolsep}{0.5pt}
    \renewcommand{\arraystretch}{0.25}
    \begin{tabular}{cccccccc}
        \makebox[20pt]{\raisebox{50pt}{\rotatebox[origin=c]{90}{\textsc{Drinking}}}} \hspace{-7pt} &
        \animage{0 0 0 0}{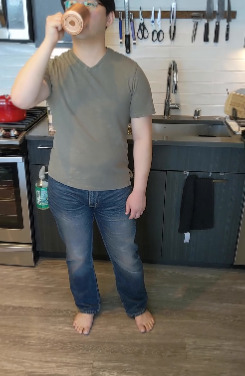} &
        \animage{0 0 0 0}{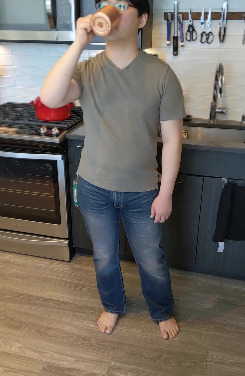} &
        \textimagelonger{0 0 0 0}{drinking-from-cup1.full_model_se3_w0.01.pred.jpg}{22.6}{\bestnum{0.0826}} &
        \textimage{0 0 0 0}{drinking-from-cup1.d-nerf.pred.jpg}{{25.7}}{0.156} &
        \textimagelonger{0 0 0 0}{drinking-from-cup1.nsff.pred.jpg}{\bestnum{29.9}}{0.0851} &
        \textimage{0 0 0 0}{drinking-from-cup1.neuralvolumes.pred.jpg}{16.6}{0.183} &
        \textimage{0 0 0 0}{drinking-from-cup1.nerf.pred.jpg}{20.1}{0.379} \\
        
        \makebox[20pt]{\raisebox{45pt}{\rotatebox[origin=c]{90}{\textsc{badminton}}}} \hspace{-7pt} &
        \animage{30 60 30 40}{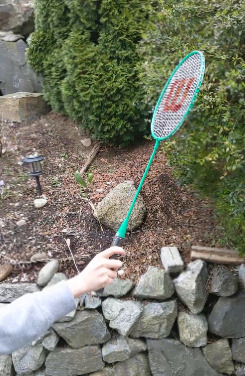} &
        \animage{30 60 30 40}{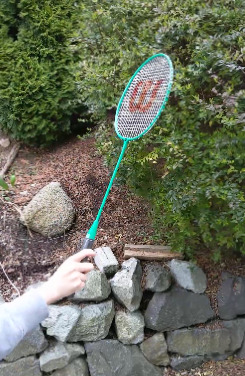} &
        \textimage{30 60 30 40}{badminton1.full_model_se3_w0.01.pred.jpg}{22.4}{\bestnum{0.118}} &
        \textimage{30 60 30 40}{badminton1.d-nerf.pred.jpg}{\bestnum{22.9}}{0.198} &
        \textimage{30 60 30 40}{badminton1.nsff.pred.jpg}{21.6}{0.217} &
        \textimage{30 60 30 40}{badminton1.neuralvolumes.pred.jpg}{{13.1}}{0.520} &
        \textimage{30 60 30 40}{badminton1.nerf.pred.jpg}{19.5}{0.360} \\
        
        \makebox[20pt]{\raisebox{50pt}{\rotatebox[origin=c]{90}{\textsc{Tail}}}} \hspace{-7pt} &
        \animage{0 0 0 0}{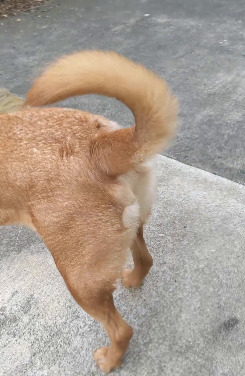} &
        \animage{0 0 0 0}{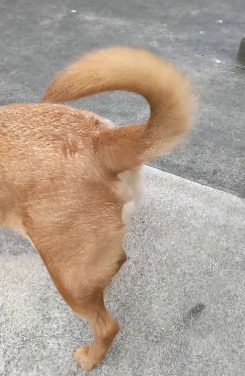} &
        \textimage{0 0 0 0}{toby-tail1.full_model_se3_w0.01.pred.jpg}{25.1}{\bestnum{0.127}} &
        \textimage{0 0 0 0}{toby-tail1.d-nerf.pred.jpg}{{26.9}}{0.353} &
        \textimage{0 0 0 0}{toby-tail1.nsff.pred.jpg}{\bestnum{30.4}}{0.252} &
        \textimage{0 0 0 0}{toby-tail1.neuralvolumes.pred.jpg}{16.5}{0.462} &
        \textimage{0 0 0 0}{toby-tail1.nerf.pred.jpg}{22.8}{0.506} \\
               
        \makebox[20pt]{\raisebox{45pt}{\rotatebox[origin=c]{90}{\textsc{Broom}}}} \hspace{-7pt} &
        \animage{60 80 00 50}{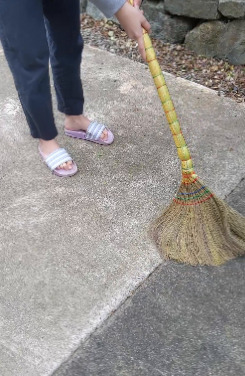} &
        \animage{30 80 30 50}{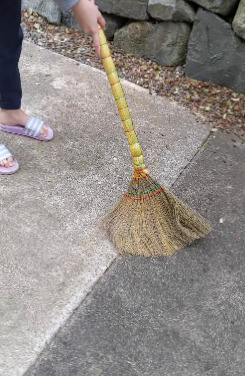} &
        \textimage{30 80 30 50}{broom2.full_model_se3_w0.01.pred.jpg}{20.7}{{0.240}} &
        \textimage{30 80 30 50}{broom2.d-nerf.pred.jpg}{{22.8}}{0.600} &
        \textimage{30 80 30 50}{broom2.nsff.pred.jpg}{\bestnum{28.9}}{\bestnum{0.159}} &
        \textimage{30 80 30 50}{broom2.neuralvolumes.pred.jpg}{{14.2}}{0.472} &
        \textimage{30 80 30 50}{broom2.nerf.pred.jpg}{18.7}{0.677} \\

        \\[2pt]
        
        & {\small Training view} & 
        {\small Novel view (GT)} & 
        {\small Ours} &
        {\small $\gamma$(t)+trans} &
        {\small NSFF~\cite{li2020neural}} &
        {\small NV~\cite{lombardi2019neural}} &
        {\small NeRF~\cite{mildenhall2020nerf}}
        \\
    \end{tabular}
    \vspace{-6pt}
    \caption{Comparisons of baselines and our method on dynamic scenes. {\footnotesize PSNR / LPIPS} metrics on bottom right with best colored \bestnum{red}. Note how better metrics do not necessarily translate to better quality.}
    \label{fig:additional-vrig-results-dynamic}
\end{figure*}

%% file: fa0_vrig_results_static.tex
\definecolor{bestcolor}{rgb}{0.85, 0.113, 0.188}

\renewcommand{\textimage}[4]{
	\begin{overpic}[width=2.35cm,unit=1mm,clip,trim=#1]{figures/vrig_results_v3/#2}
	\put (9.9,0.7) {\sethlcolor{white}\footnotesize\hl{$#3 / #4$}}
    \end{overpic}
}

\renewcommand{\animage}[2]{\includegraphics[width=2.35cm,clip,trim=#1]{figures/vrig_results_v3/#2}}
\renewcommand{\bestnum}[1]{\textcolor{bestcolor}{#1}}

\begin{figure*}[t!]
    \setlength{\tabcolsep}{0.5pt}
    \renewcommand{\arraystretch}{0.25}
    \begin{tabular}{cccccccc}
            
        \makebox[20pt]{\raisebox{50pt}{\rotatebox[origin=c]{90}{\textsc{Glasses}}}} \hspace{-7pt} &
        \animage{0 0 0 0}{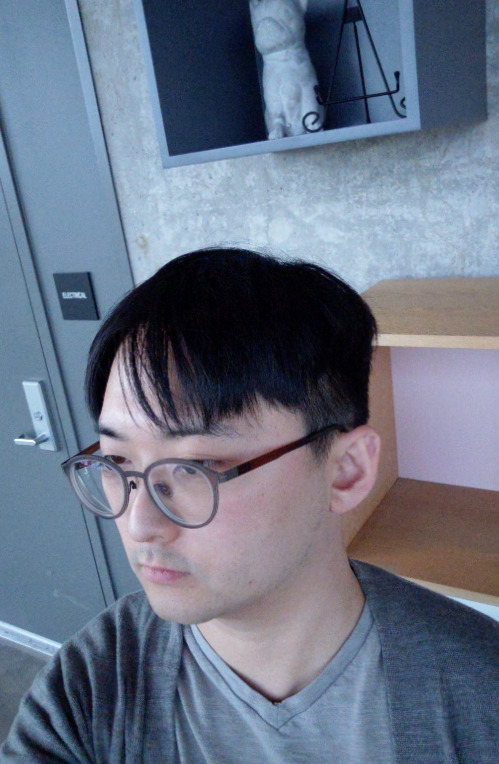} &
        \animage{0 0 0 0}{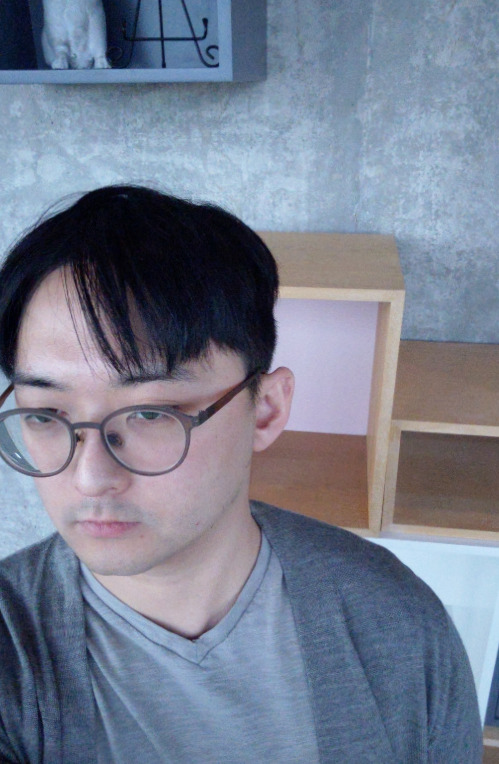} &
        \textimage{0 0 0 0}{keunhong-vrig4.full_model_se3_w0.01.pred.jpg}{\bestnum{22.1}}{\bestnum{0.329}} &
        \textimage{0 0 0 0}{keunhong-vrig4.d-nerf.pred.jpg}{19.3}{0.417} &
        \textimage{0 0 0 0}{keunhong-vrig4.nsff.pred.jpg}{14.2}{0.497} &
        \textimage{0 0 0 0}{keunhong-vrig4.neuralvolumes.pred.jpg}{14.2}{0.660} &
        \textimage{0 0 0 0}{keunhong-vrig4.nerf.pred.jpg}{15.5}{0.543} \\
        
        \makebox[20pt]{\raisebox{45pt}{\rotatebox[origin=c]{90}{\textsc{Beanie}}}} \hspace{-7pt} &
        \animage{0 0 0 0}{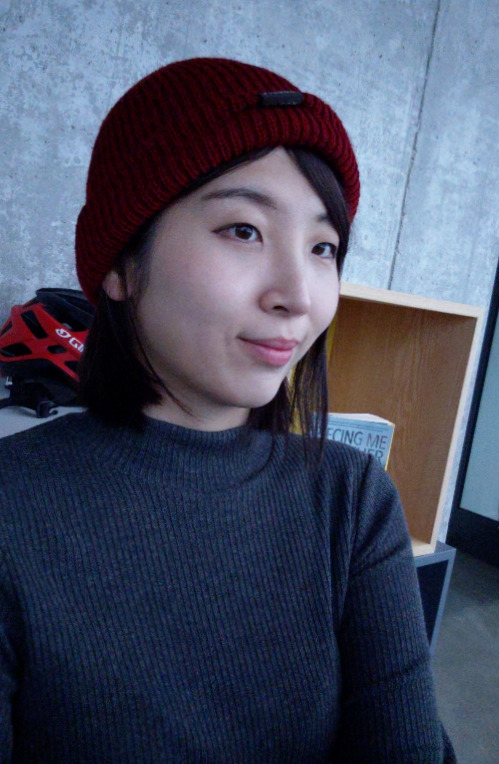} &
        \animage{0 0 0 0}{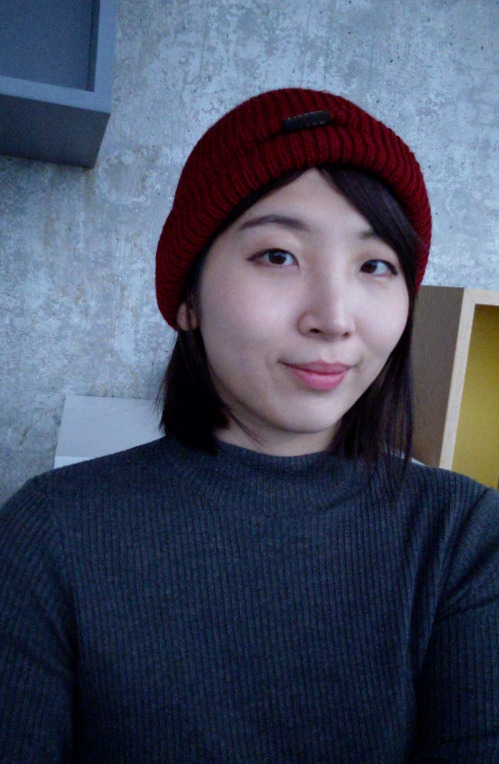} &
        \textimage{0 0 0 0}{haley-vrig-beanie.full_model_se3.pred.jpg}{\bestnum{21.6}}{\bestnum{0.410}} &
        \textimage{0 0 0 0}{haley-vrig-beanie.d-nerf.pred.jpg}{17.1}{0.512} &
        \textimage{0 0 0 0}{haley-vrig-beanie.nsff.pred.jpg}{20.8}{0.429} &
        \textimage{0 0 0 0}{haley-vrig-beanie.neuralvolumes.pred.jpg}{16.3}{0.573} &
        \textimage{0 0 0 0}{haley-vrig-beanie.nerf.pred.jpg}{13.5}{0.667} \\
        
        \makebox[20pt]{\raisebox{50pt}{\rotatebox[origin=c]{90}{\textsc{Curls}}}} \hspace{-7pt} &
        \animage{0 0 0 0}{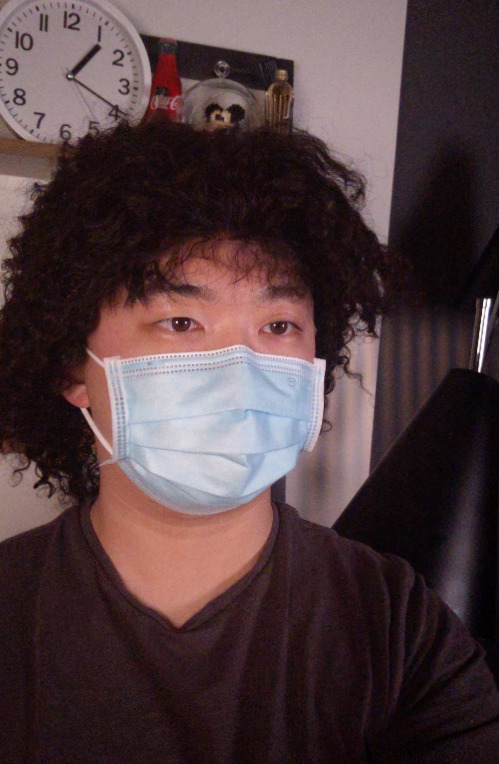} &
        \animage{0 0 0 0}{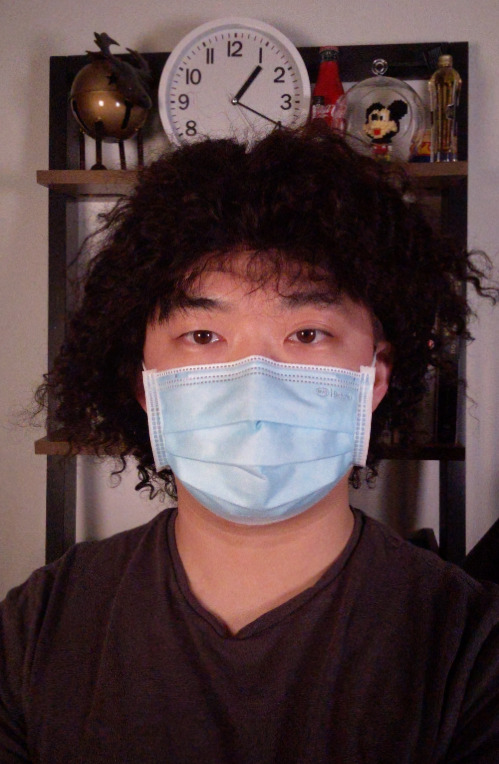} &
        \textimage{0 0 0 0}{keunhong-vrig-bookshelf.full_model_se3.pred.jpg}{\bestnum{26.2}}{\bestnum{0.250}} &
        \textimage{0 0 0 0}{keunhong-vrig-bookshelf.d-nerf.pred.jpg}{23.3}{0.357} &
        \textimage{0 0 0 0}{keunhong-vrig-bookshelf.nsff.pred.jpg}{18.8}{0.411} &
        \textimage{0 0 0 0}{keunhong-vrig-bookshelf.neuralvolumes.pred.jpg}{15.1}{0.576} &
        \textimage{0 0 0 0}{keunhong-vrig-bookshelf.nerf.pred.jpg}{12.8}{0.657} \\
        
        \makebox[20pt]{\raisebox{45pt}{\rotatebox[origin=c]{90}{\textsc{Kitchen}}}} \hspace{-7pt} &
        \animage{0 0 0 0}{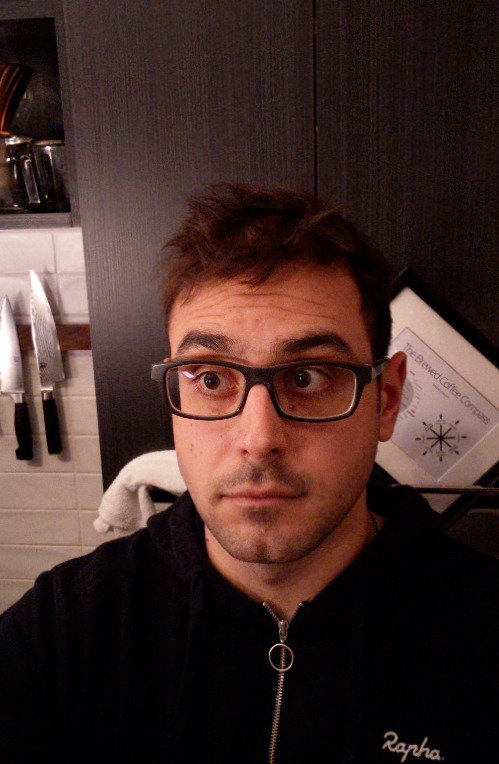} &
        \animage{0 0 0 0}{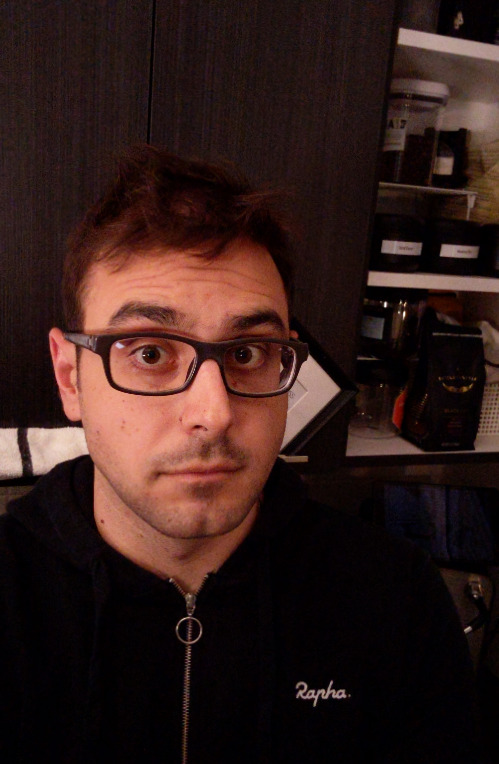} &
        \textimage{0 0 0 0}{aleks-vrig3-v2.full_model_se3_w0.01.pred.jpg}{\bestnum{25.6} }{\bestnum{0.277}} &
        \textimage{0 0 0 0}{aleks-vrig3-v2.d-nerf.pred.jpg}{22.4}{0.350} &
        \textimage{0 0 0 0}{aleks-vrig3-v2.nsff.pred.jpg}{22.1}{0.340} &
        \textimage{0 0 0 0}{aleks-vrig3-v2.neuralvolumes.pred.jpg}{17.2}{0.571} &
        \textimage{0 0 0 0}{aleks-vrig3-v2.nerf.pred.jpg}{20.8}{0.393} \\
         
        \makebox[20pt]{\raisebox{50pt}{\rotatebox[origin=c]{90}{\textsc{Lamp}}}} \hspace{-7pt} &
        \animage{0 0 0 0}{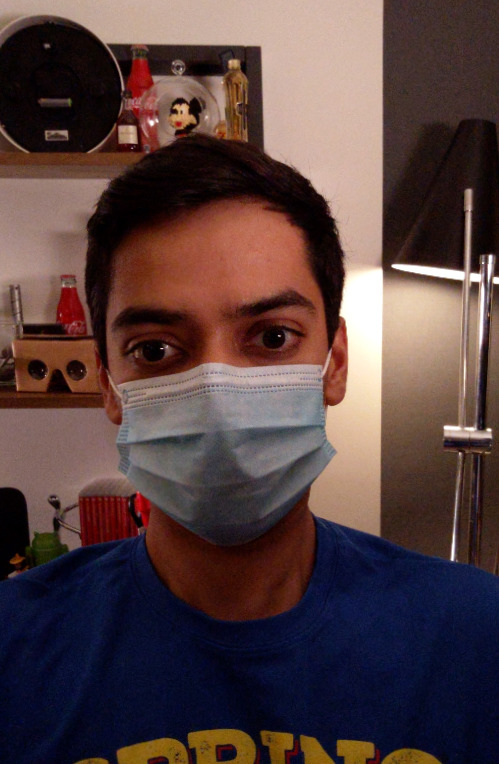} &
        \animage{0 0 0 0}{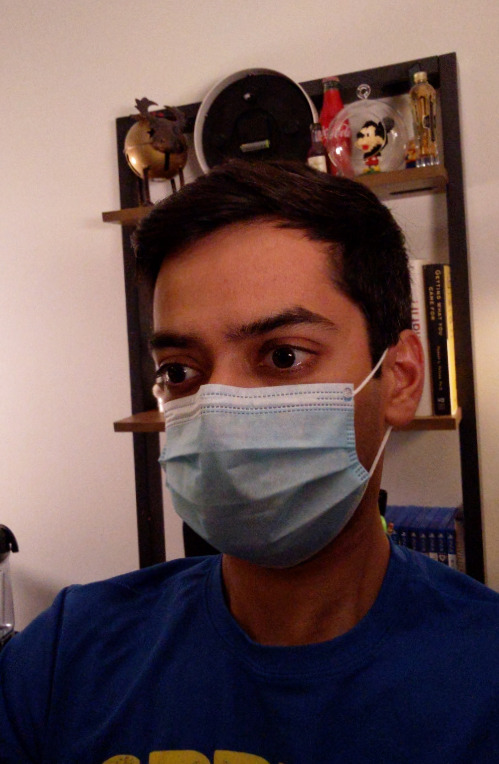} &
        \textimage{0 0 0 0}{yash-vrig-mask.full_model_se3.pred.jpg}{20.7}{0.258} &
        \textimage{0 0 0 0}{yash-vrig-mask.d-nerf.pred.jpg}{18.0}{0.336} &
        \textimage{0 0 0 0}{yash-vrig-mask.nsff.pred.jpg}{\bestnum{21.4}}{\bestnum{0.250}} &
        \textimage{0 0 0 0}{yash-vrig-mask.neuralvolumes.pred.jpg}{14.2}{0.523} &
        \textimage{0 0 0 0}{yash-vrig-mask.nerf.pred.jpg}{17.5}{0.438} \\
        
        \makebox[20pt]{\raisebox{40pt}{\rotatebox[origin=c]{90}{\textsc{Toby Sit}}}} \hspace{-7pt} &
        \animage{70 140 20 20}{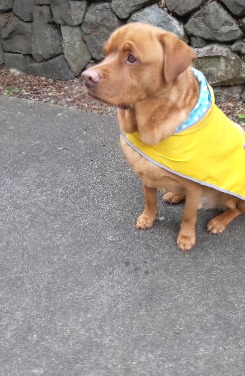} &
        \animage{50 140 40 20}{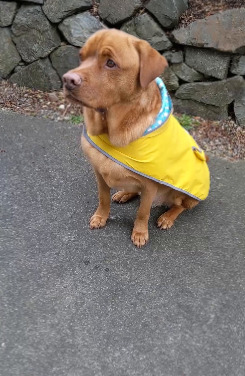} &
        \textimage{50 140 40 20}{toby-sit-outside1.full_model_se3.pred.jpg}{23.2}{\bestnum{0.143}} &
        \textimage{50 140 40 20}{toby-sit-outside1.d-nerf.pred.jpg}{{25.6}}{0.435} &
        \textimage{50 140 40 20}{toby-sit-outside1.nsff.pred.jpg}{\bestnum{27.8}}{0.179} &
        \textimage{50 140 40 20}{toby-sit-outside1.neuralvolumes.pred.jpg}{{13.4}}{0.488} &
        \textimage{50 140 40 20}{toby-sit-outside1.nerf.pred.jpg}{22.2}{0.469} \\
        
        \\[2pt]
        
        & {\small Training view} & 
        {\small Novel view (GT)} & 
        {\small Ours} &
        {\small $\gamma$(t)+trans} &
        {\small NSFF~\cite{li2020neural}} &
        {\small NV~\cite{lombardi2019neural}} &
        {\small NeRF~\cite{mildenhall2020nerf}}
        \\
    \end{tabular}
    \vspace{-6pt}
    \caption{Comparisons of baselines and our method on quasi-static scenes. {\footnotesize PSNR / LPIPS} metrics on bottom right with best colored \bestnum{red}. Note how better metrics do not necessarily translate to better quality.}
    \label{fig:additional-vrig-results-static}
    \vspace{-12pt}
\end{figure*}

%% file: f0x_failure_cases.tex
\fboxsep=0pt 
\fboxrule=0.4pt 

\begin{figure}[b]
	\centering
		\begin{subfigure}{1.0\columnwidth}
    	\centering
    	\figcelltb{0.3}{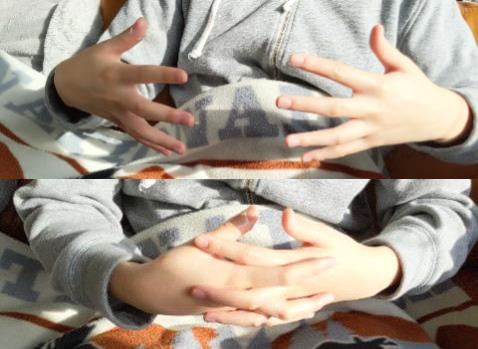}{}{clip,trim=0 0 0 0}\hspace{-1.0pt}
    	\figcelltb{0.34}{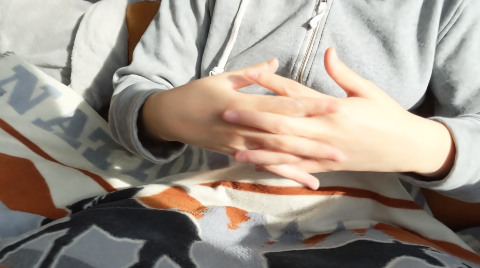}{}{clip,trim=130 35 45 35}
    	\figcelltb{0.34}{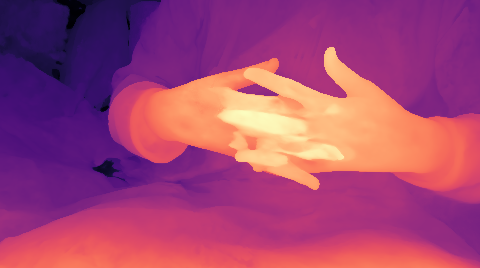}{}{clip,trim=130 35 45 35}
	\end{subfigure}
	\begin{subfigure}{1.0\columnwidth}
    	\centering
    	\figcelltb{0.3}{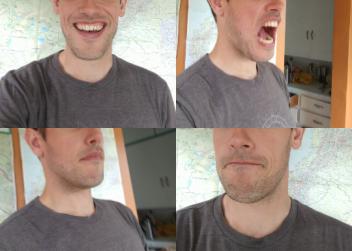}{\makecell{example inputs}}{clip,trim=0 0 0 0}\hspace{-1.0pt}
    	\figcelltb{0.34}{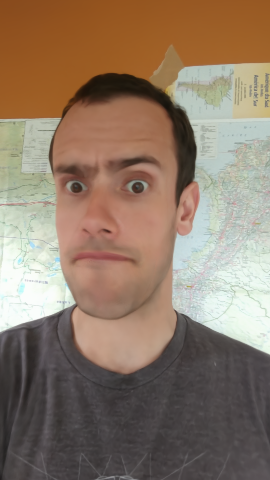}{\makecell{rendered color}}{clip,trim=0 110 0 200}
    	\figcelltb{0.34}{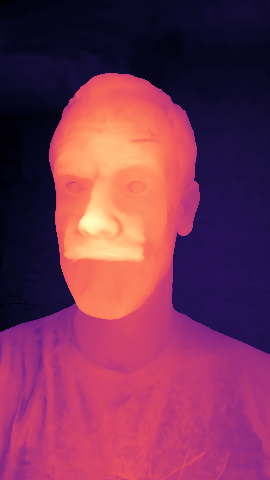}{\makecell{rendered depth}}{clip,trim=0 110 0 200}
	\end{subfigure}
	\vspace{-8pt}
	\caption{
Our method fails in the presence of topological changes. Although the rendered color views look good from some viewpoints, the recovered geometry is incorrect. 
	}
	\vspace{-15pt}
   \label{fig:failure_cases}
\end{figure}

%% file: 06_conclusion.tex
\vspace{-4pt}
\section{Conclusion}

Deformable Neural Radiance Fields extend NeRF by modeling non-rigidly deforming scenes. We show that our as-rigid-as-possible deformation prior, and coarse-to-fine deformation regularization are the key to obtaining high-quality results. We showcase the application of casual selfie captures (\emph{nerfies}), and enable high-fidelity reconstructions of human subjects using a cellphone capture. Future work may tackle larger/faster motion, topological variations, and enhance the speed of training/inference.

\section*{Acknowledgments}
We thank Peter Hedman and Daniel Duckworth for providing feedback in early drafts, and all our capture subjects for their patience, including Toby who was a good boy.

%% file: A1_appendix.tex
\appendix

\section{Details of SE(3) Field Formulation}
\label{sec:se3_details}

As mentioned in the main text, we encode a rigid transform as a screw axis~\cite{lynch2017modern} $\mathcal{S} = (\logrot; \mat{v})\in\mathbb{R}^6$ where
\begin{align}
    e^{\logrot} \equiv e^{[\logrot]_\times} = \mat{I} + \frac{\sin\theta}{\theta}[\logrot]_\times + \frac{1-\cos\theta}{\theta^2}[\logrot]_\times^2\,.
\end{align}
$\skewmat{x}$ is a skew-symmetric matrix also known as the cross-product matrix of a vector $\mat{x}$ since given two 3-vectors $\mat{a}$ and $\mat{b}$, $\skewmat{a}\mat{b}$ gives the cross product $\mat{a}\times\mat{b}$.
\begin{align}
    [\mat{x}]_\times &= \begin{pmatrix}
        0 & -x_3 & x_2 \\
        x_3 & 0 & -x_1 \\
        -x_2 & x_1 & 0
    \end{pmatrix}\,.
\end{align}
The translation encoded by the screw motion $\mathcal{S}$ can be recovered as $\mat{p} = \mat{G}\mat{v}$ where
\begin{align}
    \mat{G} &= \mat{I} + \frac{1-\cos{\theta}}{\theta^2}[\logrot]_\times + \frac{\theta - \sin\theta}{\theta^3}[\logrot]^2_\times\,.
\end{align}
The exponential of $\mathcal{S}$ can also be expressed in homogeneous matrix form $e^{\mathcal{S}} \in \SETHREE$:
\begin{align}
    e^{\mathcal{S}} &= \begin{pmatrix}
     e^{\logrot} & \mat{p} \\
     0 & 1
    \end{pmatrix}\,.
\end{align}
The deformed point is then given by $\mat{x}' = e^{\mathcal{S}} \mat{x}$.

\begin{figure}[t]
    \centering
    \includegraphics*[width=0.6\columnwidth]{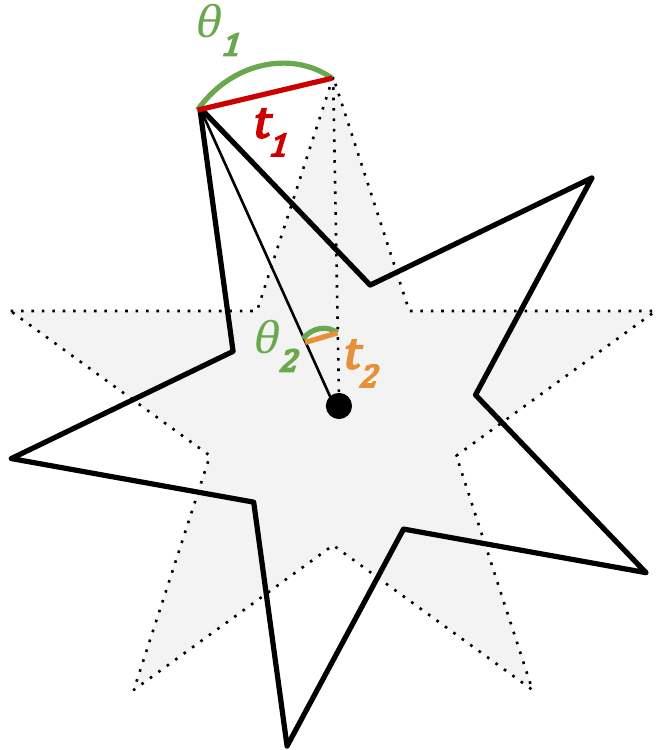}
    \caption{Here we illustrate why a rigid transformation field works better than a translation field with a simple toy example where a star is rotated counter-clockwise around its center. A translation field requires different parameters for every point to encode the rotation (e.g., $\norm{\mat{t_1}}\gg\norm{\mat{t_2}}$) whereas a rotation field only needs a single parameter to encode the rotation (e.g., $\theta_1=\theta_2$). More details in $\secref{sec:se3_details}$.}
    \label{fig:rotation_example}
\end{figure}

\paragraph{Why does an SE(3) field work better?} Consider the example in \figref{fig:rotation_example} where a star has been rotated counter-clockwise along its center. Now consider what transformation would be required at every point on the star to encode this rotation. With a translation field, points towards the center (e.g., $\mat{t_2}$) need translations of small magnitude while points towards the outside (e.g., $\mat{t_1}$) need translations of larger magnitude. Every point on the star requires a different parameter to encode a simple rotation. On the otherhand, with a rotation, every point on the star can be parameterized by a single angle which is the angle of rotation $\theta=\theta_1=\theta_2$. This makes optimization much easier since the deformation field MLP only needs to predict a single parameter across space. We illustrate this further in \secref{sec:2d_example}.

\section{Details of Coarse-to-Fine Optimization}

\mypar{Window Function} Our coarse-to-fine deformation regularization is implemented by windowing the frequency bands of the positional encoding. Eqn. 8 of the main paper defines this windowing function as a weight applied to each frequency band. We visualize our windowing function for different values of $\alpha$ in \figref{fig:anneal_window}.

\mypar{NTK} We also show a visualization of the neural tangent kernel (NTK) induced by our annealed positional encoding in \figref{fig:ntk}. This figure shows the normalized NTK for an 8 layer MLP of width $256$. Note how the bandwidth of the interpolation kernel gets narrower as the value of $\alpha$ increases.

\section{Details of Elastic Regularization}

\newcommand{\U}[0]{\mat{U}}
\newcommand{\D}[0]{\mat{D}}
\newcommand{\VV}[0]{\mat{V}}
\newcommand{\M}[0]{\mat{M}}
\newcommand{\JJ}[0]{\mat{J}}
\newcommand{\R}[0]{\mat{R}}

\paragraph{Motivation for elastic energy formulation.} Elastic energies are often implemented as the deviation of the Jacobian $\JJ$ from the closest rotation $\mat{R}$: $\norm{\mat{J}-\mat{R}}_F$~[p45]. Let $\JJ=\U\D\VV^T$ be the SVD of $\JJ$, then $\R=\U\M\VV^T$ where $\M=\diag\left(1,\ldots,1,\det(\U\VV^T)\right)$. It follows that:

\begin{small}
\begin{align}
    \norm{\JJ-\R}_F&=\norm{\U\D\VV^T-\U\M\VV^T}_F \\
    &=\norm{\U(\D-\M)\VV^T}_F \\
    &=\sqrt{\tr\left(\U(\D-\M)\VV^T\VV(\D-\M)\U^T\right)} \\
    &=\sqrt{\tr\left(\U(\D-\M)^2\U^T\right)} \\
    &=\sqrt{\tr\left((\D-\M)^2\right)} \\
    &=\sqrt{\sum_j{(\sigma_j-m_j)^2}}\,,
\end{align}
\end{small}
where $m_j$ is the $j$th diagonal of $\M$ and $\sigma_j$ is the $j$th singular value of $\JJ$. This is equivalent to penalizing the deviation of the singular values of $\JJ$ from 1. The $\M$ matrix factors in reflections as negative singular values rather a reflection in $\U$ or $\VV$. Because this formulation penalizes expansions more than contractions of the same factor, we penalize the log of the singular values directly.

\input{fa0_ntk}

\input{fa0_ntk_window}

\section{Additional Illustrations}

\paragraph{Unintentional Movement:} In \figref{fig:warp_magnitude} we show an example of how a person can move even when trying to sit still. We visualize the degree of movement by showing the difference in predicted depth as well as by showing a direct plot of the magnitude of the deformation field at the predicted depth point.

\paragraph{Domain Agnostic:} In \figref{fig:qualitative_toby}, \figref{fig:additional-vrig-results-static}, and \figref{fig:additional-vrig-results-dynamic}. we show that our method works agnostic of the type of subject.

\input{f0x_warp_magnitude}

\input{f0x_toby.tex}

\section{Additional Implementation Details}

\mypar{Architecture Details} We provide architecture details of the deformation field network and canonical NeRF networks in Figures~\ref{fig:deformation_network} and~\ref{fig:template_network} respectively.

\mypar{Training} We train our network using the Adam optimizer~\cite{kingma2014adam} with a learning rate exponentially decayed by a factor 0.1 until the maximum number of iterations is reached. The exact hyper-parameters for each configuration are provided in \tabref{table:hyper_params}.

\mypar{Background Regularization} Since the total number of background points varies per scene, we sample $16384$ points for each iteration when computing the background regularization loss in order to avoid memory issues. We additionally jitter each input point using Gaussian noise $\varepsilon\sim\mathcal{N}(0,0.001)$ and use a robust Geman-McClure loss function~\cite{geman1985bayesian} with $\alpha=-2$ and $c=0.001$ implemented as per Barron~\cite{barron2019general}.

\paragraph{Implementation:} We extend the JAX~\cite{jax2018github} implementation of NeRF~\cite{jaxnerf2020github} for our method.

\input{fa0_arch_deformation}

\input{fa0_arch_template}

\input{ta0_ssim_table}

\section{Experiment Details}

\subsection{Dataset Processing}

\mypar{Blurry Frame Filtering}
For video captures, we filter blurry frames using the variance of the Laplacian~\cite{pech2000diatom}. To compute the blur score for an image, we apply the Laplace operator with kernel size 3 and compute the variance of the resulting image.
We then filter the images based on this score to leave around 600 frames for each capture.

\mypar{Camera Registration} For camera registration, we first compute a foreground mask using a semantic segmentation network such as DeepLabV3~\cite{chen2017rethinking}. We then use COLMAP~\cite{schoenberger2016sfm} to compute the camera registration while using the mask to ignore foreground pixels when computing features. We found that this step can improve the quality of the camera registration in the presence of a moving foreground. We skip this step for captures for which we cannot obtain a segmentation mask such as for \textsc{Badminton} and \textsc{Broom}.

\mypar{Facial Landmarks} Although not necessary for our method, we use facial landmarks for selfie and full body captures to estimate a canonical frame of reference. Using this canonical frame of reference, we automatically generate visually appealing novel view trajectories of our reconstructed \emph{nerfies}, like figure-eight camera paths in front of the user. We compute the 2D facial landmarks using \mbox{MediaPipe}'s face mesh~\cite{lugaresi2019mediapipe}, and triangulate them in 3D using the Structure-from-Motion camera poses. We then set our canonicalized coordinate frame that is centered at the facemesh, with a standard orientation ($+y$ up, $+x$ right, $-z$ into the face), and with approximately metric units, by setting the scale so that the distance between the eyes matches the average interpupillary distance of $6$~cm. Note that the 3D triangulation of facial landmarks is only correct if the subject is static, which is not guaranteed in our method, but in practice we observed that the triangulation result is sufficiently good to define the coordinate frame even when the subject rotates the head side-to-side. For the animal captures, we manually generate virtual camera paths.

\subsection{Baselines}

\paragraph{Comparison to Neural Volumes:}
Neural Volumes~\cite{lombardi2019neural} reconstructs a deformable model of a subject captured by dozens of time-synchronized cameras. To apply it to our setting, where only one camera sees the subject at each time instance, we modify the encoder to network to take a single input image instead of three, as in the original method. We disable the background estimation branch and learn instead the complete scene centered around the face and scaled to a unit cube. For each frame, we render the volume from the viewpoint of the second camera of the validation rig and compute image comparison metrics. We provide quantitative comparisons in Table~1 in the main paper, and qualitative comparisons in \figref{fig:additional-vrig-results-static} and \figref{fig:additional-vrig-results-dynamic}.

We use a $128^{3}$ voxel grid, a $32^{3}$ warp field and train the network for 100k iterations for each of the five sequences. We evaluate all results using the same camera parameters and spatial resolution. We show some renderings when interpolating the camera position between training and validation views in the supplementary video.

\paragraph{Comparison to NSFF:} Concurrently to our work, Neural-Scene Flow Fields (NSFF)~\cite{li2020neural} proposes to model dynamic scenes by directly conditioning the NeRF with a position-encoded time variable $\gamma(t)$, modulating color, density, and a scene-flow prediction. Differences from our method are: (a) NSFF directly modulates the density of the NeRF by conditioning it with $\gamma(t)$ while our method uses a deformation field; (b) NSFF uses a position-encoded time variable ({\small$\gamma(t)$}) to condition each observation whereas our method uses a per-example latent code~\cite{bojanowski2018optimizing}; (c) NSFF uses depth from MIDAS~\cite{Ranftl2020} and optical flow from RAFT~\cite{teed2020raft} as supervision whereas our method only uses a photometric loss.

We quantitatively compare with NSFF in Tab. 1 of the paper and in \tabref{table:ssim_table}, and show corresponding qualitative results in \figref{fig:additional-vrig-results-static} and \figref{fig:additional-vrig-results-dynamic}. We use the official code released by the NSFF authors. The authors provided us with hyper-parameters tuned for our datasets.

\paragraph{Additional Metrics:} MS-SSIM metrics are in \tabref{table:ssim_table}.

\begin{table}[b]
\centering
\renewcommand{\arraystretch}{1.05}

\resizebox{\linewidth}{!}{
\begin{tabular}{r|ccccccc}

\toprule
Config & Resolution & Steps & Learning       & Batch  & \multicolumn{2}{c}{\# Samples} & Width \\ 
  &   &  & Rate       &  Size & Fine & Coarse & $W$ \\ 

\hline

\textsc{Full}  & 1080p & 1M    & $7.5\text{e-}4$ & 3072       & 256             & 256               & 256        \\ \hline
\textsc{Half}   & 540p  & 100K      & $1\text{e-}3$   & 8096       & 128             & 128               & 128 \\
\bottomrule
\end{tabular}
}
\caption{Here we provide the hyper-parameters used for each configuration. \textsc{Full} is the full resolution configuration used in our qualitative results. \textsc{Half} is half the resolution of \textsc{Full} and is used for our quantitative evaluation and ablation studies.}
\label{table:hyper_params}
\end{table}

\section{Additional Results}

We show qualitative results from each of the sequences presented in our quantitative evaluation (Tab. 1 of paper, \tabref{table:ssim_table}) for quasi-static scenes (\figref{fig:additional-vrig-results-static}) and dynamic scenes (\figref{fig:additional-vrig-results-dynamic}).

\begin{figure}[t]
    \centering
    \includegraphics*[width=0.7\linewidth,clip,trim=20 70 20 70]{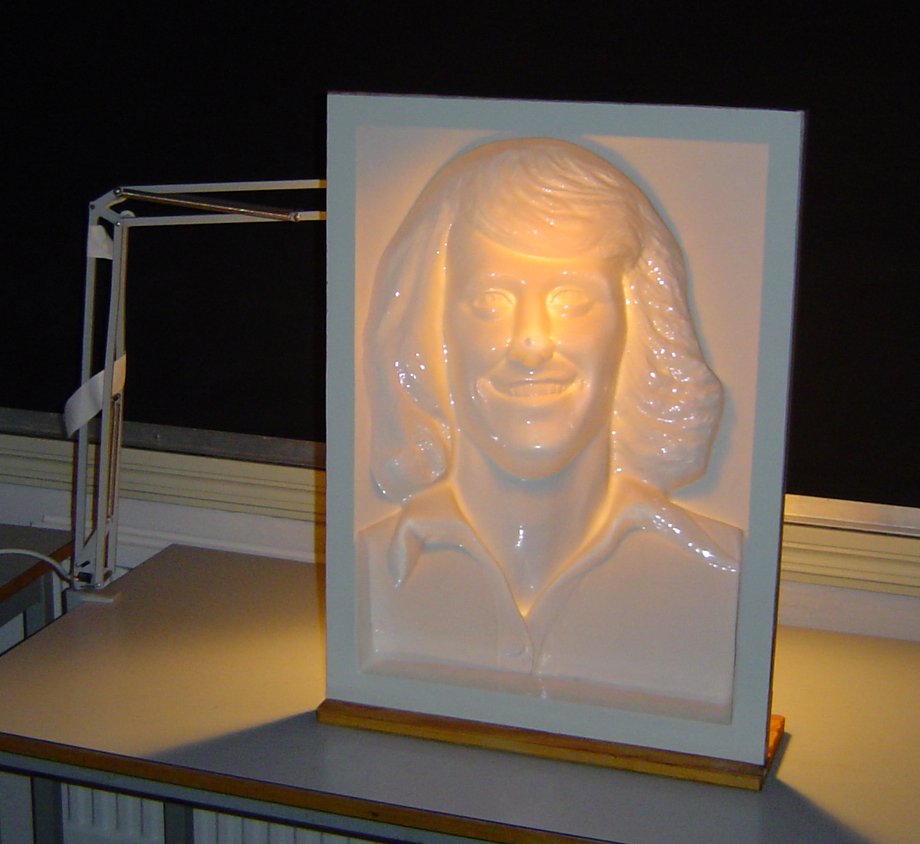}
    \caption{This concave mask of the Swedish tennis player Bjorn Borg appears to be convex due to the hollow-face illusion. By Skagedal, 2004 (Public Domain).}
    \label{fig:bjorn_borg_hollow_face}
\end{figure}

\section{Limitations}

\input{f0x_gazefollow}

\input{fa0_large_motion}

\paragraph{Topological Variation:} Our method struggles when the scene has motion which varies the topology of the scene. For example, when a person opens their mouth (as in Fig. 11 of the main text), the effective topology of their head changes. This is problematic for our method since we use a continuous deformation field parameterized by an MLP. In order to understand this, consider the mouth example and suppose that the template contains the person with their mouth open. Suppose that $\mat{x}_U$ and $\mat{x}_L$ are two adjacent points near the seam of the lips, and the $\mat{x}_U$ is on the upper lip and $\mat{x}_L$ is on the lower lip. It is then evident that a sharp discontinuity in the deformation is required to map both points to their appropriate positions on the template. Such a discontinuity is difficult for our continuous MLP to predict. We find that instead the optimization will often yield an incorrect but valid solution e.g., it will explain a closing mouth by protruding the lip and pulling it down as in Figure 11 of the paper.

\paragraph{Rapid Motion:} NeRF relies on seeing multiple observations to constrain where density lies in the volumes. In the presence of rapid motion, such as in \figref{fig:large_movement}, certain states of the scene may only be visible for a short period of time making it harder to reconstruct.

\paragraph{Orientation flips:} Optimizations solving for any parameterization of rotations are known to be non-convex due to both Gauge ambiguity and the inherent ``twistedness'' of the space of SO(3)~\cite{wilson2016rotations}. As a simple example to illustrate this, imagine trying to align two coins in 3D. If the coin is initialized in a flipped orientation where heads faces the tail side of the other coin, then the `fit' of the two coins must get worse before getting better when rotating towards the global minimum. 

We encounter the same issue when optimizing for our deformation fields. If the template of a scene is in a certain orientation, but the deformation field for an observation is initialized in the wrong orientation the method will get stuck in a local minima and result in sub-optimal alignment. We show an example of this in \figref{fig:large_movement} where frames with Toby's left side visible are reconstructed better than when Toby's right side is visible.

\input{fa0_2d_toy_train}

\paragraph{Hollow Face Illusion:} The hollow-face illusion is an optical illusion where a concave (pushed in) imprint of an object appears to be convex (pushed out) instead. A feature of this illusion is that the convex illusion appears to follow the viewer's eye. This illusion has been purposefully used in the Disneyland haunted mansion to create face busts which appear to follow you and in the popular T-Rex illusion~\cite{trex2015}. We show an example of this illusion in \figref{fig:bjorn_borg_hollow_face}.

We observe that the ambiguity which causes this illusion can also be a failure more for our method. In \figref{fig:gazefollow}, we show an example where a user fixes their gaze in the direction of the camera while capturing themselves. Instead of modeling the eye motion as a deformation, our method models the eyes concavities as can be seen in the geometry.

\section{2D Deformation Experiment}
\label{sec:2d_example}

\input{fa0_2d_toy_se2_vs_translation}

Here we analyze the behavior of a deformation field in a 2D toy setting. In this 2D setting, a "scene" is comprised of a single image which is randomly translated, rotated, and non-linearly distorted near the center. We show the full dataset in \figref{fig:2d_toy_train}. Akin to our deformable NeRF setting, the task is to reconstruct each image by using a 2D deformation field which references a single template. The template is an MLP $F : (x,y) \rightarrow (r,g,b)$ which maps normalized image coordinates $x,y\in[-1,1]$ to color values. The deformation field (i.e., 2D flow) is represented as an MLP $T : (x,y) \rightarrow (t_x, t_y)$ for a translation field or $T : (x,y) \rightarrow (\theta, p_x, p_y, t_x, t_y)$ for a rigid SE(2) transformation field. These are 2D analogs to the 3D translation field and SE(3) field described in Sec. 3.2.

\paragraph{Deformation Formulation:} \figref{fig:2d_toy_se2_vs_trans} shows how an SE(2) rigid transformation field outperforms a translation field. An SE(2) field is able to faithfully reconstruct each image with a reasonable template and smooth deformation field. On the other hand, a translation field is not able to recover a reasonable template, and as a result the reconstruction has many artifacts and the flow field is messy.

\paragraph{Positional Encoding Frequencies:} We show how changing the number of frequency bands changes the convergence behavior in \figref{fig:2d_toy_freqs}. With a small number of frequencies ($m=1$) we are able to converge to the correct orientation in the template, but cannot fully model the non-linear `swirl' towards the center of the image. If we increase the number of frequencies ($m=2\ldots6$) then while we can reconstruct the high swirl better, we start introducing artifacts due to early overfitting of the template and deformation field. With our coarse-to-fine approach we are able to both get the correct orientation without artifacts and also model the swirl.

\input{fa0_2d_toy_freqs}

%% file: fa0_ntk.tex
\begin{figure}[t]
    \includegraphics*[width=\linewidth]{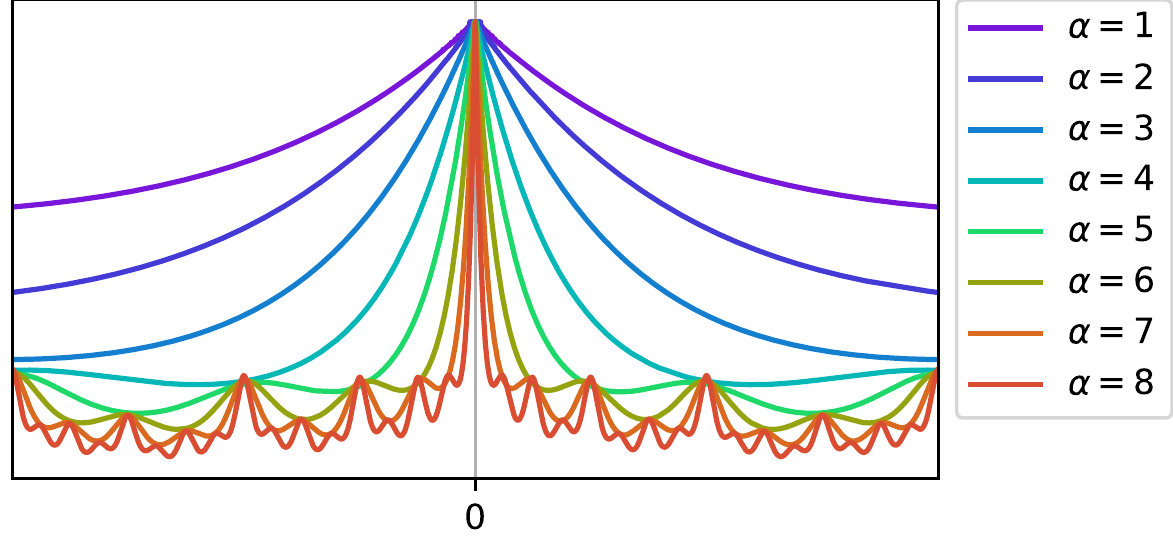}
    \caption{Visualizations of the neural-tangent kernel (NTK)~\cite{jacot2018neural} of our annealed positional encoding for different values of $\alpha$. Our coarse-to-fine optimization scheme works by easing in the influence of each positional encoding frequency through a parameter $\alpha$. This has the effect of shrinking the bandwidth of the NTK corresponding to the deformation MLP as $\alpha$ is increased, thereby allowing higher frequency deformations.}
    \label{fig:ntk}
\end{figure}

%% file: fa0_ntk_window.tex
\begin{figure}[t]
    \includegraphics*[width=\linewidth]{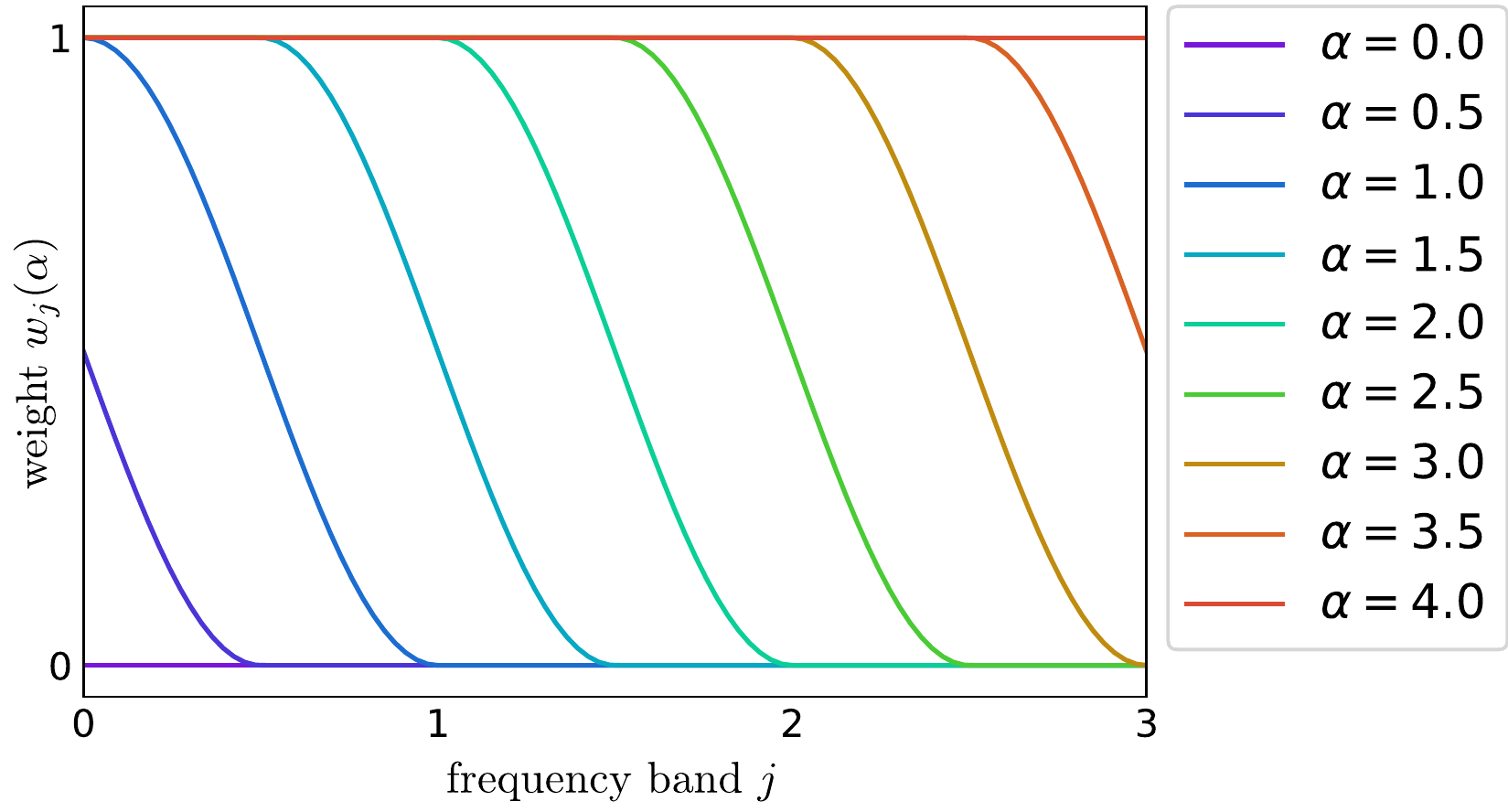}
    \caption{A visualization of the window function $w_j(\alpha)$ for the annealed positional encoding. We show an example with a maximum number of frequency bands of $m=4$ where $j\in\{0,\ldots,m-1\}$. $\alpha=0$ sets the weight of all frequency bands to zero leaving only the identity mapping, while an $\alpha=4$ sets the weight of all frequency bands to one. Increasing the value of $\alpha$ is equivalent to sliding the window to the right across the frequency bands.}
    \label{fig:anneal_window}
\end{figure}

%% file: f0x_warp_magnitude.tex
\begin{figure}[t]
	\centering
	\begin{subfigure}{\columnwidth}
    	\centering
    	\figcellb{0.188}{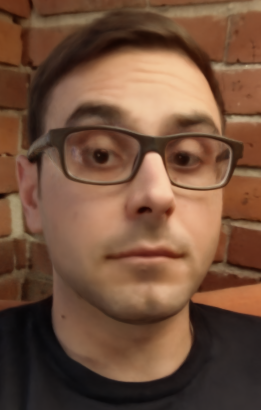}{template color}\hspace{-1pt}
    	\figcellb{0.188}{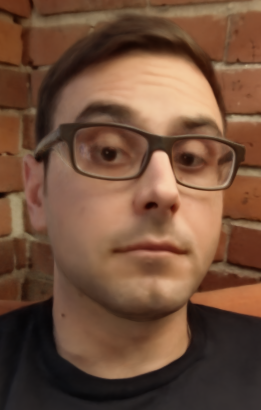}{observation color}\hspace{-1pt}
    	\figcellb{0.188}{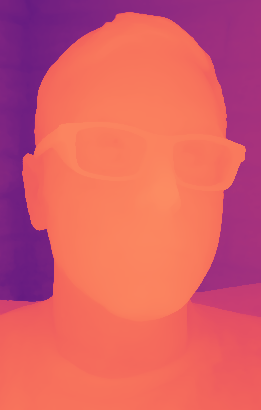}{observation depth}\hspace{-1pt}
    	\figcellb{0.188}{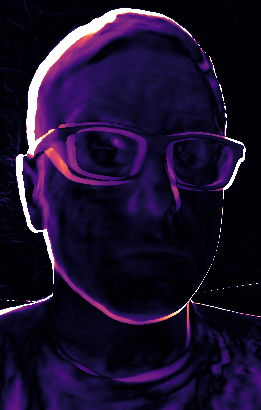}{depth difference}\hspace{-1pt}
    	\figcell{0.226}{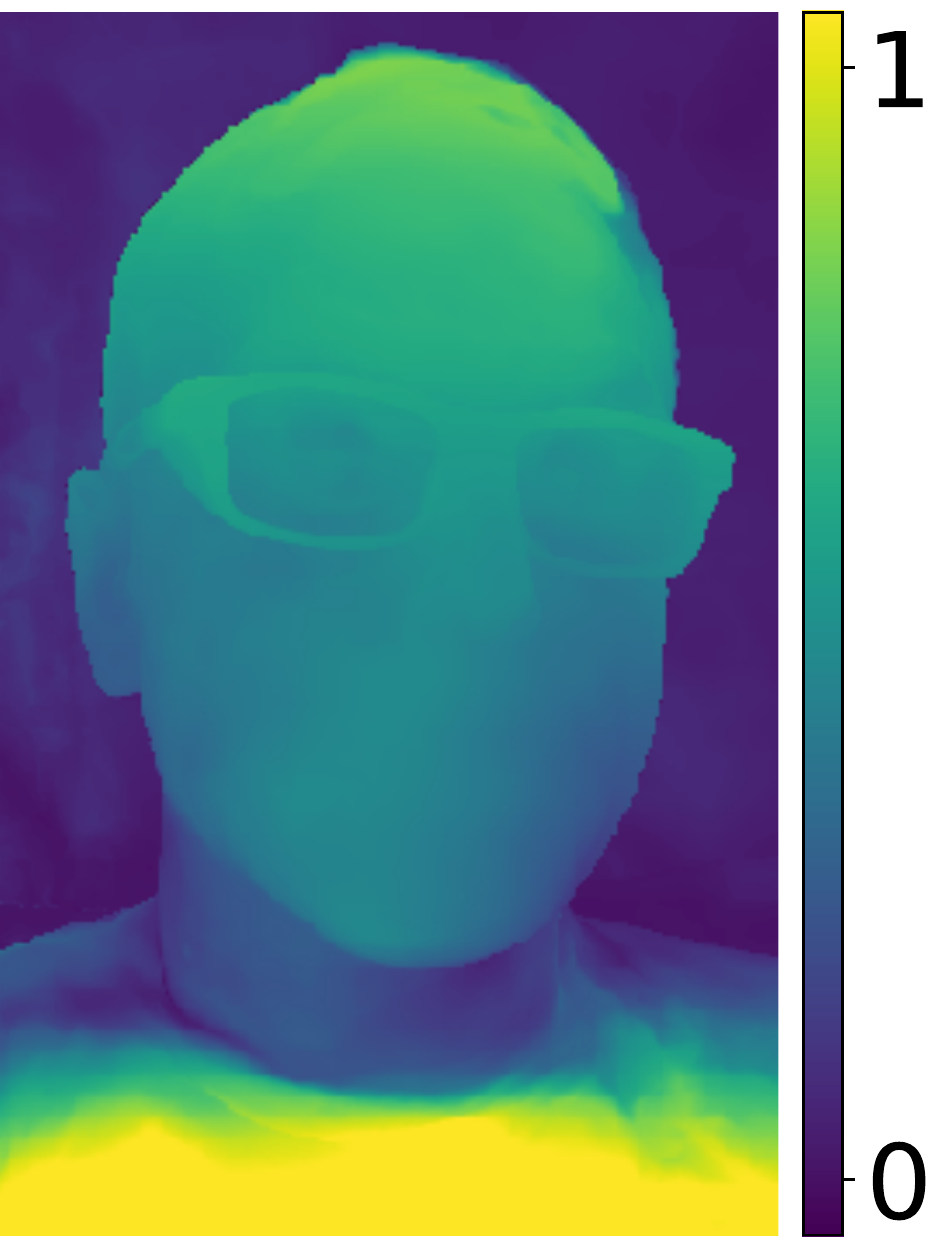}{deformation mag (cm)}
	\end{subfigure}
    \vspace{-8pt} 
	\caption{
    Users move even when trying not to. Here we visualize the depth difference and deformation magnitude between the template and an observation.
	}
  \label{fig:warp_magnitude}
  \vspace{-5pt} 

\end{figure}

%% file: f0x_toby.tex
\fboxsep=0pt 
\fboxrule=0.4pt 

\begin{figure}[t]
	\centering
	\begin{subfigure}{1.0\columnwidth}
    	\centering
    	\figcelltb{0.244}{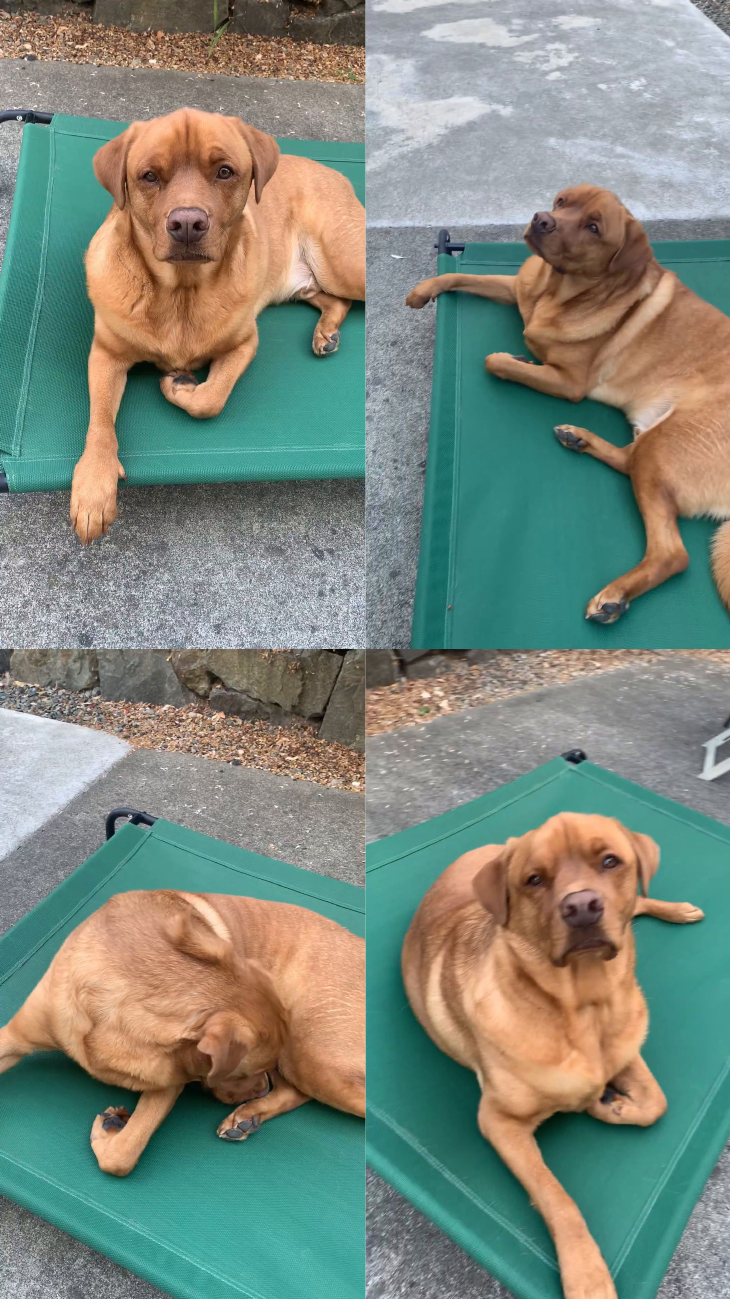}{example inputs}{clip,trim=0 40 0 20}\hspace{-1.0pt}
    	\figcelltb{0.242}{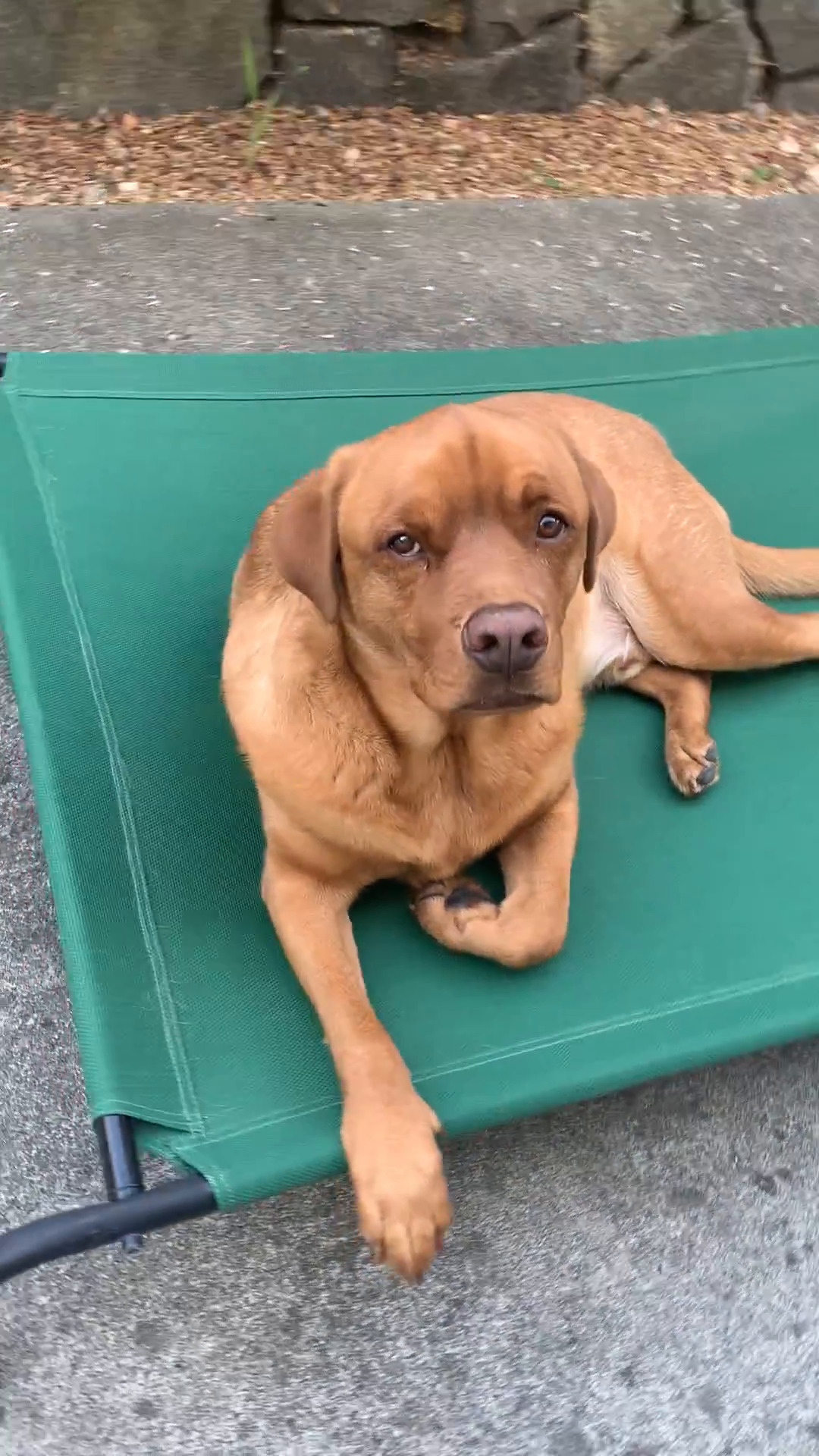}{ground truth}{clip,trim=0 40 0 250}\hspace{-1.0pt}
    	\figcelltb{0.242}{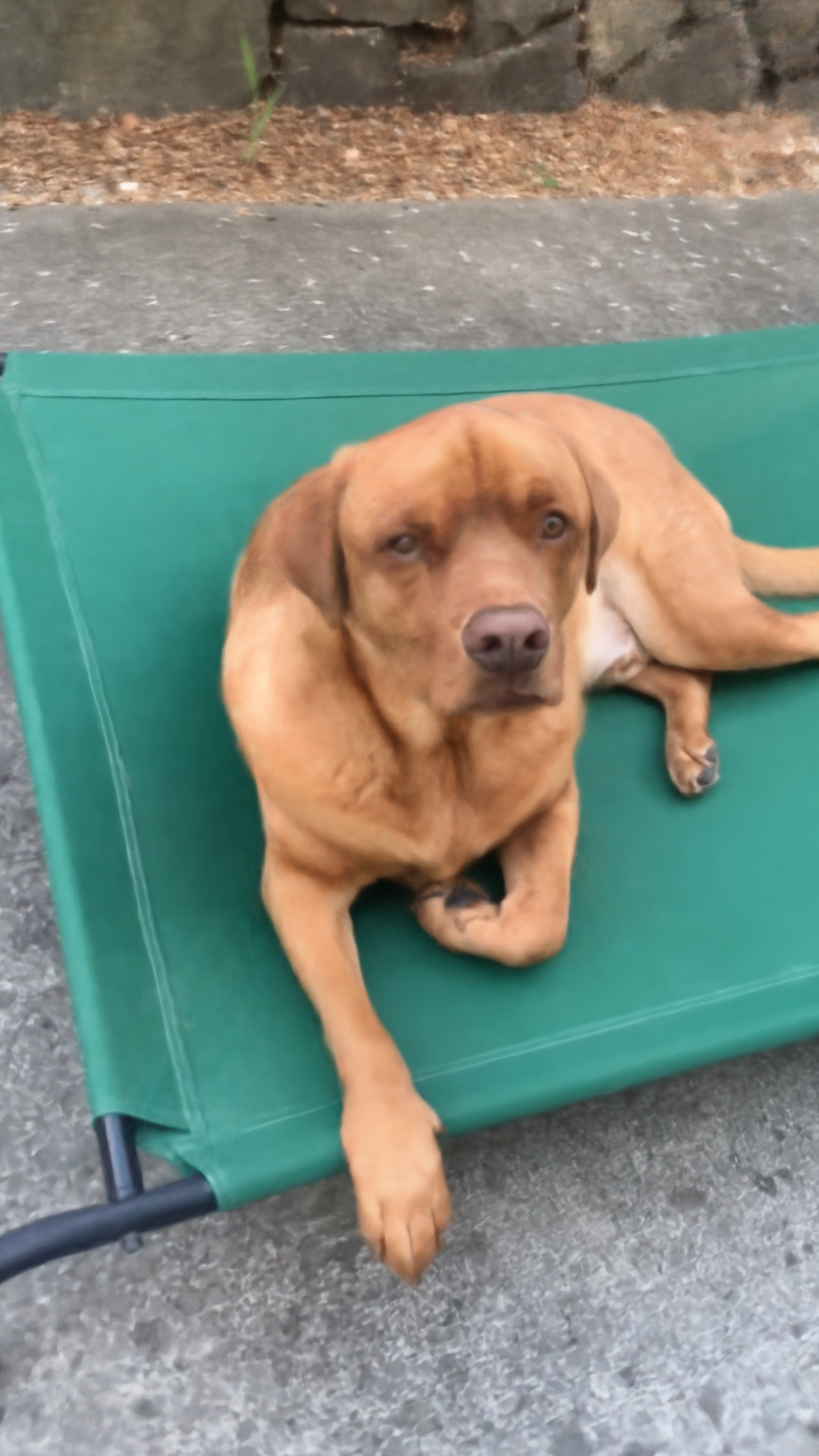}{rendered color}{clip,trim=0 40 0 250}\hspace{-1.0pt}
    	\figcelltb{0.242}{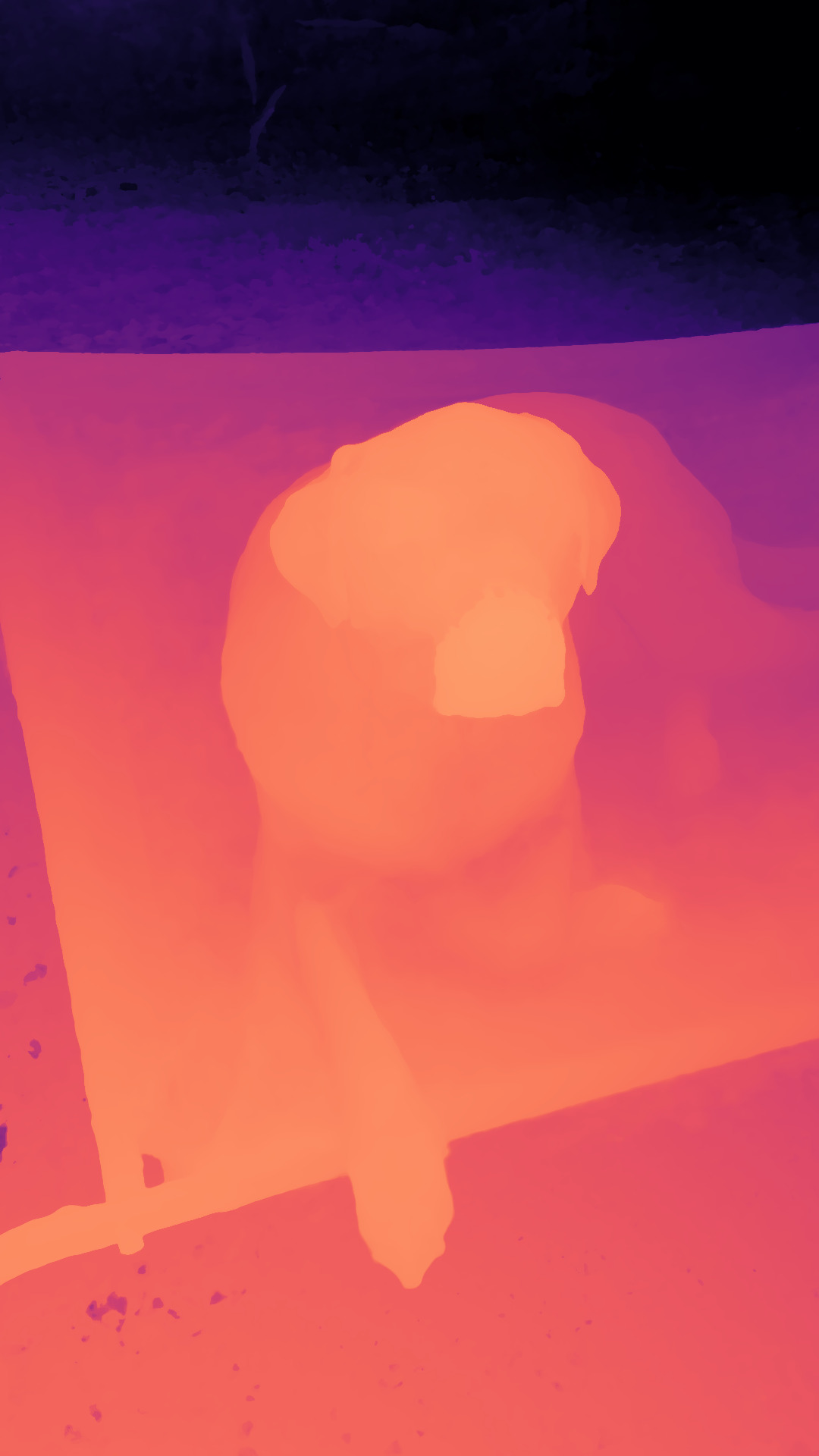}{rendered depth}{clip,trim=0 40 0 250}
	\end{subfigure}
	\vspace{-8pt}
	\caption{
	Not relying on domain specific priors enables our method to reconstruct any
	deformable object. In this case, the dog fails to stay still, yet we recover an accurate model.
	}
   \label{fig:qualitative_toby}
   \vspace{-8pt}
\end{figure}

%% file: fa0_arch_deformation.tex
\begin{figure}[t]
    \centering
    \includegraphics*[width=0.8\linewidth]{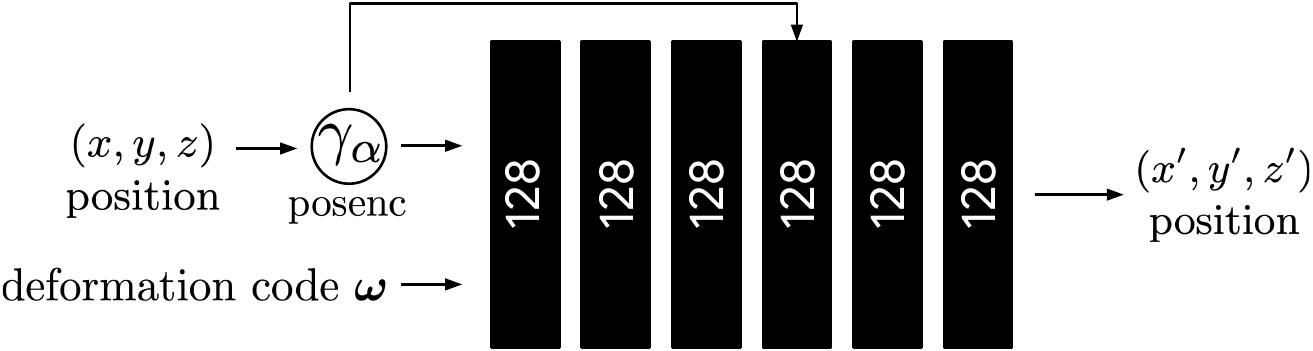}
    \caption{A diagram of our deformation network. The deformation network takes a position encoded position $\gamma_\alpha(\mat{x})$ using our coarse-to-fine annealing parameterized by $\alpha$, along with a deformation code $\mat{\omega}$ and outputs a deformed position $\mat{x}'$. The architecture is identical for all of our experiments.}
    \label{fig:deformation_network}
\end{figure}

%% file: fa0_arch_template.tex
\begin{figure}[t]
    \includegraphics*[width=\linewidth]{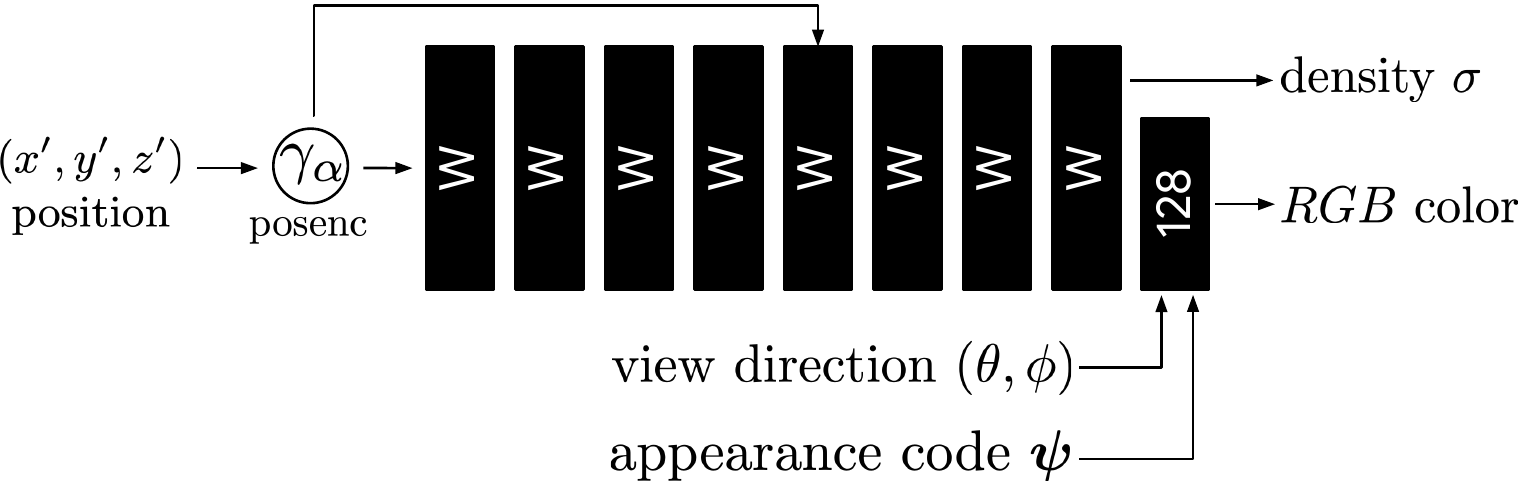}
    \caption{A diagram of the canonical NeRF network. Our network is identical to the original NeRF MLP, except we provide an appearance latent code $\mat{\psi}$ along with the view direction to allow modulating the appearance as in the NeRF-A model of~\cite{martinbrualla2020nerfw}. The width $W$ of the network is defined according to \tabref{table:hyper_params}.}
    \label{fig:template_network}
\end{figure}

%% file: ta0_ssim_table.tex
\renewcommand{\tablefirst}[0]{\cellcolor{myred}}
\renewcommand{\tablesecond}[0]{\cellcolor{myorange}}
\renewcommand{\tablethird}[0]{\cellcolor{myyellow}}

\begin{table*}[t]
\caption{
SSIM and LPIPS metrics on validation captures against baselines and ablations of our system, we color code each row as \colorbox{myred}{\textbf{best}}, \colorbox{myorange}{\textbf{second best}}, and \colorbox{myyellow}{\textbf{third best}}. Please see the main text for PSNR.
\label{table:ssim_table}
}
\centering
\resizebox{\linewidth}{!}{
\setlength{\tabcolsep}{2pt}
\input{t0x_quantitative_table_contents_ssim_0909-crop}
}

\end{table*}

%% file: t0x_quantitative_table_contents_ssim_0909-crop.tex

\begin{tabular}{l||cc|cc|cc|cc|cc|cc|cc||cc|cc|cc|cc|cc}

\toprule

& \multicolumn{ 14 }{c||}{\makecell{{\small Quasi-Static }}}
& \multicolumn{ 10 }{c}{\makecell{{\small Dynamic }}}
\\ \hline

& \multicolumn{ 2 }{c}{
  \makecell{
  \textsc{\small Glasses }
\\(78 images)
  }
}
& \multicolumn{ 2 }{c}{
  \makecell{
  \textsc{\small Beanie }
\\(74 images)
  }
}
& \multicolumn{ 2 }{c}{
  \makecell{
  \textsc{\small Curls }
\\(57 images)
  }
}
& \multicolumn{ 2 }{c}{
  \makecell{
  \textsc{\small Kitchen }
\\(40 images)
  }
}
& \multicolumn{ 2 }{c}{
  \makecell{
  \textsc{\small Lamp }
\\(55 images)
  }
}
& \multicolumn{ 2 }{c|}{
  \makecell{
  \textsc{\small Toby Sit }
\\(308 images)
  }
}
& \multicolumn{ 2 }{c||}{
  \makecell{
  \textsc{\small Mean }
  }
}
& \multicolumn{ 2 }{c}{
  \makecell{
  \textsc{\small Drinking }
\\(193 images)
  }
}
& \multicolumn{ 2 }{c}{
  \makecell{
  \textsc{\small Tail }
\\(238 images)
  }
}
& \multicolumn{ 2 }{c}{
  \makecell{
  \textsc{\small Badminton }
\\(356 images)
  }
}
& \multicolumn{ 2 }{c|}{
  \makecell{
  \textsc{\small Broom }
\\(197 images)
  }
}
& \multicolumn{ 2 }{c}{
  \makecell{
  \textsc{\small Mean }
  }
}
\\

& \multicolumn{1}{c}{ \footnotesize PSNR$\uparrow$ }
& \multicolumn{1}{c}{ \footnotesize LPIPS$\downarrow$ }
& \multicolumn{1}{c}{ \footnotesize PSNR$\uparrow$ }
& \multicolumn{1}{c}{ \footnotesize LPIPS$\downarrow$ }
& \multicolumn{1}{c}{ \footnotesize PSNR$\uparrow$ }
& \multicolumn{1}{c}{ \footnotesize LPIPS$\downarrow$ }
& \multicolumn{1}{c}{ \footnotesize PSNR$\uparrow$ }
& \multicolumn{1}{c}{ \footnotesize LPIPS$\downarrow$ }
& \multicolumn{1}{c}{ \footnotesize PSNR$\uparrow$ }
& \multicolumn{1}{c}{ \footnotesize LPIPS$\downarrow$ }
& \multicolumn{1}{c}{ \footnotesize PSNR$\uparrow$ }
& \multicolumn{1}{c|}{ \footnotesize LPIPS$\downarrow$ }
& \multicolumn{1}{c}{ \footnotesize PSNR$\uparrow$ }
& \multicolumn{1}{c||}{ \footnotesize LPIPS$\downarrow$ }
& \multicolumn{1}{c}{ \footnotesize PSNR$\uparrow$ }
& \multicolumn{1}{c}{ \footnotesize LPIPS$\downarrow$ }
& \multicolumn{1}{c}{ \footnotesize PSNR$\uparrow$ }
& \multicolumn{1}{c}{ \footnotesize LPIPS$\downarrow$ }
& \multicolumn{1}{c}{ \footnotesize PSNR$\uparrow$ }
& \multicolumn{1}{c}{ \footnotesize LPIPS$\downarrow$ }
& \multicolumn{1}{c}{ \footnotesize PSNR$\uparrow$ }
& \multicolumn{1}{c|}{ \footnotesize LPIPS$\downarrow$ }
& \multicolumn{1}{c}{ \footnotesize PSNR$\uparrow$ }
& \multicolumn{1}{c}{ \footnotesize LPIPS$\downarrow$ }
\\
\hline

  NeRF~\cite{mildenhall2020nerf}
  &$.619$
  &$.474$
  
  &$.580$
  &$.583$
  
  &$.504$
  &$.616$
  
  &$.695$
  &$.434$
  
  &$.656$
  &$.444$
  
  &$.793$
  &$.463$
  
  &$.641$
  &$.502$
  
  &$.619$
  &$.397$
  
  &$.676$
  &$.571$
  
  &$.771$
  &$.392$
  
  &$.643$
  &$.667$
  
  &$.677$
  &$.506$
  
  \\
  NeRF + latent
  &$.695$
  &$.463$
  
  &$.687$
  &$.535$
  
  &$.619$
  &$.539$
  
  &$.746$
  &$.403$
  
  &$.735$
  &$.386$
  
  &$.798$
  &$.385$
  
  &$.713$
  &$.452$
  
  &$.855$
  &$.233$
  
  &\tablethird$.800$
  &$.404$
  
  &$.850$
  &$.308$
  
  &$.688$
  &$.576$
  
  &$.798$
  &$.380$
  
  \\
  Neural Volumes~\cite{lombardi2019neural}
  &$.503$
  &$.616$
  
  &$.562$
  &$.595$
  
  &$.538$
  &$.588$
  
  &$.609$
  &$.569$
  
  &$.563$
  &$.533$
  
  &$.583$
  &$.473$
  
  &$.560$
  &$.562$
  
  &$.771$
  &$.198$
  
  &$.503$
  &$.559$
  
  &$.219$
  &$.516$
  
  &$.515$
  &$.544$
  
  &$.502$
  &$.454$
  
  \\
  NSFF\textsuperscript{\textdagger}
  &$.678$
  &$.407$
  
  &$.760$
  &$.402$
  
  &$.621$
  &$.432$
  
  &$.780$
  &$.317$
  
  &$.807$
  &$.239$
  
  &\tablefirst$.913$
  &$.208$
  
  &$.760$
  &$.334$
  
  &\tablefirst$.964$
  &\tablefirst$.0803$
  
  &\tablefirst$.917$
  &$.245$
  
  &$.840$
  &$.205$
  
  &\tablefirst$.893$
  &\tablefirst$.202$
  
  &\tablefirst$.904$
  &$.183$
  
  \\
  $\gamma(t)$ + Trans\textsuperscript{\textdagger}~\cite{li2020neural}
  &$.781$
  &$.354$
  
  &$.737$
  &$.471$
  
  &$.732$
  &$.426$
  
  &$.823$
  &$.344$
  
  &$.836$
  &$.283$
  
  &\tablesecond$.870$
  &$.420$
  
  &$.796$
  &$.383$
  
  &\tablesecond$.910$
  &$.151$
  
  &\tablesecond$.882$
  &$.391$
  
  &\tablefirst$.927$
  &$.221$
  
  &\tablesecond$.750$
  &$.627$
  
  &\tablesecond$.867$
  &$.347$
  
  \\ \hline
  Ours ($\lambda=0.01$)
  &$.826$
  &\tablefirst$.305$
  
  &\tablethird$.786$
  &\tablethird$.391$
  
  &\tablesecond$.842$
  &\tablethird$.319$
  
  &\tablefirst$.878$
  &\tablethird$.280$
  
  &\tablesecond$.888$
  &\tablethird$.232$
  
  &$.806$
  &\tablesecond$.159$
  
  &\tablesecond$.838$
  &\tablefirst$.281$
  
  &$.894$
  &$.0872$
  
  &$.754$
  &\tablefirst$.161$
  
  &\tablesecond$.926$
  &\tablefirst$.130$
  
  &$.674$
  &\tablesecond$.245$
  
  &$.812$
  &\tablefirst$.156$
  
  \\
  Ours ($\lambda=0.001$)
  &\tablefirst$.840$
  &\tablesecond$.307$
  
  &\tablesecond$.805$
  &\tablesecond$.391$
  
  &\tablefirst$.846$
  &\tablefirst$.312$
  
  &$.863$
  &\tablesecond$.279$
  
  &\tablethird$.886$
  &\tablefirst$.230$
  
  &$.805$
  &\tablethird$.174$
  
  &\tablefirst$.841$
  &\tablesecond$.282$
  
  &$.881$
  &$.0962$
  
  &$.731$
  &\tablethird$.175$
  
  &\tablethird$.922$
  &\tablesecond$.132$
  
  &$.605$
  &\tablethird$.270$
  
  &$.785$
  &\tablesecond$.168$
  
  \\
  No elastic
  &$.809$
  &$.317$
  
  &\tablefirst$.824$
  &\tablefirst$.382$
  
  &$.830$
  &$.322$
  
  &$.851$
  &$.290$
  
  &\tablefirst$.889$
  &\tablesecond$.230$
  
  &\tablethird$.821$
  &$.257$
  
  &\tablethird$.837$
  &$.300$
  
  &$.890$
  &\tablethird$.0863$
  
  &$.735$
  &\tablesecond$.174$
  
  &$.919$
  &\tablethird$.132$
  
  &$.593$
  &$.287$
  
  &$.784$
  &\tablethird$.170$
  
  \\
  No coarse-to-fine
  &\tablesecond$.828$
  &\tablethird$.312$
  
  &$.771$
  &$.408$
  
  &\tablethird$.841$
  &$.321$
  
  &\tablesecond$.877$
  &\tablefirst$.277$
  
  &$.867$
  &$.242$
  
  &$.807$
  &$.244$
  
  &$.832$
  &$.301$
  
  &$.892$
  &$.0960$
  
  &$.763$
  &$.257$
  
  &$.912$
  &$.151$
  
  &$.695$
  &$.406$
  
  &\tablethird$.815$
  &$.228$
  
  \\
  No SE3
  &$.823$
  &$.314$
  
  &$.782$
  &$.401$
  
  &$.839$
  &\tablesecond$.317$
  
  &\tablethird$.870$
  &$.282$
  
  &$.872$
  &$.235$
  
  &$.810$
  &$.206$
  
  &$.833$
  &\tablethird$.293$
  
  &\tablethird$.895$
  &$.0867$
  
  &$.715$
  &$.191$
  
  &$.899$
  &$.156$
  
  &$.599$
  &$.276$
  
  &$.777$
  &$.177$
  
  \\
  Ours (base)
  &\tablethird$.828$
  &$.319$
  
  &$.737$
  &$.456$
  
  &$.818$
  &$.345$
  
  &$.829$
  &$.323$
  
  &$.851$
  &$.254$
  
  &$.792$
  &$.184$
  
  &$.809$
  &$.314$
  
  &$.894$
  &$.127$
  
  &$.768$
  &$.298$
  
  &$.894$
  &$.173$
  
  &\tablethird$.695$
  &$.503$
  
  &$.813$
  &$.275$
  
  \\
  No BG Loss
  &$.779$
  &$.317$
  
  &$.758$
  &$.395$
  
  &$.696$
  &$.371$
  
  &$.844$
  &$.290$
  
  &$.806$
  &$.260$
  
  &$.775$
  &\tablefirst$.145$
  
  &$.776$
  &$.296$
  
  &$.893$
  &\tablesecond$.0856$
  
  &$.719$
  &$.210$
  
  &$.875$
  &$.161$
  
  &$.593$
  &$.330$
  
  &$.770$
  &$.196$
  
  \\
\bottomrule

\end{tabular}


%% file: f0x_gazefollow.tex
\fboxsep=0pt 
\fboxrule=0.4pt 

\begin{figure}[t]
	\begin{subfigure}{0.307\columnwidth}
    	\centering
    	\figcellb{0.99}{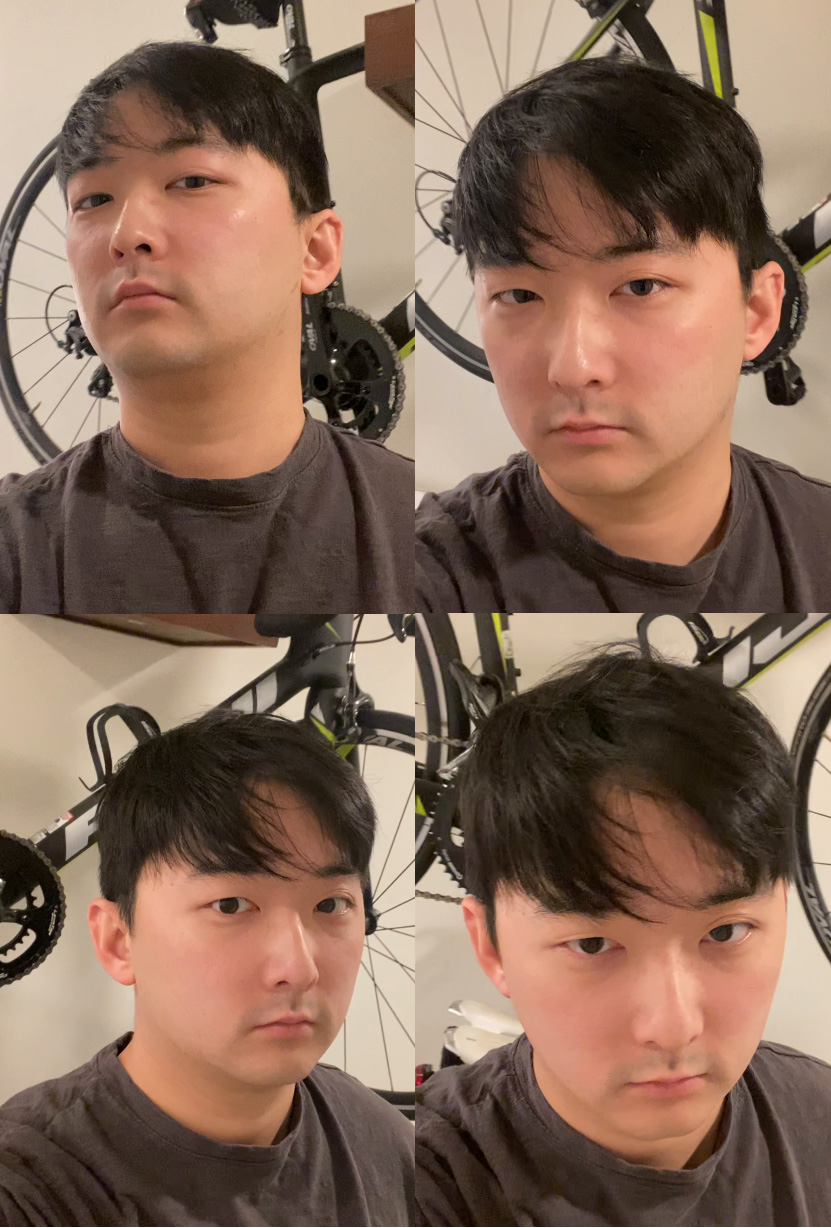}{}
    	\caption*{example inputs}
	\end{subfigure}
	\begin{subfigure}{0.69\columnwidth}
    	\centering
    	\begin{subfigure}{1.0\columnwidth}
        	\centering
        	\figcellb{0.32}{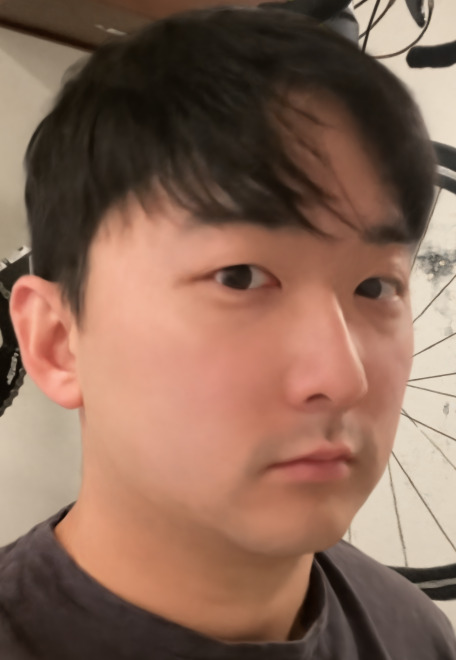}{}\hspace{-1.2pt}
        	\figcellb{0.32}{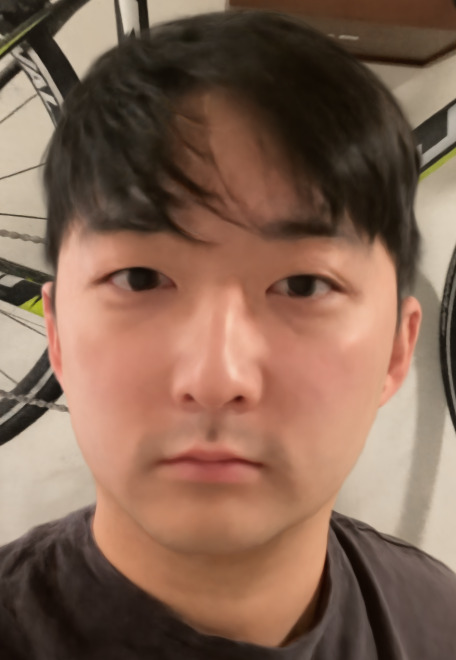}{}\hspace{-1.2pt}
        	\figcellb{0.32}{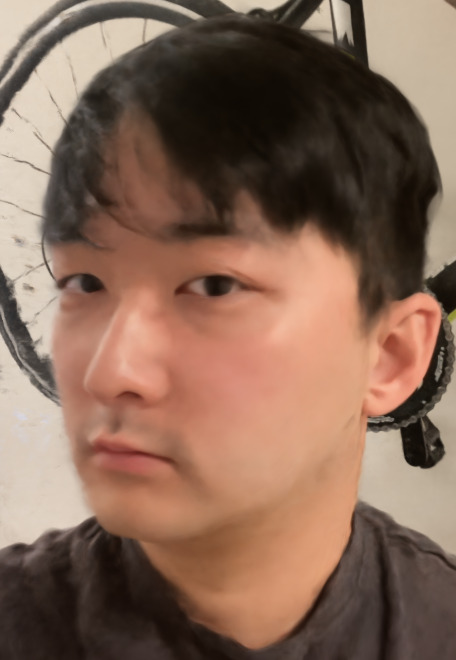}{}
    	\end{subfigure}
    	\begin{subfigure}{1.0\columnwidth}
        	\centering
        	\figcellb{0.32}{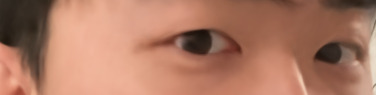}{}\hspace{-1.2pt}
        	\figcellb{0.32}{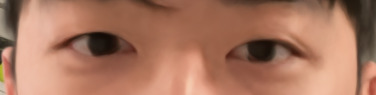}{}\hspace{-1.2pt}
        	\figcellb{0.32}{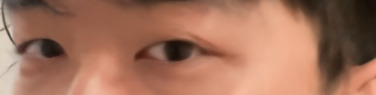}{}
    	\end{subfigure}
    	\begin{subfigure}{1.0\columnwidth}
        	\centering
        	\figcellb{0.32}{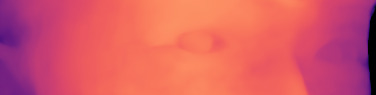}{}\hspace{-1.2pt}
        	\figcellb{0.32}{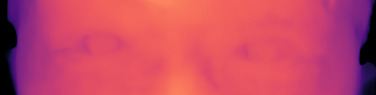}{}\hspace{-1.2pt}
        	\figcellb{0.32}{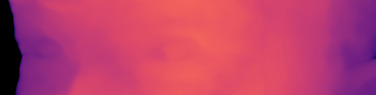}{}
    	\end{subfigure}
    	\caption*{novel views for same deformation code}
	\end{subfigure}
	\vspace{-8pt}
	\caption{If the user's gaze consistently follows the camera, the reconstructed \emph{nerfie} represents the user's gaze as geometry, akin to the Hollow-Face illusion~\cite{gregory1970intelligent}. This is apparent in the depth map and makes the reconstructed model appear as if they are looking at the camera even when the geometry is fixed. }
   \label{fig:gazefollow}
   \vspace{-10pt}
\end{figure}

%% file: fa0_large_motion.tex
\fboxsep=0pt 
\fboxrule=0.4pt 

\begin{figure}[t]
	\begin{subfigure}{0.99\columnwidth}
    	\centering
    	\figcelltb{0.15}{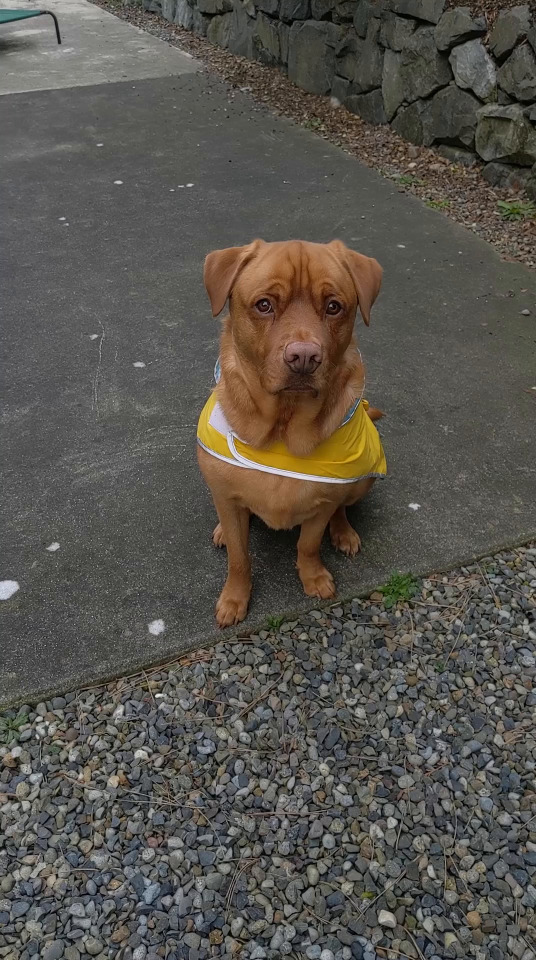}{}{}
    	\figcelltb{0.15}{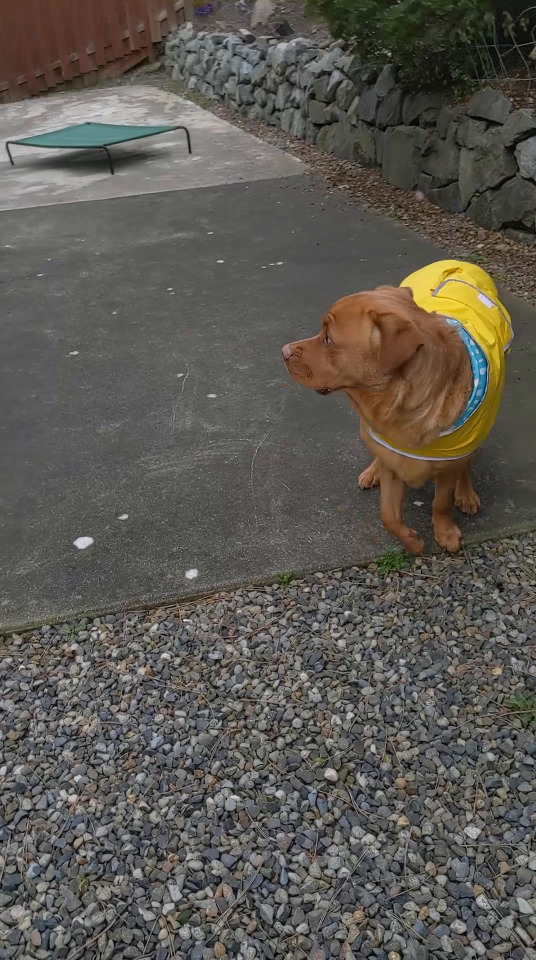}{}{}
    	\figcelltb{0.15}{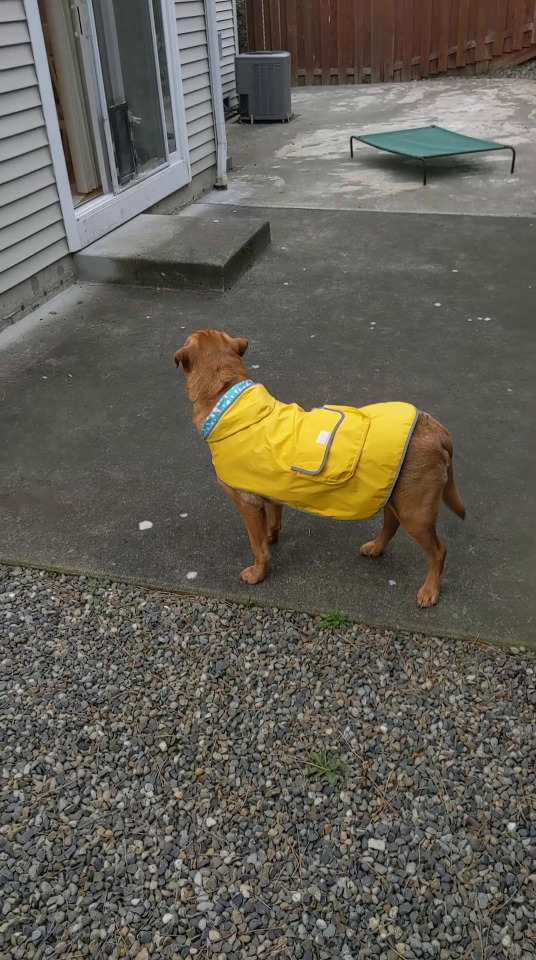}{}{}
    	\figcelltb{0.15}{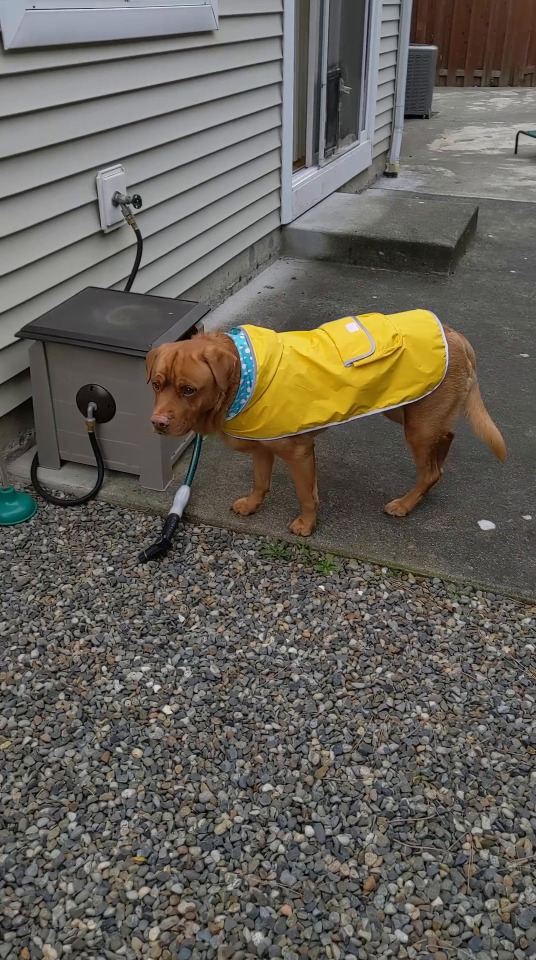}{}{}
    	\figcelltb{0.15}{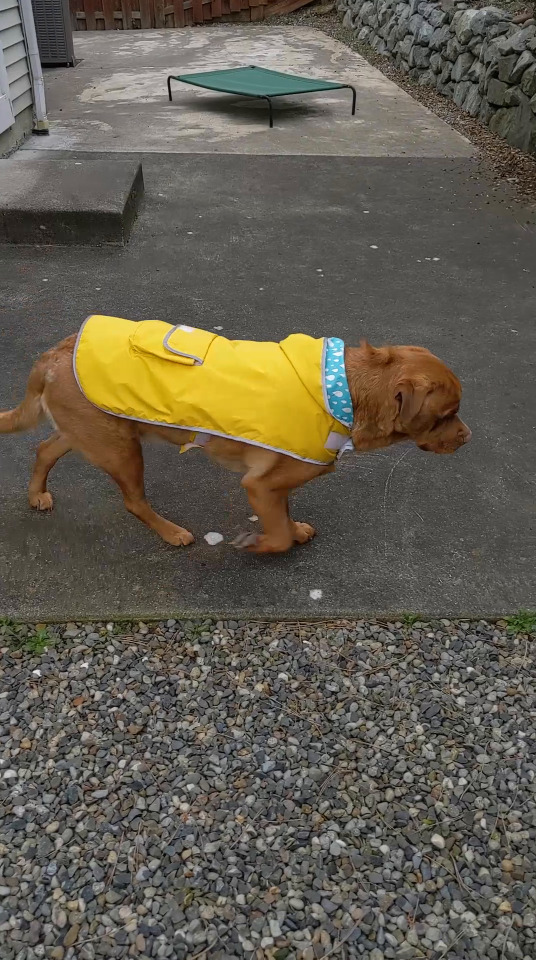}{}{}
    	\figcelltb{0.15}{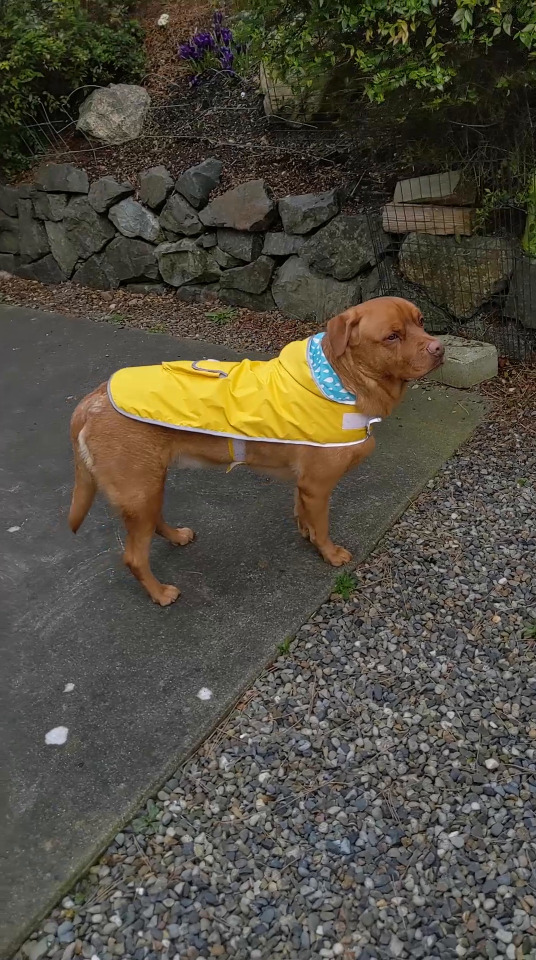}{}{}
    	\caption{input rgb}
	\end{subfigure}
	\begin{subfigure}{0.99\columnwidth}
    	\centering
    	\figcelltb{0.15}{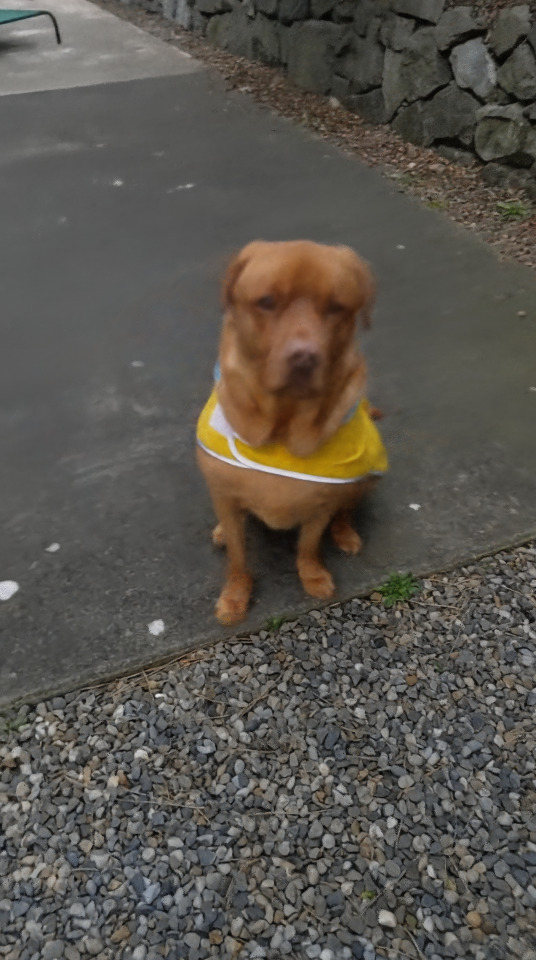}{}{}
    	\figcelltb{0.15}{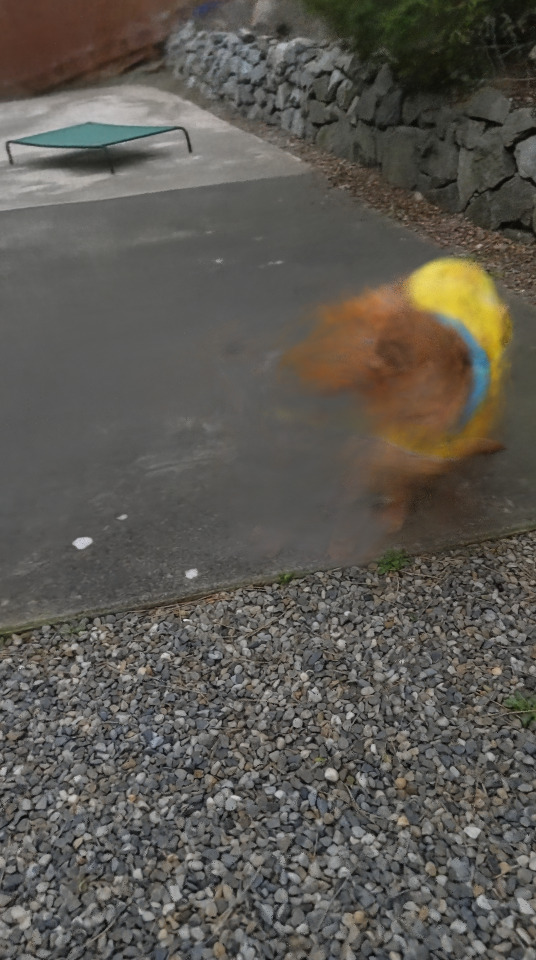}{}{}
    	\figcelltb{0.15}{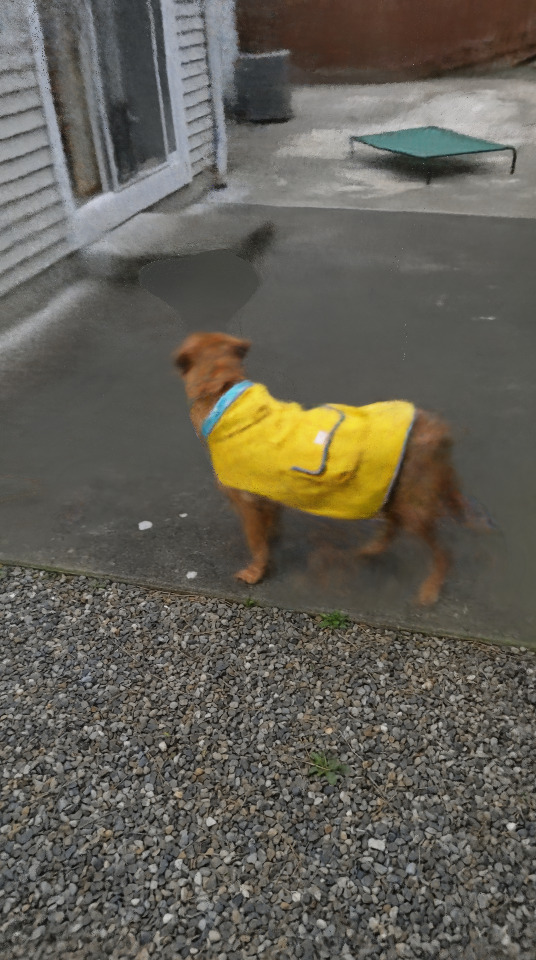}{}{}
    	\figcelltb{0.15}{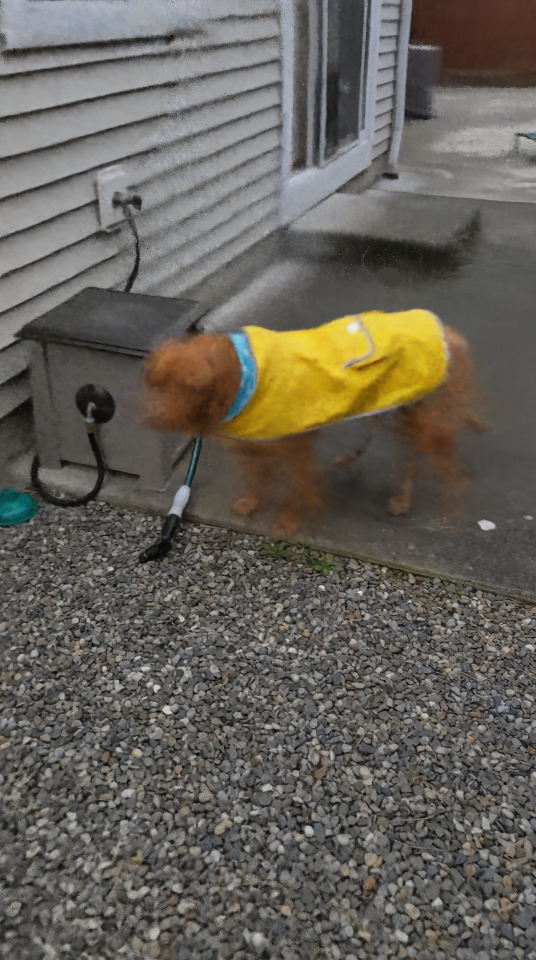}{}{}
    	\figcelltb{0.15}{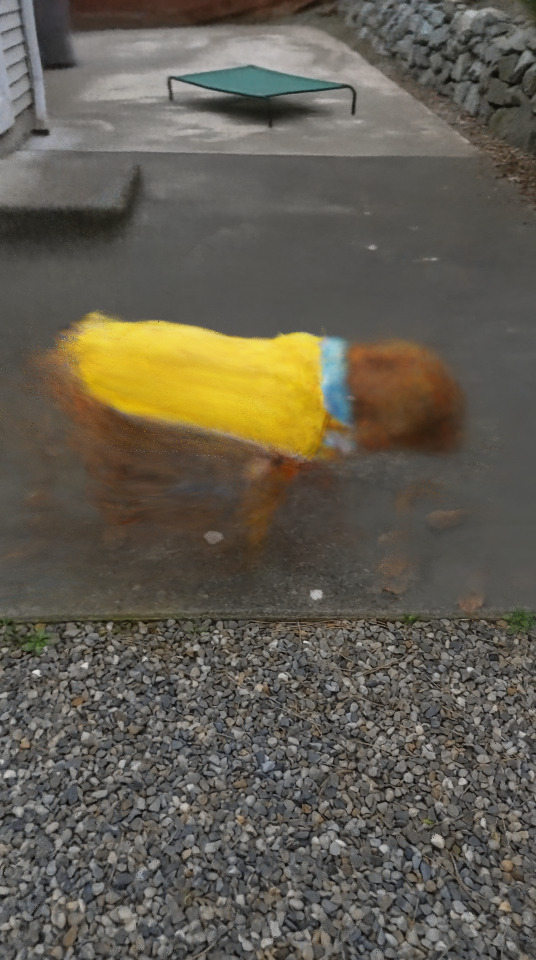}{}{}
    	\figcelltb{0.15}{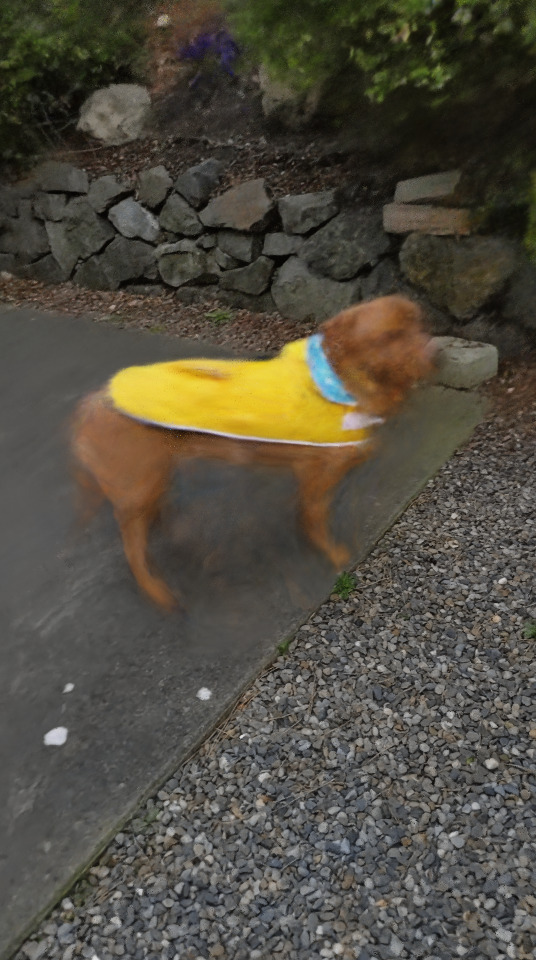}{}{}
    	\caption{rendered rgb}
	\end{subfigure}
	\begin{subfigure}{0.99\columnwidth}
    	\centering
    	\figcelltb{0.15}{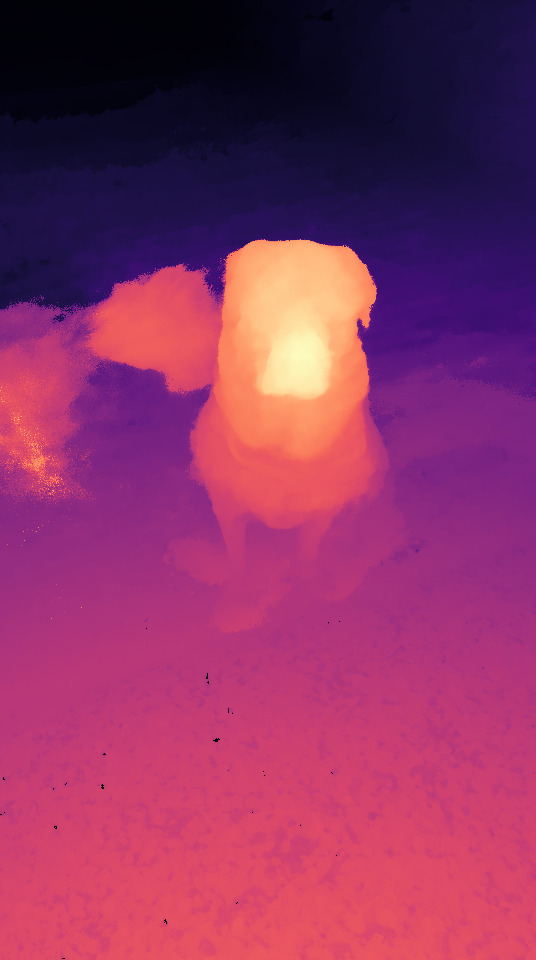}{}{}
    	\figcelltb{0.15}{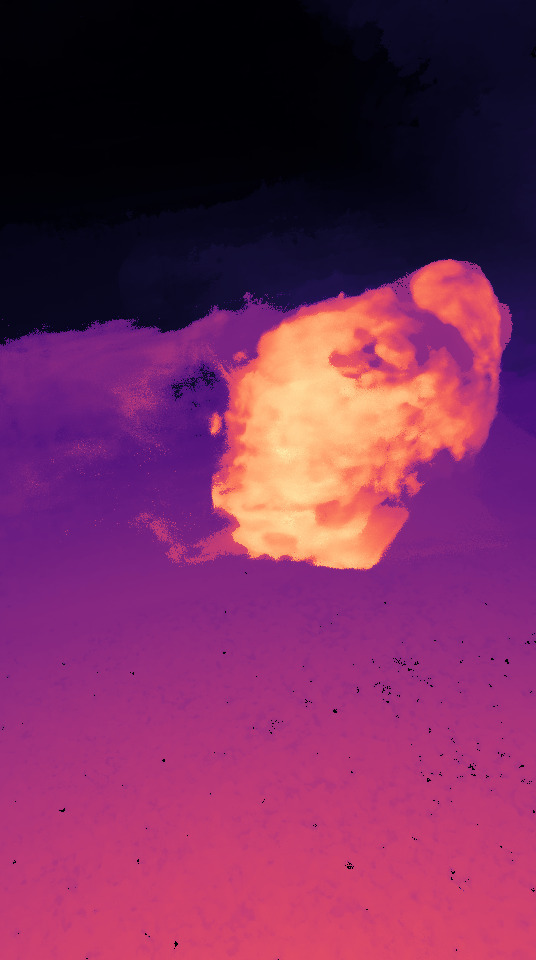}{}{}
    	\figcelltb{0.15}{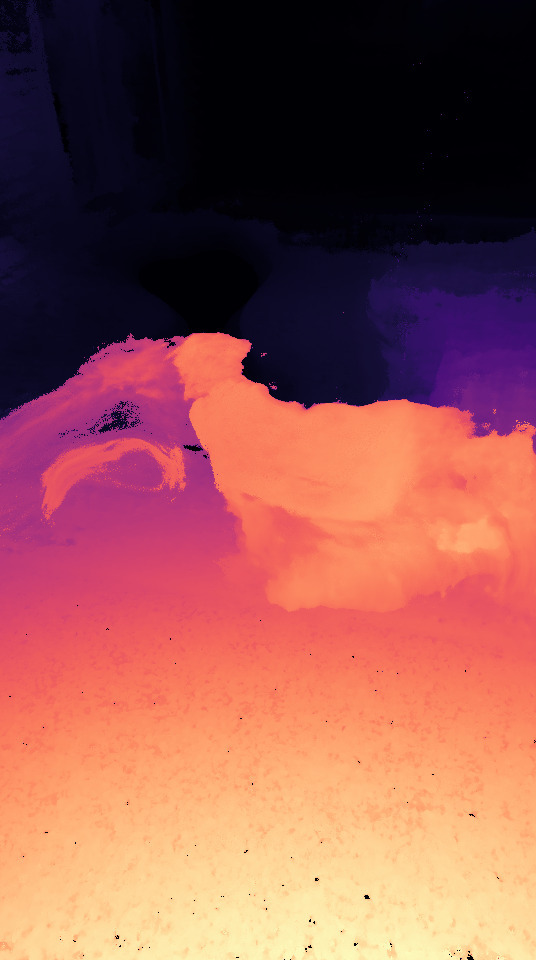}{}{}
    	\figcelltb{0.15}{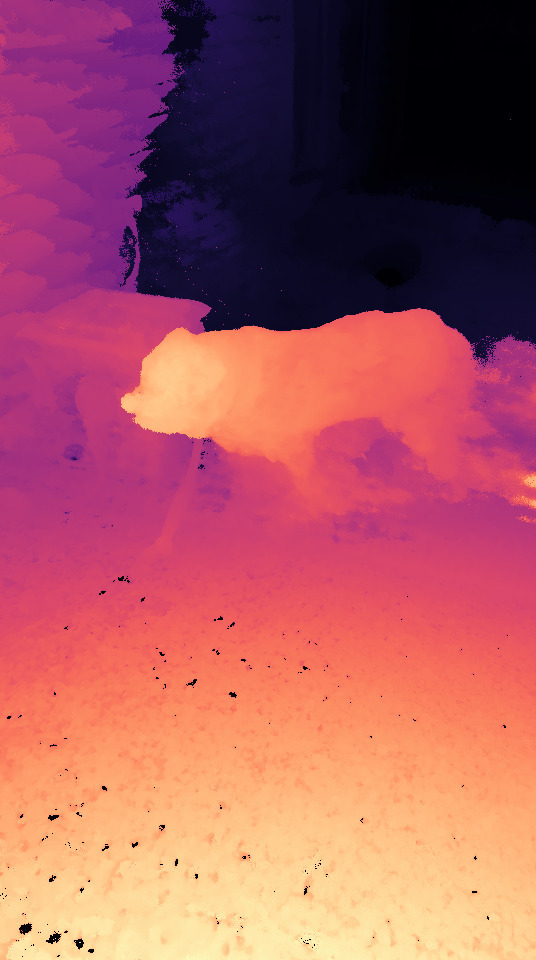}{}{}
    	\figcelltb{0.15}{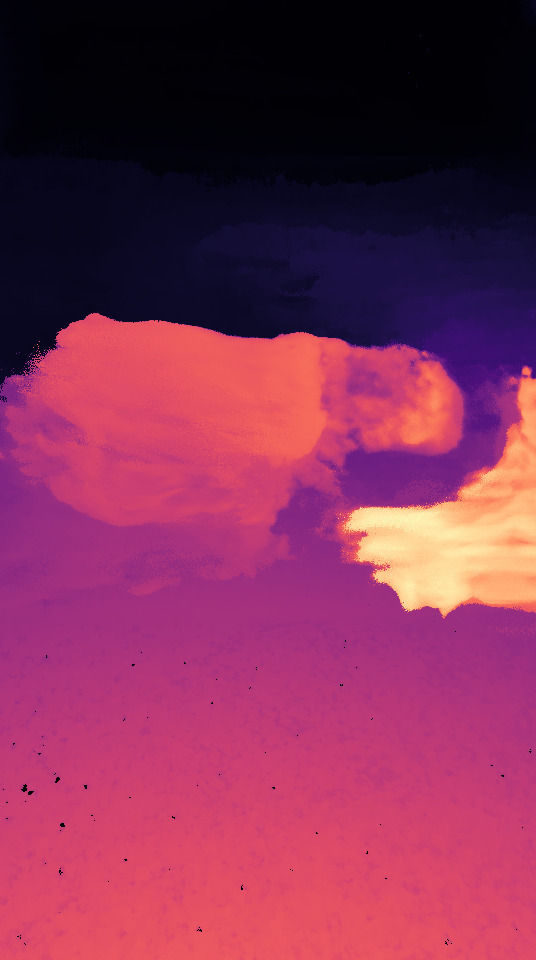}{}{}
    	\figcelltb{0.15}{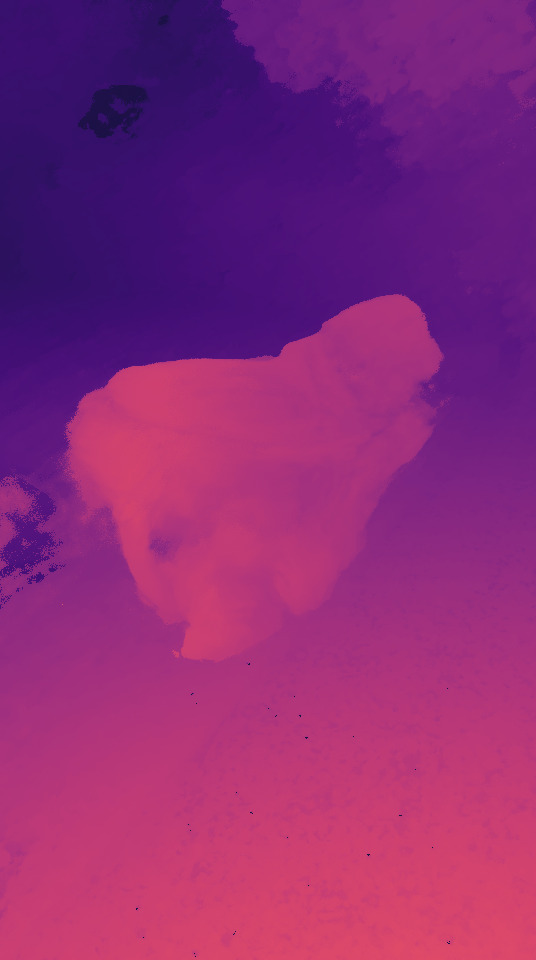}{}{}
    	\caption{rendered depth}
	\end{subfigure}
	\vspace{-5pt}
	\caption{This example shows Toby the dog moving around freely, showcasing two limitations of our method. (1) \emph{Rapid motion}: Because Toby moves quite fast, the camera only sees him in certain poses for a short amount of time, resulting in a sparse set of observations for certain poses. This can make those poses under-constrained. (2) \emph{Orientation flips}: Toby wanders back and forth, showing different sides of his body. Depending on which orientation Toby is modeled as in the template, it is difficult for the deformation field to predict a flipped orientation.}
   \label{fig:large_movement}
   \vspace{-10pt}

\end{figure}

%% file: fa0_2d_toy_train.tex
\newcommand{\spacemantrain}[1]{\includegraphics[height=1.95cm,clip]{figures/2d_toy/training/#1}}

\begin{figure}[t!]
    \centering
    \setlength{\tabcolsep}{0.5pt}
    \renewcommand{\arraystretch}{0.25}
    \begin{tabular}{cccc}
        \spacemantrain{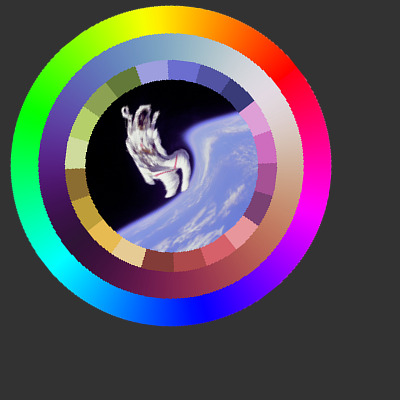} &
        \spacemantrain{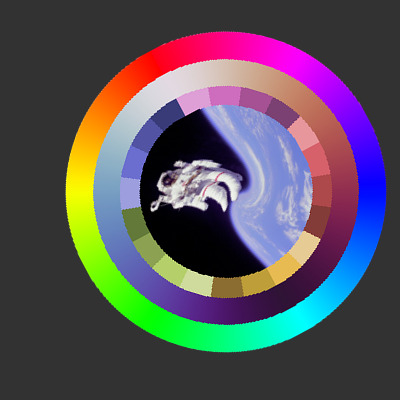} &
        \spacemantrain{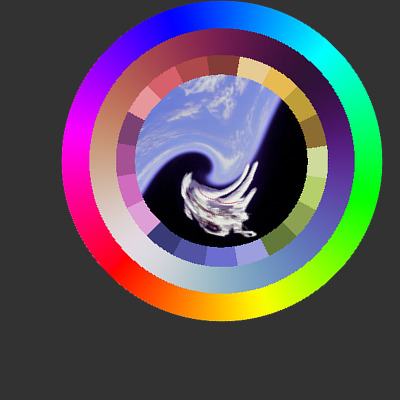} &
        \spacemantrain{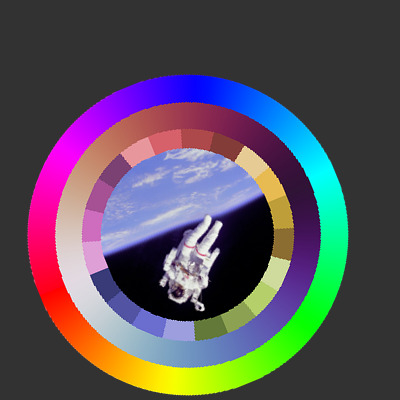} \\
        \spacemantrain{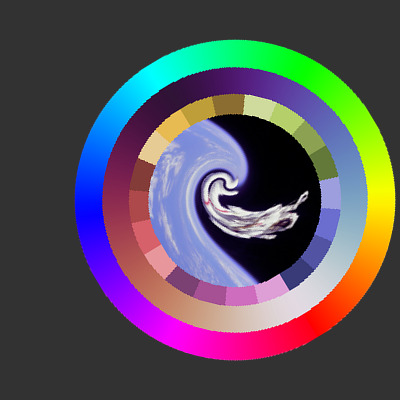} &
        \spacemantrain{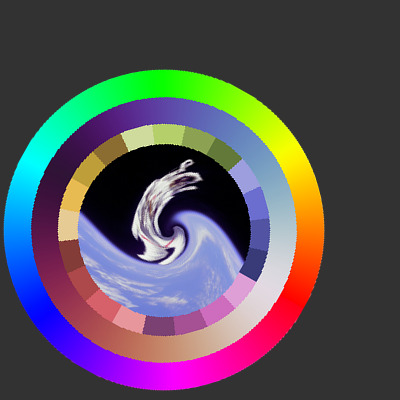} &
        \spacemantrain{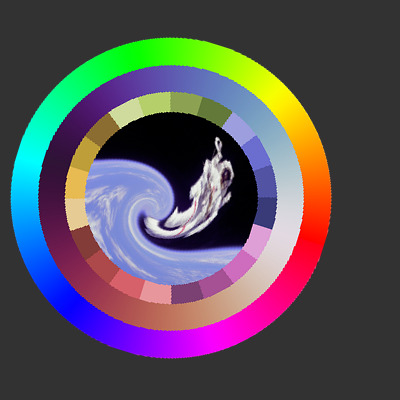} &
        \spacemantrain{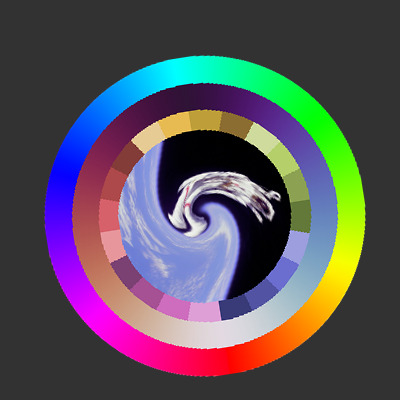} \\
        \spacemantrain{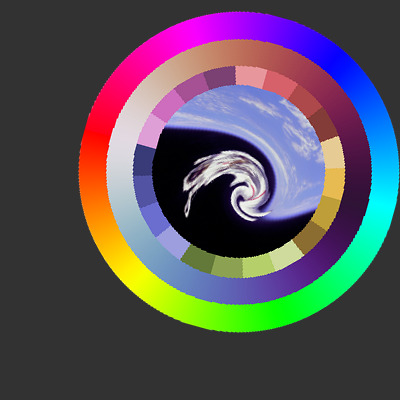} &
        \spacemantrain{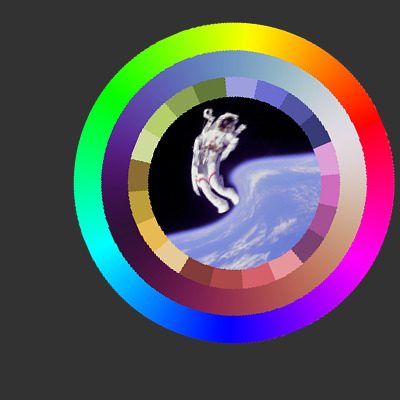} &
        \spacemantrain{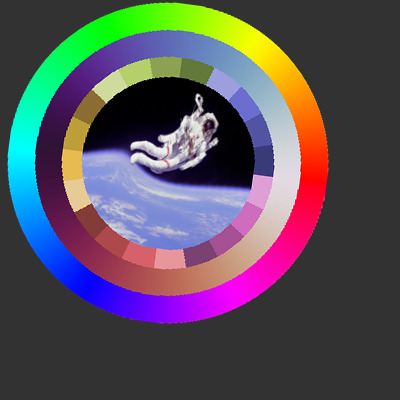} &
        \spacemantrain{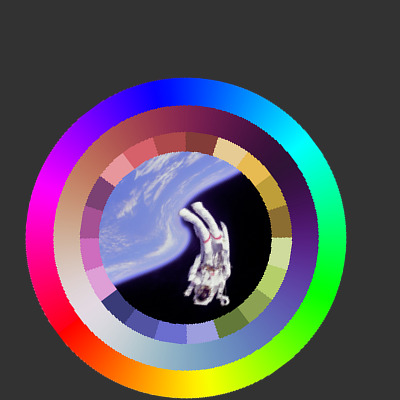} \\
        \spacemantrain{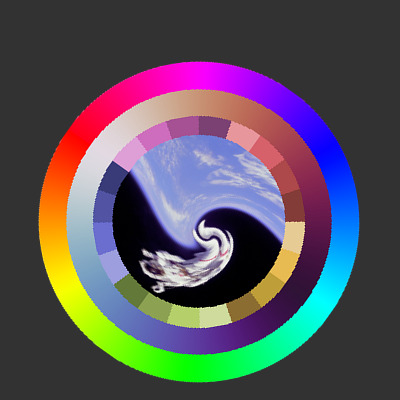} &
        \spacemantrain{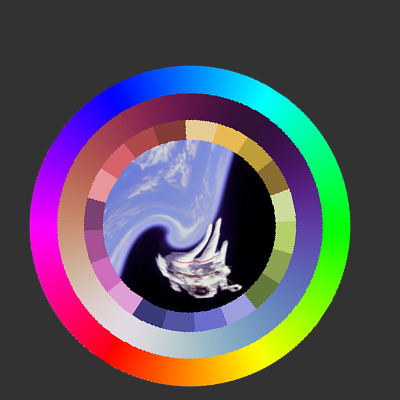} &
        \spacemantrain{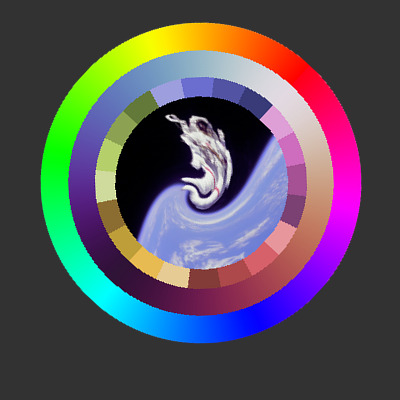} &
        \spacemantrain{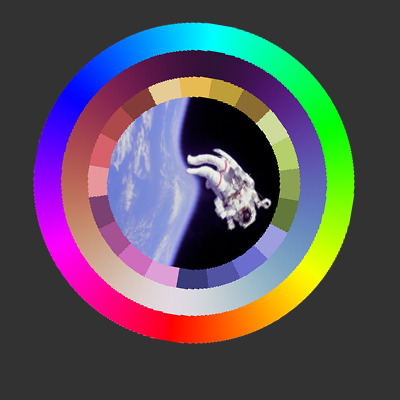} \\
        \spacemantrain{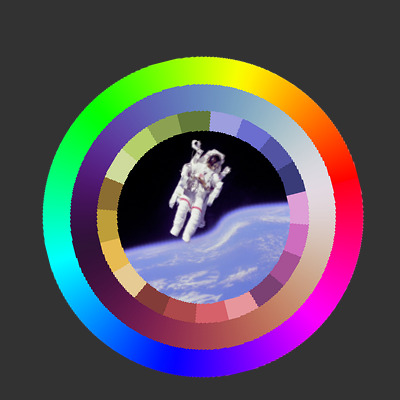} &
        \spacemantrain{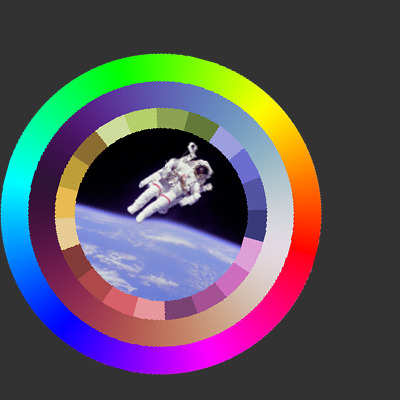} &
        \spacemantrain{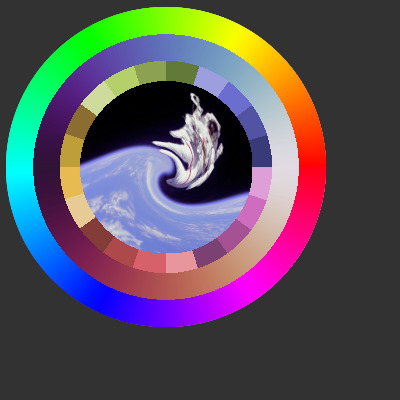} &
        \spacemantrain{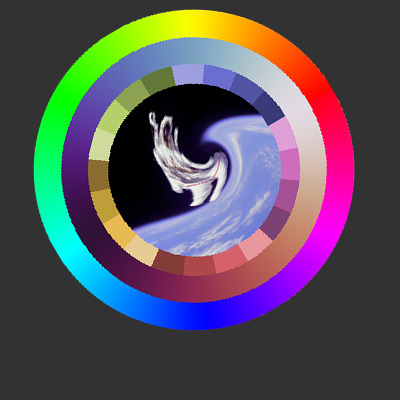}
    \end{tabular}
    \vspace{-6pt}
    \caption{Our 2D toy dataset comprised of an image with a random translation, a random rotation, and a random non-linear distortion near the center. Astronaut photo by Robert Gibson (1984, Public Domain). }
    \label{fig:2d_toy_train}
    \vspace{-5pt}
\end{figure}

%% file: fa0_2d_toy_se2_vs_translation.tex
\newcommand{\spacemanresultsmall}[1]{\includegraphics[height=2.4cm,clip]{figures/2d_toy/results/#1}}

\begin{figure}[t!]
    \centering
    \setlength{\tabcolsep}{0.5pt}
    \renewcommand{\arraystretch}{0.25}
    \begin{tabular}{cccc}
        \makebox[15pt]{\raisebox{35pt}{\rotatebox[origin=c]{90}{\small{Reconstruction}}}} \hspace{-6pt} &
        \spacemanresultsmall{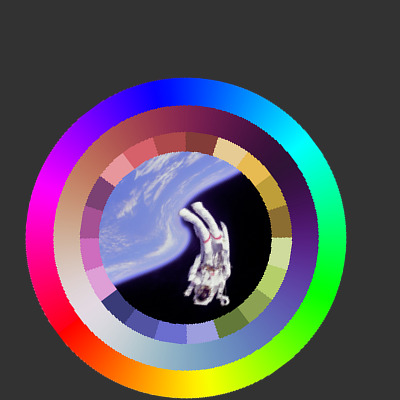} &
        \spacemanresultsmall{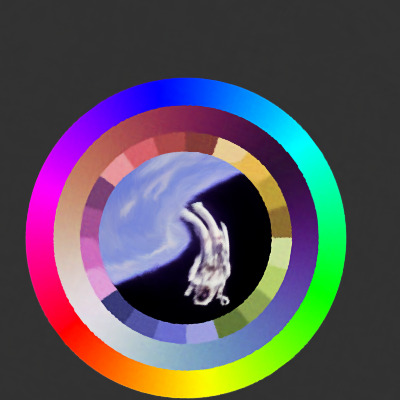} &
        \spacemanresultsmall{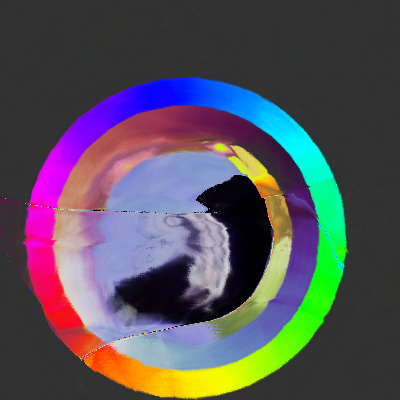}  \\
        \makebox[20pt]{\raisebox{35pt}{\rotatebox[origin=c]{90}{\small{Template}}}} \hspace{-6pt} &
        &
        \spacemanresultsmall{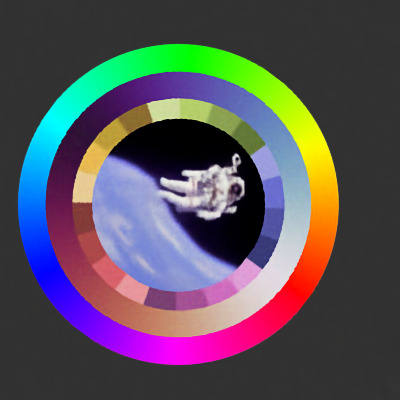} &
        \spacemanresultsmall{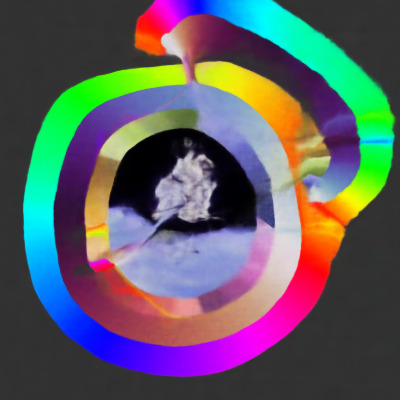}  \\
        \makebox[20pt]{\raisebox{35pt}{\rotatebox[origin=c]{90}{\small{Flow}}}} \hspace{-6pt} &
        &
        \spacemanresultsmall{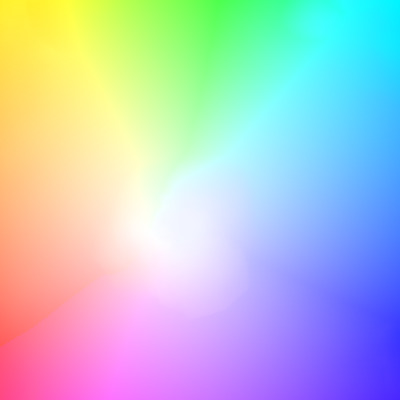} &
        \spacemanresultsmall{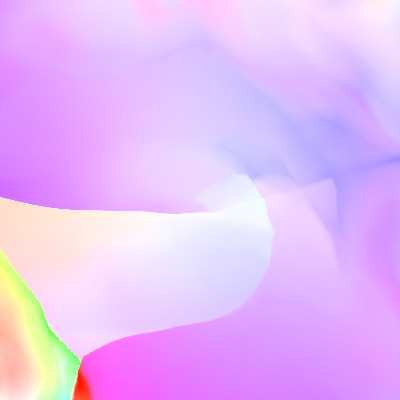}  \\ [2pt]
        &
        {\small ground truth} & 
        {\small SE(2) field} &
        {\small translation field}
    \end{tabular}
    \vspace{-6pt}
    \caption{A comparison of our SE(2) field and a translation field. The translation parameterization has difficulty rotating groups of distant pixels whereas the rigid transformation successfully finds the correct orientation.}
    \label{fig:2d_toy_se2_vs_trans}
    \vspace{-5pt}
\end{figure}

%% file: fa0_2d_toy_freqs.tex
\newcommand{\spacemanresult}[1]{\includegraphics[height=2.65cm,clip]{figures/2d_toy/results/#1}}

\begin{figure*}[t!]
    \centering
    \setlength{\tabcolsep}{0.5pt}
    \renewcommand{\arraystretch}{0.25}
    \begin{tabular}{ccccccc}
        \makebox[25pt]{\raisebox{37pt}{\rotatebox[origin=c]{90}{\small{Reconstruction}}}} \hspace{-6pt} &
        \spacemanresult{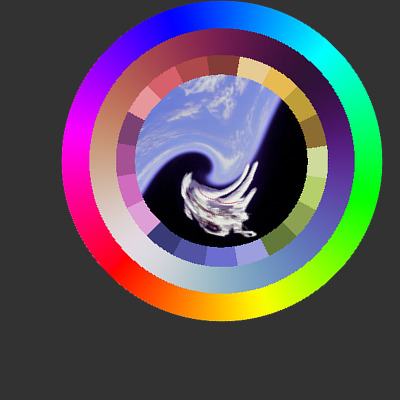} &
        \spacemanresult{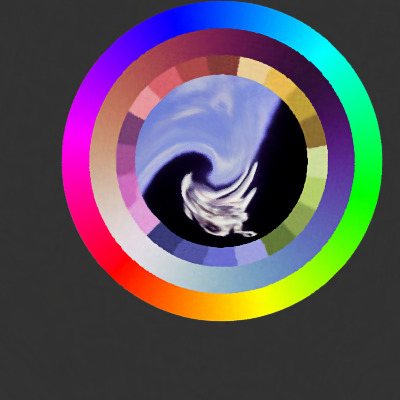} &
        \spacemanresult{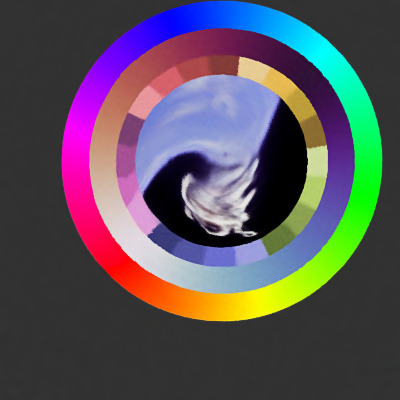} &
        \spacemanresult{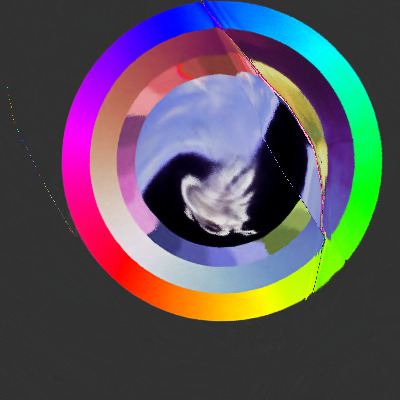} &
        \spacemanresult{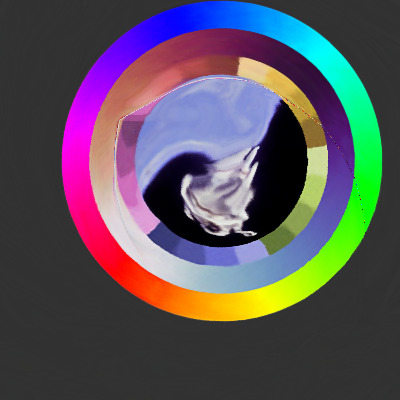} &
        \spacemanresult{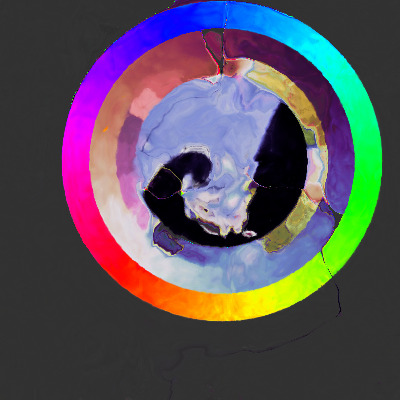} \\
         
        \makebox[20pt]{\raisebox{37pt}{\rotatebox[origin=c]{90}{\small{Abs. Error}}}} \hspace{-6pt} &
         &
        \spacemanresult{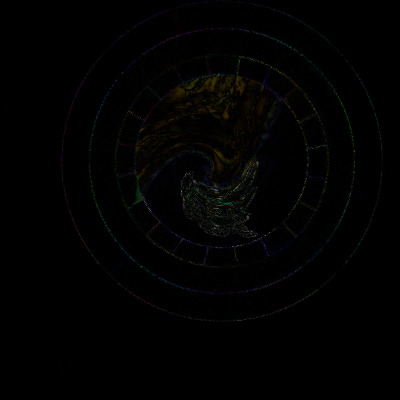} &
        \spacemanresult{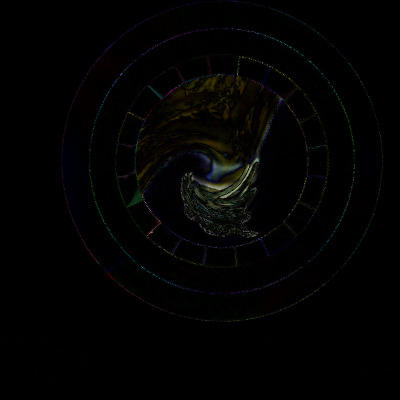} &
        \spacemanresult{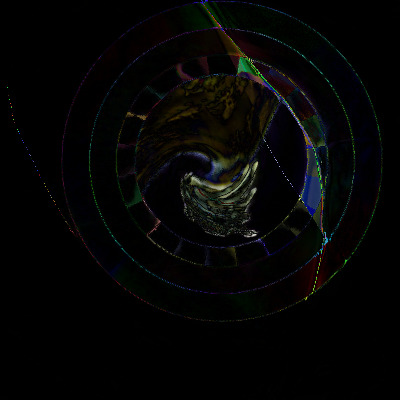} &
        \spacemanresult{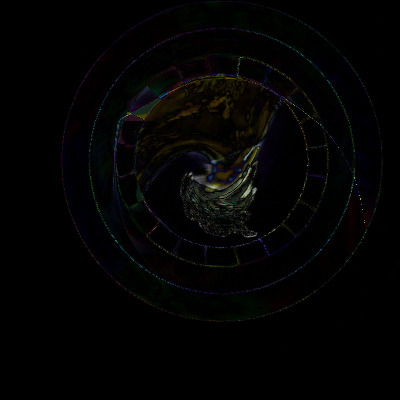} &
        \spacemanresult{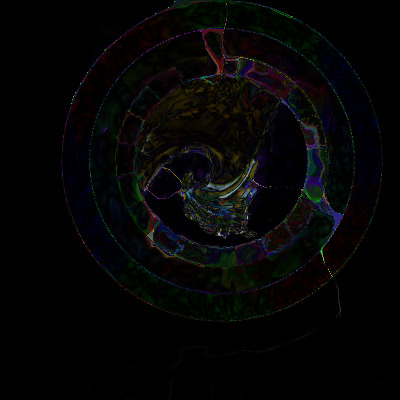} \\

        \makebox[20pt]{\raisebox{37pt}{\rotatebox[origin=c]{90}{\small{Template}}}} \hspace{-6pt} &
         &
        \spacemanresult{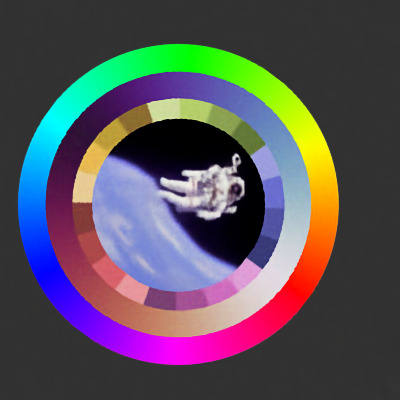} &
        \spacemanresult{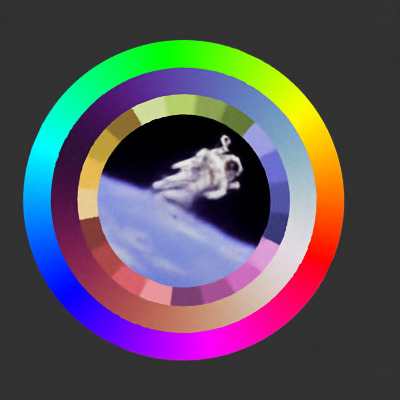} &
        \spacemanresult{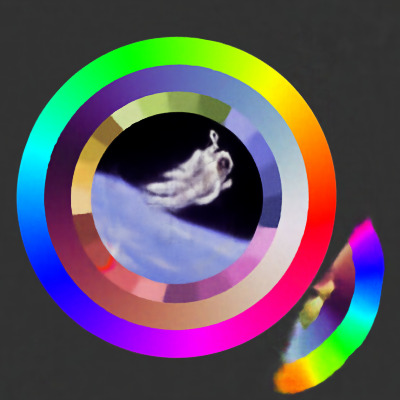} &
        \spacemanresult{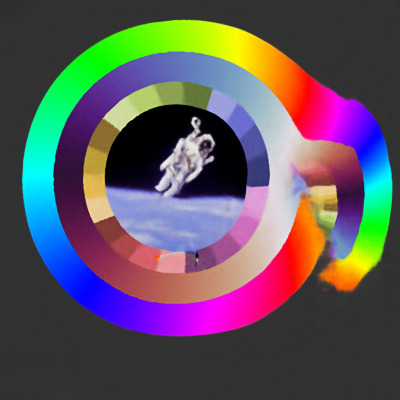} &
        \spacemanresult{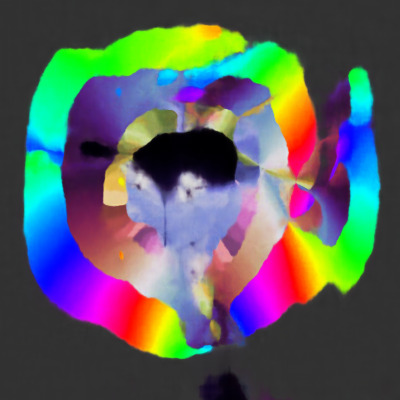} \\
       
        \makebox[20pt]{\raisebox{37pt}{\rotatebox[origin=c]{90}{\small{Flow}}}} \hspace{-6pt} &
        &
        \spacemanresult{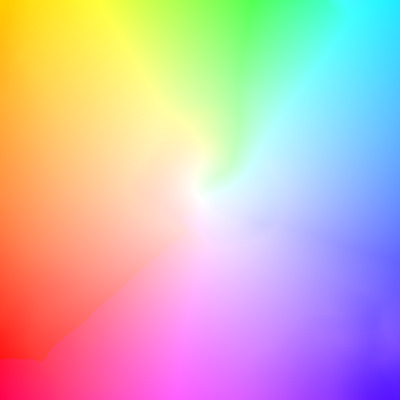} &
        \spacemanresult{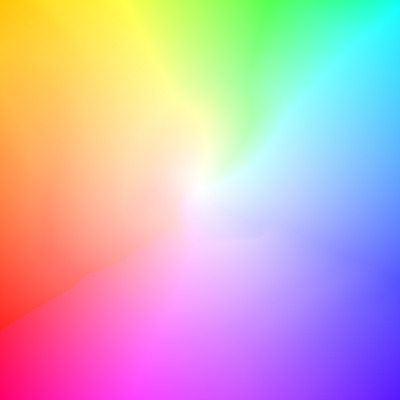} &
        \spacemanresult{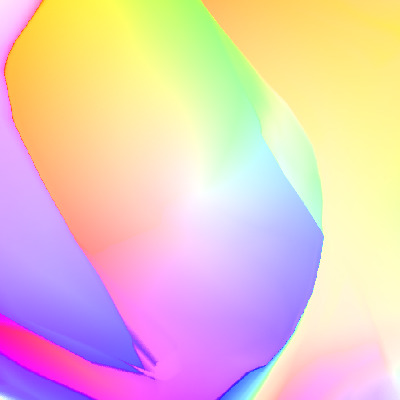} &
        \spacemanresult{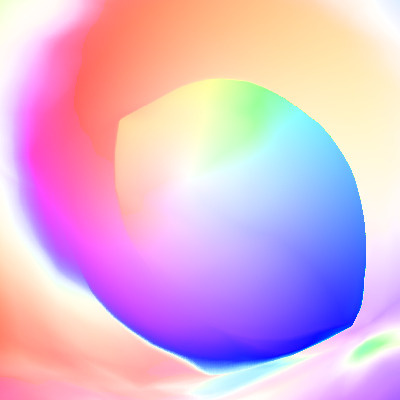} &
        \spacemanresult{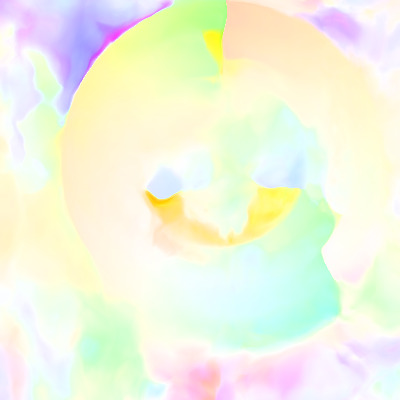} \\ [2pt]
        &
        {\small ground truth} & 
        {\small \makecell{coarse-to-fine\\($\alpha\in[0,6]$)}} & 
        {\small $m=1$} &
        {\small $m=2$} &
        {\small $m=4$} &
        {\small $m=6$}
        \\
    \end{tabular}
    \vspace{-6pt}
    \caption{We show how the optimization changes depending on the number of frequency bands in the positional encoding of the deformation field. With 1 frequency, the model find the correct orientation of all images but is unable to model the high frequency distortion near the center. If we increase the frequencies then the templates overfits early and gets stuck in bad local minima. With our coarse-to-fine technique we are less prone to local minima while also modeling high frequency details.}
    \label{fig:2d_toy_freqs}
    \vspace{-5pt}
\end{figure*}